\documentclass[twoside,11pt]{article}

\usepackage{jmlr2e}
\usepackage[utf8]{inputenc} 
\usepackage[T1]{fontenc}    
\usepackage{hyperref}       
\usepackage{url}            
\usepackage{booktabs}       
\usepackage{amsfonts}       
\usepackage{nicefrac}       
\usepackage{microtype}      
\usepackage{lipsum}
\usepackage{bm}
\usepackage{amsmath}
\usepackage{algorithm}
\usepackage[noend]{algpseudocode}
\usepackage{graphicx}
\usepackage{dsfont}
\usepackage[mathscr]{eucal}
\usepackage{xcolor}
\usepackage{appendix}
\usepackage{mathrsfs} 
\usepackage{enumitem}
\usepackage{amstext}
\usepackage{verbatim}
\usepackage{pdflscape}
\usepackage{afterpage}
\usepackage{colortbl}
\usepackage{rotating}

\usepackage[normalem]{ ulem }
\usepackage{soul}

\newtheorem{claim}{Claim}
\newtheorem{rem}{Remark}
\newtheorem{lem}{Lemma}

\newtheorem{cor}{Corollary}
\newtheorem{assumption}{Assumption}
\newtheorem{argument}{Element}

\hypersetup{
colorlinks = true,
citecolor = {magenta},
linkcolor = {blue}
}

\newcommand\underrel[2]{\mathrel{\mathop{#2}\limits_{#1}}}
\newcommand{\ppm}{$\pm$}

\newcommand{\subsubsubsection}[1]{\paragraph{#1}\mbox{}\\}
\definecolor{LightCyan}{rgb}{0.88,1,1}

\newcolumntype{a}{>{\columncolor{LightCyan}}c}

\newcounter{protocol}
\makeatletter
\newenvironment{protocol}[1][htb]{%
  \let\c@algorithm\c@protocol
  \renewcommand{\ALG@name}{Subroutine}
  \begin{algorithm}[#1]%
  }{\end{algorithm}
}
\makeatother

\usepackage{lastpage}
\jmlrheading{22}{2021}{1-\pageref{LastPage}}{3/20; Revised
	4/21}{9/21}{20-261}{Lorenzo Dall'Amico, Romain Couillet, Nicolas Tremblay}
\ShortHeadings{A Unified Framework for Spectral Clustering in Sparse Graphs}{Dall'Amico, Couillet and Tremblay}

\firstpageno{1}

\begin{document}

\title{A Unified Framework for \\Spectral Clustering in Sparse Graphs}

\author{\name Lorenzo Dall'Amico$^{\dagger}$ \email lorenzo.dall-amico@gipsa-lab.fr\\
 \name Romain Couillet$^{\dagger~*}$ \email romain.couillet@univ-grenoble-alpes.fr\\
 \name Nicolas Tremblay$^{\dagger}$ \email nicolas.tremblay@gipsa-lab.fr \vspace{0.3cm} \\ 
 \addr $^{\dagger}$GIPSA-lab
	Universit\'e Grenoble Alpes, CNRS, Grenoble INP\\
	11 Rue des Mathématiques, 38400 Saint-Martin-d'Hères, France \vspace{0.3cm} \\
	\addr $^{*}$Laboratoire d'Informatique de Grenoble (LIG), Universit\'e Grenoble Alpes\\
	700 Av. Centrale, 38401 Saint-Martin-d'Hères, France}

\editor{Tina Eliassi-Rad}

\maketitle

\begin{abstract}%
	This article considers spectral community detection in the regime of sparse networks with heterogeneous degree distributions, for which we devise an algorithm to efficiently retrieve communities. Specifically, we demonstrate that a well parametrized form of \emph{regularized Laplacian matrices} can be used to perform spectral clustering in sparse networks without suffering from its degree heterogeneity. Besides, we exhibit important connections between this proposed matrix and the now popular \emph{non-backtracking} matrix, the \emph{Bethe-Hessian} matrix, as well as the standard Laplacian matrix. Interestingly, as opposed to competitive methods, our proposed improved parametrization inherently accounts for the hardness of the classification problem.
	These findings are summarized under the form of an algorithm capable of both estimating the number of communities and achieving high-quality community reconstruction.
\end{abstract}


\section{Introduction}
\label{sec:intro}

Graph theory has found many applications in a variety of domains that span from modern biology, to technology and social sciences \citep{barabasi2016network}. Although the underlying represented systems may be fundamentally different, some common features emerge in complex networks. Some of these are  \citep{newman2003structure}:
\begin{enumerate}[noitemsep]	
	\item  Their \emph{heterogeneous degree distribution}: the number of connections of each node (their \emph{degree}) is far from following a homogeneous distribution and, instead, is typically a broad distribution \citep{barabasi1999emergence} in which few nodes (called \emph{hubs}) have a very large degree while the majority has only few connections.
	\item  \emph{Sparsity}: the average degree is typically much smaller than the size of the graph and a small fraction of all the possible connections are present \citep{barabasi2013network}; consequently, the average degree can be considered as roughly independent of the size of the graph.
	\item the \emph{clustering effect}: as a consequence of the nodes affinity encoded by edges \citep{borgatti2011network}, in real graphs typically emerge small groups of densely connected nodes, called \emph{clusters} or \emph{communities}.	
\end{enumerate}
The remainder of the article focuses on \emph{community detection}, that is in determining a good community assignment of the nodes of a given graph. While point 3.\@ of the list above is of central concern, the (nuisance) properties 1.\@ and 2.\@ need to be accounted for to guarantee high performance community assignments on real-world graphs. 

\subsection{Community Detection}

One of the most natural tasks in graph theory is community detection, i.e., the identification of similarity groups on a given graph. Practically, for an unweighted and undirected graph $\mathcal{G}(\mathcal{V},\mathcal{E})$ with $|\mathcal{V}| = n$ nodes and $|\mathcal{E}|$ edges, community detection consists in finding a non-overlapping partition of the nodes that identifies underlying communities in a completely unsupervised manner. There is no unique definition of a community, but a general criterion is to impose that nodes in the same community have more inter-connections than nodes in different communities, as a consequence of the stronger affinity among members of the same community \citep{fortunato2010community}.

\medskip

A possible way of formalizing this intuition consists in defining the class labels as the solution of an optimization problem such as \emph{MinCut}, \emph{RatioCut}, and \emph{NormalizedCut} \citep{von2007tutorial}. These definitions do not make any assumption on the underlying graph but only on what a satisfactory community assignment should be like. The aforementioned optimization problems are, however, NP-hard and only approximate solutions can be obtained. Another possible approach (adopted in the following) is instead to consider community detection as a statistical inference problem. The graph $\mathcal{G}(\mathcal{V},\mathcal{E})$ is seen as a realization of a random process in which the class-label assignment is encoded by some hidden parameters of the generative model that have to be inferred. To simultaneously account for sparsity and heterogeneity that typically characterize real graphs, in this article we consider graphs generated according to a $k$-class sparse degree-corrected stochastic block model (DC-SBM) \citep{karrer2011stochastic}, that is
\begin{equation}
\mathbb{P}(A_{ij}=1|\ell_i,\ell_j,\theta_i,\theta_j) = \theta_i\theta_j \frac{C_{\ell_i,\ell_j}}{n} \quad 1 \leq i < j \leq n
\label{eq:DC-SBM}
\end{equation}
in which $\bm{\ell} = \{1,\cdots,k\}^n$ denotes the node-wise labelling vector ($\ell_i = p$ if node $i$ is in class $p$), $\bm{\theta}$ is the vector of intrinsic connection ``probabilities'' which are used to produce an arbitrary degree distribution and are independent of the label vector $\bm{\ell}$. The entries of $\bm{\theta}$ are positive independent random variables $\theta_i \in [\theta_{\rm min}, \theta_{\rm max}]$, with $\theta_{\rm min}>0$, having first moment
$\mathbb{E}[\theta_i] = 1$ and second moment $\mathbb{E}[\theta_i^2]=\Phi$. We denote with $C \in \mathbb{R}^{k \times k}$ the symmetric class-affinity matrix, with entries independent of $n$.
The term $1/n$ in \eqref{eq:DC-SBM} bounds the average degree $\bar{d} = \frac{1}{n}\bm{1}_n^TA\bm{1}_n = O_n(1)$ to an $n$-independent value, setting the problem in the sparse regime. 

\medskip

A major advantage of a well defined generative model results in statistically tractable conclusions, like the existence in asymptotically large graphs ($n\to\infty$) generated from the (degree-corrected) stochastic block model of a limiting \emph{detectability threshold}. Specifically, for $k = 2$ classes, one can in general identify a parameter $\alpha$, function of the community statistics and such that, beyond a threshold ($\alpha > \alpha_c$), partial label reconstruction can be theoretically achieved, whereas below the threshold ($\alpha < \alpha_c$) no algorithm can perform better than random guess. For $k > 2$ classes, the situation is more involved: it is conjectured \citep{decelle2011asymptotic} that there  exists an \emph{easy} detection zone in which a non-trivial clustering can be found in polynomial time, a \emph{hard} detection zone in which non-trivial clustering can be found but only in exponential time and an \emph{impossible} detection zone in which no algorithm can find a non-trivial partition. 

In the very specific setting where $k=2$ and both classes are of equal size ($n/2$), the detectability threshold assumes a simple expression. Letting $C_{\ell_i,\ell_j} = c_{\rm in}$ if $\ell_i = \ell_j$ and $c_{\rm out}$ otherwise, in the \emph{stochastic block model} (for which $\bm{\theta} = \bm{1}_n$) it was initially conjectured \citep{decelle2011asymptotic} and later proved \citep{mossel2012stochastic,massoulie2014community} that the condition to non-trivial clustering is given by
\begin{equation}
\alpha \equiv \frac{c_{\rm in} - c_{\rm out}}{\sqrt{c}} > 2 \equiv \alpha_c,
\label{eq:detectability_SBM}
\end{equation} 
where $c = (c_{\rm in} + c_{\rm out})/2$. Equation~\eqref{eq:detectability_SBM} was later generalized in \citep{gulikers2015impossibility, gulikers2016non} to the DC-SBM but now with $\alpha_c = 2/\sqrt{\Phi}$ (we recall that $\Phi = \mathbb{E}[\theta_i^2]$).

\medskip

If on one side random generative models like the DC-SBM give a solid theoretical understanding, the applicability to real graphs of a DC-SBM inspired clustering algorithm is a serious concern to be taken into account and which will be addressed here. 

\subsection{Spectral Clustering: Related Work}

\setcounter{algorithm}{-1}

\begin{algorithm}[b!]
	\begin{algorithmic}[1]
		\State \textbf{Input} : $\mathcal{G}$, $k$
		\State Choose the matrix $M\in \mathbb{R}^{n \times n}$, a suited representation of the graph
		\State  Find the $k$ largest (or smallest) eigenvalues of $M$ and stack the corresponding eigenvectors in the columns of a matrix $\bm{X} \in \mathbb{R}^{n\times k}$
		\State  (\emph{Optional}) Normalize the rows of $\bm{X}$
		\State Estimate community labels as the output of some small dimensional clustering method (such as \emph{k-means} or \emph{expectation maximization}) performed on the points in $\mathbb{R}^k$ defined by the rows of $\bm{X}$
		\State \Return Estimated label community vector.
		\caption{A typical spectral clustering algorithm on a graph $\mathcal{G}$ with $k$ classes}
		\label{alg:0}
	\end{algorithmic}
\end{algorithm}

Regardless of the definition of a community considered (as a solution of an optimization problem or of an inference problem), different classes of algorithms can be considered to find exact or approximate solutions for community detection. Spectral clustering is a class of algorithms according to which each node of the graph is mapped to a low dimensional space. Spectral clustering hence provides a small dimensional graph embedding from which nodes are partitioned using common clustering algorithms, like \emph{k-means}. The name \emph{spectral} comes from the fact that the embedding is obtained using some eigenvectors (later referred to as \emph{informative}) of a suitable graph representation matrix. Spectral clustering allows one to obtain an approximate fast solution of the optimal class assignment both in the case in which clusters are defined as the solution of an NP-hard optimization problem \citep{von2007tutorial, zhang2018understanding} or of an inference problem \citep{krzakala2013spectral}.

Algorithm~\ref{alg:0} provides the general structure of a $k$-class spectral clustering algorithm based on an ``affinity'' matrix $M$.
Algorithm~\ref{alg:0} relies on the fact that some of the eigenvectors of these appropriate affinity matrices contain the class structure of the graph. Step~3 of Algorithm~\ref{alg:0} underlines that the informative eigenvalues are generally found to be the largest or smallest eigenvalues, so that there is an algorithmically clear definition of which eigenvectors should be used for clustering.

\medskip

In order to review the most relevant spectral clustering algorithms for community detection, let us define the adjacency matrix $A \in \{0,1\}^{n\times n}$ of the graph $\mathcal G$ by $A_{ij} = \mathds{1}_{(ij)\in\mathcal{E}}$ where $\mathds{1}_{x}$ equals $1$ if the condition $x$ is verified and zero otherwise, and the diagonal degree matrix $D = {\rm diag}(A\bm{1}_n)$, where $\bm{1}_n$ is the vector with all entries equal to one. In \citep{fiedler1973algebraic}, and in the case of $k=2$ communities, it was proposed to reconstruct communities using the eigenvector corresponding to the second smallest eigenvalue of the \emph{combinatorial graph Laplacian} matrix $L = D - A$. It was then shown \citep[see \emph{e.g.}][]{von2007tutorial} that this eigenvector provides a relaxed solution of the \emph{RatioCut} problem.
Based on this result, one can build a spectral algorithm that, referring to Algorithm~\ref{alg:0}, sets $M = L$; in this case, the informative eigenvectors correspond to the smallest eigenvalues and step~4 is not performed. Similarly, the \emph{normalized graph Laplacian} matrices ($L^{\rm rw} = D^{-1}A$ and $L^{\rm sym} = D^{-1/2}AD^{-1/2}$) can be used to solve a relaxed form of the \emph{NCut} problem \citep{von2007tutorial}. In \citep{shi2000normalized}, $M = L^{\rm rw}$, the informative eigenvectors correspond to the $k$ largest eigenvalues and step~4 is again not performed, while in \citep{ng2002spectral}, $M = L^{\rm sym}$, the informative eigenvectors correspond to the $k$ largest eigenvalues and step~4 is performed. Another classical choice of matrix $M$ is in~\citep{Newman-2006}, where the authors propose an alternative method to define communities through a NP-hard optimization problem by maximizing the so-called modularity of the graph. For $k=2$ classes, a relaxed solution can be obtained by considering the largest eigenvector of the modularity matrix  $A - \frac{\bm{d}\bm{d}^T}{2|\mathcal{E}|}$, where $\bm{d} = A\bm{1}_n$ is the degree vector. For more than two communities, the algorithm is slightly more involved and deviates from the general structure of Algorithm~\ref{alg:0}.\\

\begin{figure}[t!]
	\centering
	\includegraphics[width = 0.8\columnwidth]{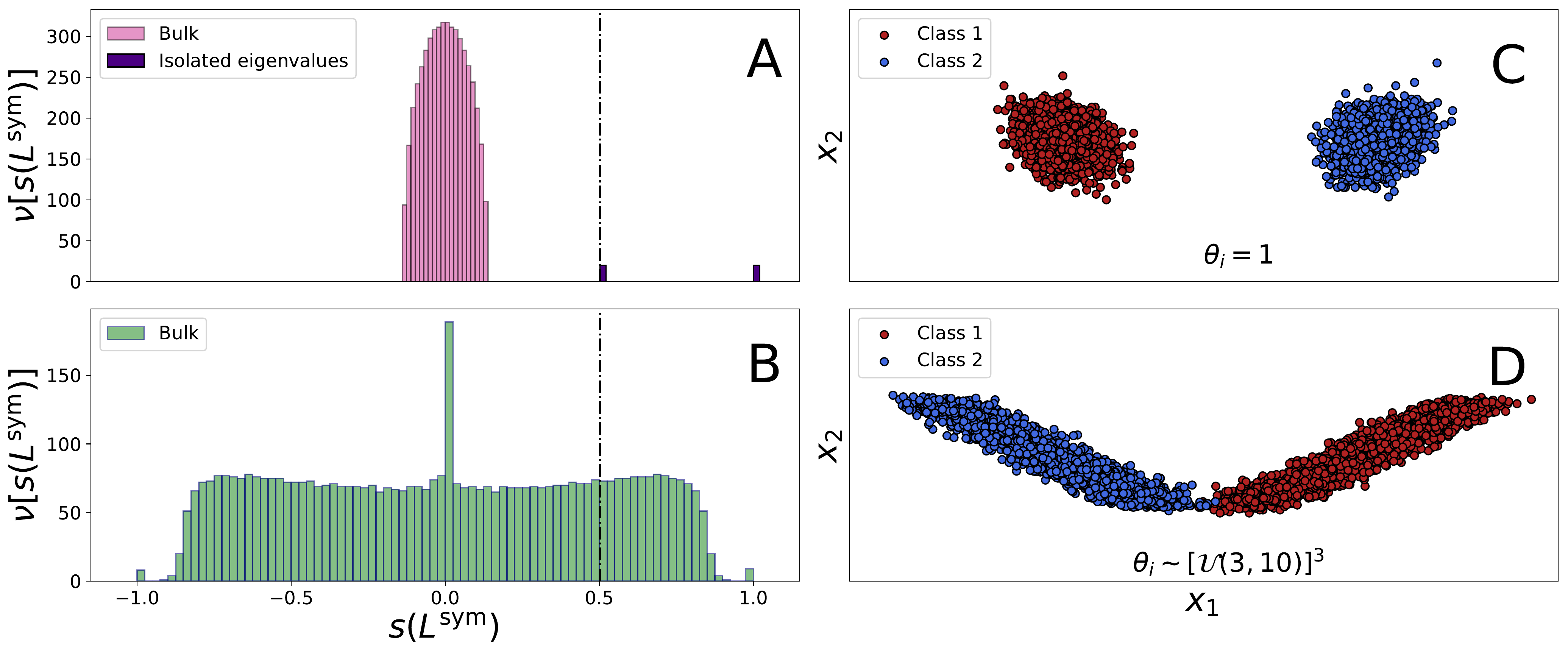}
	\vspace{-0.3cm}
	\caption{For all the four plots $n = 5000$, $k = 2$, $\bm{\theta} = \bm{1}_n$ if not otherwise specified. (Left) spectrum of the matrix $L^{\rm sym}$ in the: A) dense regime ($c_{\rm in}/n = 0.06$, $c_{\rm out}/n = 0.02$); B) sparse regime ($c_{\rm in} = 6$, $c_{\rm out} = 2$). (Right) first eigenvectors $\bm{x}_1$ vs second eigenvector $\bm{x}_2$ of $L^{\rm sym}$ in the dense regime for: C) $\bm{\theta} = \bm{1}_n$; D) $\theta_i \sim [\mathcal{U}(3,10)]^3$.}
	\label{fig:sparse_het}
	\vspace{-0.7cm}
\end{figure}

Many results justifying the earlier methods exist in the asymptotic limit where $n \gg 1$, in the so-called \emph{dense} regime, in which the average degree (number of connections) grows with the size ($n$) of the network. However, as we anticipated in the introduction, real networks tend instead to be \emph{sparse} as well as \emph{heterogeneous} in their degree distribution, following broad distributions such as power laws. These two characteristics (sparsity and heterogeneity) make it difficult to theoretically predict the behavior and performances of Algorithm~\ref{alg:0} and to choose a proper matrix $M$ to work with. From a linear algebra viewpoint, sparsity in the node degrees has been shown to spread the eigenvalues of the Laplacian matrix, with the deleterious effect to ``swallow'' the isolated (smallest or largest) informative eigenvalues within the so-called ``bulk'' of uninformative eigenvalues: as a result, the corresponding informative eigenvectors are no longer found to be associated with dominant (largest or smallest) eigenvalues, as shown in Figure \ref{fig:sparse_het}B, and as opposed to Figure \ref{fig:sparse_het}A. Besides, by losing the isolation of informative eigenvalues, the associated eigenvectors tend to merge with the eigenvectors associated to close-by (non-informative) eigenvalues. Heterogeneity in the degrees also induces a severe negative effect: it modifies the amplitude of the $i$-th entry (for all $1\leq i\leq n$) of the informative eigenvectors by a non-trivial function of the degree of node $i$. This further compromises the efficiency of the last classification step (usually performed through \emph{k-means}), as shown in Figure \ref{fig:sparse_het}D, to be opposed to Figure \ref{fig:sparse_het}C. Several contributions have thus tackled the challenging problem of devising efficient spectral clustering in the sparse regime and for an arbitrary degree distribution.

\medskip 

One prominent line of research was devoted to defining spectral algorithms capable of retrieving communities as soon as theoretically possible ($\alpha > \alpha_c$, Equation~\eqref{eq:detectability_SBM}) on sparse graphs generated from the SBM, for which we recall that there is a sharp transition between the ``detectable'' and ``undetectable'' situations. The authors of~\citep{krzakala2013spectral} proposed an algorithm based on the (non-symmetric) non-backtracking matrix $B_{(ij)(lm)} = \delta_{jl}(1-\delta_{im})$, with $B~\in~ \{0,1\}^{2|\mathcal{E}|\times 2|\mathcal{E}|}$. In 
\citep{massoulie2014community} it was indeed shown that the eigenvectors associated to the largest (in modulus) eigenvalues of $B$ have a non-trivial correlation with the actual underlying communities, as soon as $\alpha > \alpha_c$. A closely related algorithm is the one proposed in~\citep{saade2014spectral} which instead uses the eigenvectors attached to the smallest eigenvalues of the Bethe-Hessian matrix $H_r = (r^2-1)I_n + D -  rA \in  \mathbb{R}^{n \times n}$, for $r=\sqrt{\rho(B)}$ ($B$ being the non-backtracking matrix just defined and $\rho(\cdot)$ the spectral radius). These algorithms have the benefit of achieving non-trivial classification down to the detectability threshold, but they have only been investigated under a sparse SBM setting; hence they do not cope with degree heterogeneity. The extension to the  sparse DC-SBM case was treated in~\citep{gulikers2016non}, in which the authors show that spectral clustering on $B$ also works down to the (generalized) threshold.

All these results are powerful as they propose algorithms capable of reaching the information-theoretic threshold, but they also have inherent weaknesses as they only guarantee a positive correlation of their output classification with the underlying true structure. Specifically, these positive correlations do not imply that the classification performance is maximal. In particular, even in~\citep{gulikers2016non} where spectral clustering on the DC-SBM is studied, the problem of eigenvector pollution due to degree heterogeneity (which is one of our main focuses in this article) is not discussed and \emph{a fortiori} not corrected.

\medskip

Another line of research which studies the reconstruction of sparse DC-SBM networks suggests to exploit regularization (i.e., using $D_{\tau} = D + \tau I_n$ and $A_{\tau}=A + \tau\bm{1}_n\bm{1}_n^T$ in place of $D$ and $A$) as a solution to maintain the location of the informative eigenvalues in their dominant (smallest or largest) positions \citep{amini2013pseudo,joseph2013impact,lei_consistency_2015,le2018concentration}. A particularly interesting method in this direction is proposed in \citep{qin2013regularized} which uses the \emph{regularized symmetric reduced Laplacian} matrix $L_{\tau}^{\rm sym} = D_{\tau}^{-1/2}AD_{\tau}^{-1/2}$,
for $\tau = \bar{d}$, the average degree. In terms of Algorithm~\ref{alg:0}, the algorithm proposed in \citep{qin2013regularized} sets $M = L_{\tau}^{\rm sym}$, searches for the $k$ eigenvectors corresponding to the $k$ largest eigenvalues of $M$, and then performs the normalization step~4 on the rows of the resulting matrix $\bm{X}$. While \citep{qin2013regularized} deals with graphs with arbitrary degree distribution (generated from the DC-SBM), the authors do not discuss whether their algorithm can achieve non-trivial clustering down to the threshold, i.e. as soon as $\alpha > \alpha_c$. We will bring new conclusions in this direction in Section~\ref{sec:reg_lap}.

\subsection{Contributions}

In the state-of-the-art methods we presented, while sparsity is not properly accounted for in \citep{shi2000normalized, ng2002spectral}, the line of research based on $B$ (and consequently $H_r$) \citep{krzakala2013spectral,saade2014spectral,bordenave2015non,gulikers2016non} does account for sparsity and provides methods to reach the detectability threshold; yet, all they guarantee is the possibility to obtain a positive, possibly suboptimal, correlation between the algorithm output and the underlying community structure. As for the works on regularization \citep{qin2013regularized, le2015estimating}, they only establish theoretical results of perfect community recovery far from the (most interesting) detection threshold.

The present work solves both the issues of sparsity and heterogeneity at once by proposing a simple spectral clustering scheme which, under a sparse DC-SBM setting, provably performs non-trivial clustering as soon as $\alpha>\alpha_c$ and is robust to degree heterogeneity, as it retrieves eigenvectors not infected by the node degrees. This claim is supported by three parallel arguments having the side advantage to unify in a joint framework the ideas of \citep{krzakala2013spectral, saade2014spectral,shi2000normalized, fiedler1973algebraic, qin2013regularized}.

We further discuss how our algorithm, analyzed under the DC-SBM assumption, can be extended to real graphs. In particular, the algorithm provides an accurate estimate of the number of communities when unknown, and is observed to systematically achieve good scores (both in terms of modularity or log-likelihood of the DC-SBM posterior probability) on all real-world graphs which we experimented on. 

In summary, our main contributions are as follows:
\begin{enumerate}[noitemsep]
    \item We devise a practical spectral algorithm exploiting the eigenvectors corresponding to the smallest eigenvalues of the Bethe-Hessian matrix $H_r$ for a set of properly chosen values of $r$. We further provide strong arguments and extensive numerical simulations to claim that the proposed method achieves non-trivial reconstruction down to the DC-SBM detectability threshold.
    \item The proposed algorithm is extensively tested on synthetic DC-SBM graphs (for which the performance is measured in terms of overlap between the estimated and ground-truth labels) as well as on real networks (for which the performance is measured in terms of modularity and DC-SBM log-likelihood). In both cases our algorithm outperforms or is on par with the standard competing spectral techniques for sparse graphs.
	\item In passing, we provide a new vision to spectral clustering, and in particular a compelling new connection between our proposed approach and all aforementioned standard spectral methods, so far treated independently. In particular, we show that the \emph{regularized reduced random walk Laplacian} matrix $L^{\rm rw}_{\tau} = D_{\tau}^{-1}A$ (that shares the same eigenvalues as $L_{\tau}^{\rm sym}$) can be used to perform non-trivial clustering down to the detectability threshold for a choice of $\tau$ strongly related to our proposed algorithm.
\end{enumerate}

A Python implementation of the codes needed to reproduce the results of this article can be found at \href{https://lorenzodallamico.github.io/codes/}{lorenzodallamico.github.io/codes/}: all data, experiments and running times refer to this implementation. On top of the Python codes, we developed a more efficient Julia implementation of the proposed Algorithm~\ref{alg:2}, available in the \href{https://github.com/lorenzodallamico/CoDeBetHe.jl}{CoDeBetHe.jl} package which gains a factor ${\rm x}5 \sim {\rm x}10$ in terms of computational time over the Python implementation.

\medskip

The remainder of the article is organized as follows:  Section \ref{sec:connection} formally introduces the problem at hand and provides three complementary supporting arguments to the claim that the eigenvectors corresponding to the smallest eigenvalues of a set of Bethe-Hessian matrices $H_r$, with different values of $r$, can be used to reconstruct classes. Section~\ref{sec:reg_lap} draws the connection between the above $H_r$ and the regularized Laplacian and proves that, under a correct choice of $\tau$ related to $r$ above, the matrix $L_{\tau}^{\rm rw}$ is also a suitable candidate to reconstruct communities. 
Section~\ref{sec:sim} introduces the final form of the proposed algorithm and provides simulation outputs on real networks. Section~\ref{sec:conclusions} finally closes the article.

\subsection{Notation}

\begin{itemize}[noitemsep]
	\item Matrices are indicated with standard font capital letters ($M$). The only exception is $\bm{X}$, the matrix concatenating the eigenvectors of the spectral clustering algorithm.
	\item $M_{\bullet,p}$ indicates the $p$-th column of $M$ and $M_{p,\bullet}$ its $p$-th row.
	\item Vectors are denoted in bold $\bm{v}_p$, while scalar and vector elements in standard font $v_{p,i}$.
	\item We denote by $s_i^{\uparrow}(M)$ the $i$-th smallest eigenvalue of a Hermitian matrix $M$, i.e., $s_1^{\uparrow}(M)\leq s_2^{\uparrow}(M) \leq\dots\leq s_n^{\uparrow}(M)$, and $s_i^{\downarrow}(M)$ the $i$-th largest. For a non-Hermitian matrix $M$, we denote by  $s_{R,i}^{\uparrow/\downarrow}(M)$ the $i$-th smallest/largest real eigenvalue, while $s_i^{\uparrow/\downarrow_{|\cdot|}}(M)$ indicates the $i$-th smallest/largest \emph{in modulus}. When using the notation $s_i(M)$ we consider a generic eigenvalue of $M$.
	\item $S(M) = \{s_i(M),~1\leq i\leq n \}$  denotes the set of eigenvalues of $M\in\mathbb{R}^{n\times n}$.
	\item The notation $\bm{1}_n$ indicates the vector of size $n$ containing all ones: $\bm{1}_n^T = (1,1,\cdots,1)$.
	\item We denote the neighborhood of the node $i$ of a graph with adjacency matrix $A$ by  $\partial i = \{j \in \mathcal{V}~:~A_{ij}=1\}$.
\end{itemize}

\section{Informative Eigenvectors of $B$ and $H_r$ in the Sparse Regime}
\label{sec:connection}

Consider an undirected and unweighted graph $\mathcal{G}(\mathcal{V},\mathcal{E})$ with $|\mathcal{V}|=n$ nodes and $|\mathcal{E}|$ edges.
Our objective is to devise a community detection algorithm on $\mathcal{G}$ which is resilient to the typically  heterogeneous and sparse nature of real graphs, and also accounts for the fact that the actual grouping of nodes in communities is not clear-cut in a real graph (i.e., there usually exists no ``optimal'' number of communities and no ``ground-truth'' for the allocation of nodes into communities).

While our proposed algorithm (Algorithm~\ref{alg:2}) is applicable to any arbitrary graph (as discussed in Section~\ref{sec:sim}), for analytic purposes, we start the article by assuming that $\mathcal G$ is constituted of exactly $k$ communities and generated as a sparse DC-SBM graph; that is, the edges of $\mathcal G$ are drawn independently  according to Equation~\eqref{eq:DC-SBM}. Before delving into the technical details though, we first provide a short digest of our main result.

\subsection{Informal Statement of the Main Result}

Let us present our central result (Claim~\ref{claim:1}) and the resulting new algorithm for community detection in simple terms. The result revolves around the shape of the eigenvectors attached to the smallest (algebraic) eigenvalues of the Bethe-Hessian matrix  $H_r \in \mathbb{R}^{n \times n}$, defined as
\begin{align}
H_r = (r^2-1)I_n + D - rA,
\label{eq:def_BH}
\end{align}
for carefully chosen values of $r \geq 1$; we recall that $A$ and $D$ are the adjacency and degree matrices of $\mathcal{G}(\mathcal{V},\mathcal{E})$, respectively, while $I_n$ is the identity matrix of size $n$.

To each $k$-class graph, community detection is performed by first associating a set of $k$ Bethe-Hessian matrices $\{H_{\zeta_p}\}_{p = 1,\dots,k}$, with $\zeta_p$ defined so that the $p$-th smallest eigenvalue of $H_{\zeta_p}$ is equal to zero. We then extract from each matrix the eigenvector $\bm{x}_p$ attached to the zero eigenvalue (so that $H_{\zeta_p}\bm{x}_p = 0$). The eigenvectors $\{\bm{x}_p\}_{p = 1,\dots,k}$ are stacked in the columns of the matrix $\bm{X}$ defined in Algorithm~\ref{alg:0} and used to produce the small dimensional node embedding on which the \emph{k-means} algorithm is applied. 

Claim~\ref{claim:1} justifies the relevance of the above procedure as a highly performing community detection method. Specifically, for a $k$-class DC-SBM graph, Claim~\ref{claim:1} states that:
\begin{itemize}[noitemsep]
	\item The largest value of $\zeta_p$ that can be defined is $\zeta_k$. We show that, as a consequence,  the $(k+1)$-th smallest eigenvalue of $H_r$ is always positive for all $r$ and the maximal number of negative eigenvalues of $H_r$ as function of $r$ is precisely equal to $k$. This allows one to build an estimator for the number of classes. 
	\item The eigenvectors $\{\bm{x}_p\}_{p = 1,\dots,k}$ and the $k$ dominant eigenvectors of the non-backtracking matrix $B$ are correlated to the class structure under the same hypothesis. This, for $k=2$ classes of equal size, means that $\{\bm{x}_1, \bm{x}_2\}$ can be used to reconstruct communities as soon as theoretically possible, i.e. when $\alpha > \alpha_c$, (Equation~\eqref{eq:detectability_SBM}).
	\item The entries of $\{\bm{x}_p\}_{p = 1,\dots, k}$ are not polluted by the degree heterogeneity; hence do not suffer the problem shown in Figure~\ref{fig:sparse_het}D that hinders the performance of \emph{k-means}.
\end{itemize}

The properties of $\{\bm{x}_p\}$ allow us to define an efficient algorithm for community detection in sparse graphs with a heterogeneous degree distribution, solving at once the two main challenges discussed in Section~\ref{sec:intro}.

\subsection{Main Result}

In this section, we proceed with the formal statement of Claim~\ref{claim:1}, together with a clear definition of the hypotheses under which it is formulated.

\subsubsection{Model and Setting}
Let $\mathcal G$ be a $k$ class DC-SBM generated graph. We consider graphs $\mathcal{G}$ whose classes are not necessarily of equal size and denote by $0<\pi_{p}<1$, for every class $1\leq p\leq k$, the fraction of nodes belonging to class $p$.
We correspondingly define the diagonal matrix $\Pi = {\rm diag}(\pi_1,\ldots,\pi_k)$ (note that in particular ${\rm Tr}(\Pi) = 1$).
We next denote by $(\nu_p,\bm{v}_p)$ the eigenvalue-eigenvector pairs of $C\Pi$, i.e., $C\Pi\bm{v}_p = s_p^{\downarrow}(C\Pi)\bm{v}_p \equiv \nu_p \bm{v}_p$, where we recall $s_p^{\downarrow}(C\Pi)$ denotes the $p$-th largest eigenvalue of $C\Pi$, hence the eigenvalues are sorted as $\nu_1\geq \ldots\geq \nu_k$. Note that the eigenvalues of $C\Pi$ are all real as $C\Pi$ has the same spectrum as $\Pi^{1/2}C\Pi^{1/2}$, which is a symmetric and real matrix (thus diagonalizable in $\mathbb{R}$). In order to set ourselves under an asymptotically non-trivial community detection setting, the following assumptions on matrix $C\Pi$ need to be posed: 
\begin{assumption}
The symmetric matrix $C$ and the diagonal matrix $\Pi$ with ${\rm Tr}(\Pi) = 1$ have entries independent of $n$ and satisfy the following hypotheses:
\begin{enumerate}
    \item $(C\Pi)_{pq} > 0,~\forall~1\leq p,q\leq k$
    \item $C\Pi\bm{1}_k = c\bm{1}_k$, for some positive $c$
    \item For $1\leq p\leq  k$, $\nu_p > \sqrt{\frac{c}{\Phi}}$. 
\end{enumerate}
\label{ass:cpi}
\end{assumption}

Assumption~\ref{ass:cpi} is fundamental to our analysis. Its three key requests deserve a detailed motivation and explanation:
\begin{enumerate}
    \item $(C\Pi)_{pq} > 0$: this implies that there is a non-null probability of connection between any two classes.
    As a consequence, one may apply the Perron-Frobenius theorem on $C\Pi$: the eigenvalue of $C\Pi$ of largest modulus, denoted with $c>0$, is positive, simple and its corresponding eigenvector is the only one that can be chosen with all positive entries. By definition of the eigenvalues $\{\nu_p\}_{p=1}^k$, we have in particular $c=\nu_1$.
    \item $C\Pi\bm{1}_k = c\bm{1}_k$: this assumption imposes that the Perron-Frobenius eigenvector of $C\Pi$ is the vector of all ones, $\bm{1}_k$. 
    From Equation~\eqref{eq:DC-SBM} and the independence of the entries of $A$, one can show that this assumption implies that $\mathbb{E}[\bar{d}]$, the expected value of the average degree, is equal to $c$. In the two-class symmetric case, $c = (c_{\rm in} + c_{\rm out})/2$ as introduced in Equation~\eqref{eq:detectability_SBM}. Assumption~\ref{ass:cpi} also importantly ensures that the average degree does not depend on the class; in fact, the expected degree of  each node $i$ belonging to class $p$ equals $\theta_i (C\Pi\bm{1}_k)_p = \theta_i c$, which is independent of $p$. 
   this is a standard assumption in the literature and sets the problem under a scenario for which the degree of each node, being independent of its label, does not bring any information on the class structure \citep{ krzakala2013spectral,decelle2011asymptotic,bordenave2015non}. 
     \item $\nu_p > \sqrt{c/\Phi}$ : this assumption is made for consistency between the actual number of classes $k$ and the number of classes $\hat{k}$ that can be detected with the spectral methods described in the following and consists in imposing $k = \hat{k}$. Recalling that $\nu_1 = c$, this assumption notably implies that $c\Phi > 1$, which is a necessary and sufficient condition to have a giant component in $\mathcal{G}$, as stated in Theorem~\ref{th:giant} in Appendix~\ref{app:giant}. 
     In the case of two classes of equal sizes, the condition $\nu_p > \sqrt{c/\Phi}$ is equivalent to setting the problem above the detectability threshold $\alpha > \alpha_c$.
    As a consequence of this assumption, for all large $n$ with high probability, the spectrum of the affinity matrices under study ($B$, $H_r$, $L^{\rm rw}_\tau$, etc.) can be decomposed as the union of a set of contiguous (non-informative) eigenvalues, collectively referred to as the \emph{bulk}, and of a set of $k$ (informative) \emph{isolated} eigenvalues $s_1,\ldots,s_k$ found at non-vanishing distance from the bulk (that is, up to multiplicity, $s_i$ is isolated if for some $\epsilon > 0$, we have for all $n$ and $j\neq i$, $|s_i-s_j|> \epsilon$). As opposed to bulk eigenvalues, the isolated eigenvalues are referred to as \emph{informative} as their corresponding eigenvectors are non-trivially correlated to the community structure. 
   \end{enumerate}

It must be noted that, from Assumption~\ref{ass:cpi}, we necessarily have $\nu_p > 0$ for all $1\leq p\leq k$. However, the results of this paper remain valid if one replaces the third point of Assumption~\ref{ass:cpi} by $|\nu_{p}| > \sqrt{\frac{c}{\Phi}}$ (that is, $\nu_p$ for $p\geq 2$ may be of arbitrary sign and only its modulus may be lower-bounded; $\nu_1=c$, being the Perron-Frobenius eigenvalue, is necessarily positive in any case). As the notations to describe this more general case are more cumbersome, we prefer here to focus on the simpler case where $\nu_{p} > \sqrt{\frac{c}{\Phi}}$ in the core of the article and defer the discussion of the general case to Section~\ref{subsec:extension}.

\subsubsection{Characterization of the Informative Eigenvectors of $H_r$ and $B$}

Before stating the main claim of the article, we first need to introduce the values $r=\zeta_p$ which must be carefully selected to best operate on the Bethe Hessian $H_r$.
 \begin{definition}[$\zeta_p$] Consider an arbitrary graph $\mathcal{G}$ composed of $n_{\rm CC}$ connected components. Let $\mathcal{G}^{(j)}$ be the subgraph reduced to the $j$-th connected component, and $H_{r}^{(j)}$ its associated Bethe-Hessian matrix, for $r \in \mathbb{R}$ (Equation~\eqref{eq:def_BH}). Let $p$ be an integer between $1$ and $n$. We define, if it exists, $\zeta_p^{(j)}$ as:
\begin{align}
	\label{eq:zeta_j}
	\zeta_p^{(j)}= \underrel{r\geq1}{\rm min}\{r : s_p^{\uparrow}(H^{(j)}_r) = 0\},
\end{align} 
where $s_p^{\uparrow}(\cdot)$ is the $p$-th smallest eigenvalue. In words, $\zeta_p^{(j)}$ is, if it exists, the smallest value of the parameter $r\geq1$ such that the $p$-th smallest eigenvalue of $H_r^{(j)}$ is null.
 \end{definition}

Here we list a few important properties of these values:
\begin{enumerate}
    \item At $r=1$, $H_1^{(j)} = D^{(j)} - A^{(j)}$ is the combinatorial Laplacian of the subgraph associated to the $j$-th connected component. As this subgraph is by definition connected, it is well known~\citep{chung_spectral_1997} that its smallest eigenvalue is null and that $s_{p\geq2}^{\uparrow}\left(H_1^{(j)}\right) > 0$. Consequently, $\zeta_{1}^{(j)}=1$ always exists and if $\zeta_{p\geq 2}^{(j)}$ exists, it is strictly superior to 1.
    \item If $\zeta_{p\geq 2}^{(j)}$ exists, then $\zeta_{2}^{(j)}, \ldots, \zeta_{p-1}^{(j)}$ necessarily exist and are smaller or equal to $\zeta_{p}^{(j)}$. In fact, if $\zeta_p^{(j)}$ exists, it means that at $r=\zeta_p^{(j)}$, $0$ is the $p$-th smallest eigenvalue of $H_r^{(j)}$: the $p-1$ smallest are thus $\leq0$. Given that these $p-1$ smallest are $\geq0$ at $r=1$ and by continuity of $s_p^{\uparrow}(H_r^{(j)})$, they necessarily cross zero before $\zeta_p^{(j)}$. Similarly, if $\zeta_{p\geq 2}^{(j)}$ does not exist, then all $\zeta_q^{(j)}$ for $q>p$ do not exist.
    \item Empirically, on as many connected graphs as we could think of, we have observed that the function $s_p^{\uparrow}(H_r)$ for $r\geq 1$ either never crosses zero (in which case $\zeta_p$ does not exist), crosses zero exactly twice and is convex between these two crossings (in which case $\zeta_p$ is the lowest of the two values), or, in very symmetric cases, touches zero exactly once without crossing it (in which case $\zeta_p$ is that value). 
\end{enumerate}

We are now in position to state our main result. 
\begin{claim}
	Let $\mathcal{G}(\mathcal{V},\mathcal{E})$ be generated according to Equation~\eqref{eq:DC-SBM}, that is, a DC-SBM with $k$ communities. Let $D$ and $A$ be its degree and adjacency matrices, and $H_{r} = (r^2-1)I_n + D - rA$, for $r \in \mathbb{R}$ its Bethe-Hessian matrix. 
    Provided that Assumption~\ref{ass:cpi} is satisfied, we have with high probability for large $n$ that:
	\begin{itemize}
	    \item There is only one connected component $j$ for which $\zeta_{p\geq 2}^{(j)}$ exists: it is the giant component. In the following, abusing notation, we simply write $\zeta_p$ instead of $\zeta_p^{(j)}$ to refer to the $\zeta$'s associated to this giant component. One has $\zeta_1=1$ and, if it exists, $\zeta_{p\geq 2}$ verifies\footnote{This statement allows to define $\zeta_p$ with respect to the Bethe-Hessian matrix of the whole graph, instead of the Bethe-Hessian matrix of its giant component. The fact that Eq.~\eqref{eq:zeta} is verified with high probability is not evident. In fact,  considering all the disconnected components at once might change the ordering of the eigenvalues. More details are to be found in Appendix~\ref{app:zeta_p}.}
	   \begin{align}
	   \label{eq:zeta}
	       \zeta_p = \underrel{r>1}{\rm min}\{r : s_p^{\uparrow}(H_r) = 0\}.
	   \end{align}
	    \item The largest $p$ for which $\zeta_p$ exists is equal to $k$, the number of underlying communities of the DC-SBM.\footnote{Precisely, we will see that $s_{k+1}^{\uparrow}(H_r) > 0$ for all $r>1$ with high probability, so that $\zeta_{k+1}$ is not defined; in fact it was interestingly shown in \citep{saade2014spectral} and we will verify that $s_{k+1}^{\uparrow}(H_{\sqrt{c\Phi}})\downarrow 0^+$ as $n\to\infty$, but the limit is never reached.} One has  $1=\zeta_1<\zeta_2\leq\ldots\leq\zeta_{k}\leq\sqrt{c\Phi}$. More precisely:
	\begin{align}
	   \forall~ p \text{ s.t. }\; 1\leq p\leq k, \quad \zeta_{p} = \frac{c}{\nu_p} + o_n(1).
	\label{eq:zeta_th} 
	\end{align} 
	\item For $2\leq p\leq k$ and $\zeta_p \leq r \leq \sqrt{c\Phi}$, the $p$ smallest eigenvalues of $H_r$ are isolated. In particular, zero is an isolated eigenvalue of $H_{\zeta_p}$. Its corresponding eigenvector $\bm{x}_p$ is correlated to the community structure and the entries of $\bm{x}_p$ are in expectation (over realizations of $A$) independent of the degrees of the graph.
	\end{itemize}
	\label{claim:1}
\end{claim}

In simple terms, Claim~\ref{claim:1} predicts that, in a graph of $k$ communities, spectral clustering can be successfully performed by successively retrieving the eigenvector associated to the null eigenvalue of each of the matrices $H_{\zeta_2},\ldots,H_{\zeta_k}$. Note that these eigenvectors have null entries for nodes that are not connected to the giant component. Thus, all the nodes not connected to the giant component will be arbitrarily classified: this is a structural consequence of sparsity, for which complete recovery is not feasible \citep{mossel2014belief}.

In the specific case of $k=2$ classes of equal size with $C_{\ell_i,\ell_j}=c_{\rm in}$ if $\ell_i=\ell_j$ and $c_{\rm out}$ if not, Claim \ref{claim:1} states that the $\zeta$'s exist only up to $p=k=2$, and that\footnote{One can indeed easily show that  $\nu_2=\frac{c_{\rm in}-c_{\rm out}}{2}$ in this case of two equal-sized classes} $\zeta_2 = (c_{\rm in } + c_{\rm out})/(c_{\rm in} - c_{\rm out})+o_n(1)$. Also, the relevant community information is found in the eigenvector $\bm{x}_2$ associated with the zero eigenvalue (the second smallest) of $H_r$ for $r=\zeta_2$. As we will see next, this value of $r$ differs from the choice $r=\sqrt{\rho(B)}$ advocated by \citep{saade2014spectral}, unless $\alpha=\frac{c_{\rm in } - c_{\rm out}}{\sqrt{c}}=\alpha_c=\frac{2}{\sqrt{\Phi}}$, i.e., exactly at the phase transition point. 

\begin{figure}
	\centering
	\includegraphics[width=0.8\columnwidth]{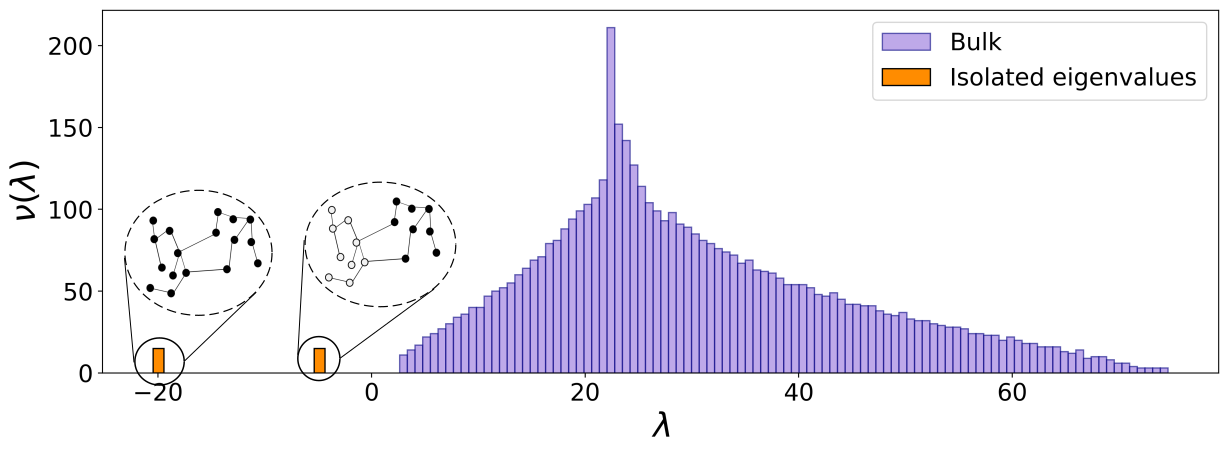}
	\caption{Spectrum of the matrix $H_r$ for $r = 3/2\sqrt{c\Phi}$. On the $x$ axis the eigenvalues, on the $y$ axis the respective histogram. In purple the bulk of uninformative eigenvalues and in orange the two isolated eigenvalues---zoomed to simplify the readability---. The two drawings give  a sketchy representation of the \emph{ferromagnetic}  ground state (all black spins) and of the informative state (half black and half white).	
		For this network, $n = 5000$, $k=2$, $\bm{\pi} = \bm{1}_k/2$, $c_{\rm in} = 8$, $c_{\rm out} = 3$, $r = 6.4$, $\theta_i \sim [\mathcal{U}(3,5)]^4$.}
	\label{fig:BH}
\end{figure}

Figure~\ref{fig:BH} provides a visual representation of the typical spectrum of $H_r$ for $r$ not too far from a $\zeta_p$: the eigenvalue of $H_r$ closest to zero is clearly isolated and the eigenvector associated to this eigenvalue is strongly aligned to the community structure.
In Figure~\ref{fig:zeta}, for a typical realization of a DC-SBM, $s_p^{\uparrow}(H_r)$ is represented as a function of $r$ for $p=1, 2, 3$ and $4$ in solid lines and $p=5$ in dotted line. In this instance, $\zeta_5$ does not exist (and all subsequent $\zeta_p$'s do not exist either), $\zeta_1=1$, and $\zeta_2$, $\zeta_3$ and $\zeta_4$ exist and lie between $1$ and $2$.

\medskip

The non-obvious parts of the claim are (i) of course that such $\zeta_p$'s do exist up to $p=k$, (ii) that they are concentrated around a deterministic value depending on the statistics of the model, (iii) importantly, that zero is indeed an isolated eigenvalue in the spectrum of $H_{\zeta_p}$, (iv) that the associated eigenvector is informative for the underlying community structure and that is not infected by the degrees of the graph.

\medskip

The statement of Claim~\ref{claim:1} is formulated as a claim in the sense that, while efforts have been made to rigorously prove parts of the result \citep[see, \emph{e.g.},][]{ coste2019eigenvalues}, the mathematical tools required to fully prove Claim~\ref{claim:1}, to the best of our knowledge, do not exist yet. Instead, the remainder of this section will propose three complementary supporting elements, arising from non-rigorous but convincing approximations, in particular borrowing arguments from the field of statistical physics.

\subsection{Supporting Arguments to Claim~\ref{claim:1}}

This section details our heuristic arguments in support of Claim~\ref{claim:1}. Specifically, we will successively show
\begin{itemize}
	\item under Section~\ref{subsec:connection.BP} that the vectors $\bm{x}_p$, solution of $H_{\zeta_p}\bm{x}_p = 0$, are informative in the sense that they are correlated with the community labels.
	\item under Section~\ref{subsec:connection.ising}, that the informative null eigenvalue of $H_{\zeta_p}$ (associated to the informative eigenvector $\bm{x}_p$) is located in $p$-th smallest position and is isolated. This result is algorithmically crucial to determine $\zeta_p$ itself.
	\item under Section~\ref{subsec:resilience}, that the entry $x_{p,i}$ is essentially independent of $d_i$, the degree of node $i$; precisely, it is strictly independent of $d_i$ on average (over random allocation of the labels) and only depends on $d_i$ through a noise term of order $1/\sqrt{d_i}$ otherwise.
\end{itemize}

\subsubsection{Linearization of the Belief Propagation Equations}
\label{subsec:connection.BP}

We first proceed to our agenda by arguing that the eigenvectors $\bm{x}_p$ of $H_{\zeta_p}$ are correlated with the community labels.

\medskip

A first approach to the question of community detection in sparse graphs consists in estimating the community allocation probabilities $\mathbb{P}(\ell_i|A)$ using belief propagation (BP) approximation, based on the local-tree approximation of sparse graphs. The BP node marginal is given by \citep[Equation~28]{decelle2011asymptotic}, which we recall here:
	\begin{align}
   \mathbb{P}(\ell_i|A) &\approx \frac{\pi_{\ell_i}}{Z_i}e^{-h_{\ell_i}} \prod_{j \in \partial i}\sum_{\ell_j} C_{\ell_i,\ell_j} \eta_{ij}(\ell_j),
   \label{eq:marginal}
\end{align}
in which $Z_i$ is the probability normalization constant, and the $\eta_{ji}(\ell_i)$ (the ``messages'') and $h(\ell_i)$ are the solutions to  \citep[see][Equations~26,27]{decelle2011asymptotic}:
\begin{align}
\eta_{ji}(\ell_i) &= \frac{1}{Z_{ji}} \pi_{\ell_i}e^{-h_{\ell_i}}\prod_{m \in \partial i \setminus j}\sum_{\ell_m} C_{\ell_m,\ell_i}\eta_{im}(\ell_m) \label{eq:fixed_BP} \\
h_{\ell_i} &= \frac{1}{n}\sum_{k \in \mathcal{V}} \sum_{\ell_k} C_{\ell_i,\ell_k}\mathbb{P}(\ell_k|A) \label{eq:field}.
\end{align}
Equation~\eqref{eq:fixed_BP} admits the trivial $\eta_{ji}(\ell_i) = \pi_{\ell_i}$ which corresponds to the so-called \emph{paramagnetic fixed point} \citep{zhang2015nonbacktracking}. Let us perform a linear expansion of Equation~\eqref{eq:fixed_BP} around this trivial fixed point by denoting $\eta_{ji}(\ell_i) = \pi_{\ell_i} + b_{ji}(\ell_i)$. Letting $\bm{b}^T = [\bm{b}_1^T,\dots,\bm{b}_k^T] \in \mathbb{R}^{2|\mathcal{E}|k}$, with $\bm{b}_p \in \mathbb{R}^{2|\mathcal{E}|}$ containing the entries of $b_{ij}(p)$, we obtain as in \citep{krzakala2013spectral}: 
\begin{align}
    (T \otimes B)\bm{b} = \bm{b}+o_n(1)
    \label{eq:linear_B}
\end{align}
where $\otimes$ is the Kronecker product, $T = \frac{\Pi C}{c}-\Pi\bm{1}_k\bm{1}_k^T$ and $B \in \{0,1\}^{2|\mathcal{E}|\times2|\mathcal{E}|}$ is the non-backtracking matrix, which we recall is defined on the set of directed edges $\mathcal{E}^d$ of $\mathcal{G}$ as:
\begin{align}
    B_{(ij)(kl)}=\delta_{jk}(1-\delta_{il}),~\forall~(ij),(kl)~\in~\mathcal{E}^d,
    \label{eq:B}
\end{align}
in which $\delta$ is the Kronecker symbol. Before proceeding, we recall an important property of the spectrum of $B$ rigorously established in \citep{bordenave2015non,gulikers2016non} and which we will extensively use in the following. Under Assumption~\ref{ass:cpi}, with high probability, the matrix $B$ has $k$ isolated real eigenvalues equal to\footnote{Recall that $\nu_p$ is the $p$-th largest eigenvalue of $C\Pi$.} $\nu_p\Phi+o_n(1)$, $1 \leq p \leq k$, while all other (complex) eigenvalues are contained within a disk of the complex plane of radius $\sqrt{\rho(B)} = \sqrt{c\Phi}+o_n(1)$.

\medskip

From Equation \eqref{eq:linear_B}, our interest is in the eigenvalues of $(T\otimes B)$ equal (or close) to one. From the properties of the Kronecker product, the set of eigenvalues of $(T\otimes B)$ is $S(T\otimes B) = \{s_i(B)s_j(T)~:~1 \leq i \leq 2|\mathcal{E}|, 1\leq j \leq k\}$. This induces a one-to-one relation between the $k-1$ non-zero eigenvalues\footnote{Note that $s_1^{\uparrow}(T) = 0$.} of $T$ and the $k-1$ eigenvalues of $B$, that need to satisfy the relation $s_q(T)s_t(B)=1$ for some $t,q$. From the expression of $T$, it comes that the $p$-th largest eigenvalue of $T$ satisfies $s_p^{\downarrow}(T) = \frac{\nu_{p+1}}{c}$ for $1 \leq p \leq k-1$ (see Appendix~\ref{app:T} for details) so that, for $2 \leq p \leq k$, there must exist $\zeta_p\in S(B)$ such that:

\begin{figure}[t!]
	\centering
	\includegraphics[width=\columnwidth]{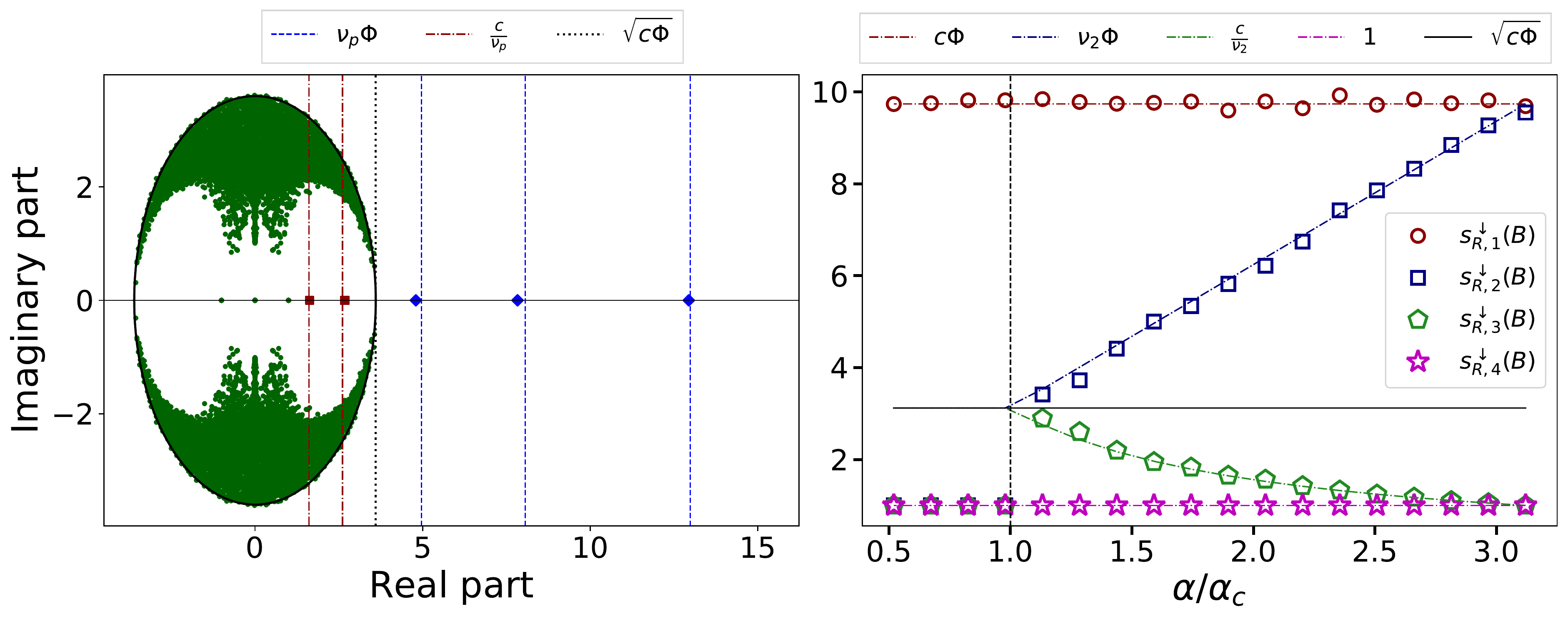}
	\caption{(Left): spectrum of the matrix $B$ on the complex plane. The green dots represent the uninformative eigenvalues, the blue diamonds the real eigenvalues outside the bulk and the red squares the real eigenvalues inside the bulk. The lines indicate the theoretical values. For this network $n=5000$, $k = 3$, $\theta_i \sim[\mathcal{U}(3,10)]^3$, $c = 8$. The off-diagonal elements of $C$ are distributed according to $C_{p>q} \sim \mathcal{N}(c_{\rm out}, c_{\rm out}/k)$ with  $c_{\rm out} = 4$ and the diagonal elements are then fixed to have $C\Pi\bm{1}_k = c\bm{1}_k$. The element of  $\bm{\pi}$ are distributed as $\pi_i \sim  \mathcal{N}(1/k,1/2k)$. The entries are then rescaled sot that ${\rm Tr}(\Pi) = 1$. (Right): position of the first four real eigenvalues of $B$ as a function of $\alpha$. For this simulation $n = 5000$, $k=2$, $\theta_i = [\mathcal{U}(3,10)]^3$, $c_{\rm in}: 12 \to 7$, $c_{\rm out}: 0 \to 5$ with $c = 6$.}
	\label{fig:B}
\end{figure}

\begin{align}
    \zeta_p = \frac{c}{\nu_p} + o_n(1).
    \label{eq:def1_zeta}
\end{align}
To these eigenvalues correspond $k-1$ \emph{exact} eigenvectors $\bm{g}_p \in\mathbb{R}^{2|\mathcal{E}|}$ of $B$ satisfying $B\bm{g}_p = \zeta_p\bm{g}_p$ (with $\bm{g}_p = \bm{b}_p + o_n(1)$) that are naturally ``informative'' as they are associated to structural eigenvalues of $T$. Also, from a belief propagation standpoint, these eigenvectors are small deviations from the uninformative fixed point, so must point towards informative directions.

Note importantly here that Equation~\eqref{eq:def1_zeta} \emph{defines} $\zeta_p$ as a real eigenvalue of the matrix $B$. So far, this definition needs not correspond to the claimed definition of $\zeta_p$ as introduced in Claim~\ref{claim:1}: Section~\ref{subsec:connection.ising} will show that these two definitions are indeed equivalent.
In addition, by Assumption~\ref{ass:cpi} and the Perron-Frobenius theorem, $\nu_1 = c$ has unit multiplicity,
so that $\zeta_p > 1+o_n(1)$ for all $p > 1$. Along with Equation~\eqref{eq:def1_zeta}, we thus have $1+o_n(1) < \zeta_p \leq \sqrt{c\Phi} + o_n(1)$, and thus $\zeta_p$ is asymptotically confined inside the bulk of radius $\sqrt{\rho(B)}=\sqrt{c\Phi}+o_n(1)$ of $B$. Figure~\ref{fig:B} confirms, in agreement with the formal result of \citep{coste2019eigenvalues} obtained in the slightly non sparse regime ${\rm log}(n)/c = o_n(1)$, that inside the disk of radius $\sqrt{\rho(B)}$, these are the only informative eigenvalues of $B$. Specifically, the real eigenvalues of $B$ inside the bulk are divided between (i) the (non-informative) eigenvalues $-1,0,1$, and (ii) the $k$ eigenvalues $\gamma = \zeta_p \approx \frac{c}{\nu_p} \leq \sqrt{c\Phi}$ just described.

\medskip 

Having identified the $k-1$ informative eigenvectors $\bm{g}_{2 \leq p \leq k} \in \mathbb{R}^{2|\mathcal{E}|}$ of $B$, we now consider how to process $\bm{g}_p$ in order to obtain a vector of size $n$ whose entries are in one to one mapping with the nodes of the graph. From the linearization of the marginal probability distribution of the labels (Equation~\eqref{eq:marginal}), the term $\sum_{j \in \partial i} b_{ij}(\ell_i) \approx \sum_{j \in \partial i}g_{\ell_i,ij} \equiv x_{\ell_i,i}$ brings information on the label of node $i$, hence on the class structure. 
Now, let $\bm{g}$ be the solution of $B\bm{g} = \gamma \bm{g}$ for some $\gamma$ and $g^{\rm in}_i = \sum_{j \in \partial i} g_{ij}$. Then, thanks to the Ihara-Bass formula \citep{krzakala2013spectral,terras2010zeta}, we find that
\begin{align}
&[(\gamma^2-1)I_n + D - \gamma A]\bm{g}^{\rm in} = 0.
\label{eq:Hashimoto}
\end{align}
The matrix in brackets appears to be the Bethe-Hessian matrix $H_{r}=(r^2-1)I_n+D-rA$, here for $r = \gamma$. This provides an explicit link between $B$ and $H_r$. According to our earlier discussion, the informative eigenvectors of the matrix $B$ have corresponding eigenvalues $\gamma = \zeta_p$ and there are consequently $k-1$ informative eigenvectors $\bm{x}_p$ for $2 \leq p \leq k$ such that
\begin{eqnarray}
H_{\zeta_p}\bm{x}_p = 0.
\end{eqnarray}
Note that, for $\zeta_1 = 1$ we have $H_1 = D-A$ and $\bm{x}_1$ is the vector $\bm{1}_n$, which is irrelevant to reconstruct communities. 

\medskip

Let us specifically focus on the particular case of two classes of equal size and consider the question of the detectability threshold. In this case the matrix $C\Pi$ has a unique non-trivial eigenvalue equal to $\nu_2 = (c_{\rm in} - c_{\rm out})/2$. 
In this setting, $B$ has two informative eigenvalues $\approx \nu_2\Phi$ and $\approx \frac{c}{\nu_2}$ on either side of the disk (bulk) of radius $\sqrt{c\Phi}$. As the detection problem becomes harder, the two eigenvalues get closer together until they (asymptotically) meet right at the detectability transition where $\nu_2\Phi=\sqrt{c\Phi}$ (and $\alpha=\alpha_c$). Further increasing the detection difficulty (that is, for $\alpha < \alpha_c$), the two eigenvalues now become complex, each being the complex conjugate of the other. This behavior is shown in Figure \ref{fig:B} (right panel). Summarizing, we have
\begin{align*}
\nu_2\Phi = \frac{c}{\nu_2}= \sqrt{c\Phi}, \quad  &{\rm when} ~\alpha = \alpha_c\\
\nu_2\Phi >  \sqrt{c\Phi}>\frac{c}{\nu_2}> 1, \quad  &{\rm when} ~\alpha > \alpha_c. \label{eq:inequality_B}
\end{align*}

The fact that the second eigenvalue of largest amplitude $\approx \nu_2\Phi$ of $B$ remains isolated down to the \emph{detectability threshold} is the strongest argument in favor of the algorithm proposed by \citep{krzakala2013spectral}. The authors in \citep{krzakala2013spectral} however ignored the effect on the corresponding eigenvalue $\approx \frac{c}{\nu_2}$ which, from our present discussion, is similar.
This behavior can be extended to more than two classes, obtaining $\zeta_p \approx \frac{c}{\nu_p}\leq \nu_p\Phi$ with equality at the transition point where $\nu_p\Phi = \sqrt{c\Phi}$. This can be summarized as follows:
\begin{argument}
\label{el:1}
	Under Assumption~\ref{ass:cpi} for all large $n$ with high probability, the non-backtracking matrix $B$ of a graph $\mathcal{G}(\mathcal{V},\mathcal{E})$ generated from a $k$-class DC-SBM has $k-1$ isolated real eigenvalues \emph{inside} the disk of radius $\sqrt{c\Phi}$ (bulk). These eigenvalues are found at positions $1 < \zeta_p = \frac{c}{\nu_p} + o_n(1)$ for $2\leq p\leq k$. Besides, the eigenvectors $\bm{g}_p$ of $B$ associated with these eigenvalues $\zeta_p$ bring information about the community structure, which can be extracted through the vectors $\bm{x}_p \in \mathbb{R}^n$, defined as:
	\begin{equation*}
	x_{p,i} = \sum_{j \in \partial i} g_{p,ij}, \quad B\bm{g}_p = \zeta_p\bm{g}_p ; 
	\end{equation*}
	or equivalently satisfying
	\begin{equation*}
	H_{\zeta_p}\bm{x}_p = 0.
	\end{equation*}
\end{argument}

Element \ref{el:1} provides a first statement of Claim \ref{claim:1}, according to which the class information should be retrieved from the vector $\bm{x}_p$ solution to $H_{\zeta_p}\bm{x}_p = 0$, and that $\zeta_p = c/\nu_p + o_n(1)$. This argument however does not specify the location (in the ordered list of the eigenvalues) of the null eigenvalue of $H_{\zeta_p}$ to which the eigenvector $\bm{x}_p$ corresponds nor the structure of the eigenvector $\bm{x}_p$, and in particular its dependence on the degrees of the graph.

The subsequent sections will cover these aspects.

\subsubsection{The Excited States of the Ising Hamiltonian on $\mathcal{G}$}
\label{subsec:connection.ising}

In this section, through a statistical physics mapping between the graph $\mathcal G$ and a system of interacting \emph{spins}, we aim to informally justify why the smallest eigenvalues of the matrix $H_r$ correspond to ``informative states'' of the system and why the zero eigenvalue of $H_{\zeta_p}$ (associated by definition with the informative eigenvectors $\bm{x}_p$) should be isolated and in the $p$-th smallest position of the spectrum of $H_{\zeta_p}$.

\subsubsubsection{~The Smallest Eigenvalues of $H_r$ are Informative}

Consider the Ising model of interacting \emph{spins} $\sigma_i \in \{\pm 1\}$ on the graph $\mathcal{G}(\mathcal{V},\mathcal{E})$, in which the intensity of spin coupling is controlled by a parameter $r$. Its Hamiltonian $\mathcal H({\bm\sigma})$ expresses the energy of a given configuration ${\bm\sigma}$ as \citep[see \emph{e.g.}][]{mezard2009information}:
\begin{align}
\mathcal{H}(\bm{\sigma}) = -\sum_{(ij)\in\mathcal{E}} {\rm ath}\left(\frac{1}{r}\right)\sigma_i\sigma_j.
\label{eq:Ising_G}
\end{align}
The parameter ${\rm ath}(1/r)$ is related to the temperature of the system and it appears through an inverse hyperbolic tangent for computational ease. 
The vector $\bm{\sigma}$ is a random variable, distributed according to the Boltzmann distribution $\mu(\bm{\sigma})$: 
\begin{equation}
\mu(\bm{\sigma}) = \frac{1}{Z}e^{-\mathcal{H}(\bm{\sigma})}.
\label{eq:Boltzmann_weight}
\end{equation}
Equation \eqref{eq:Boltzmann_weight} implies that the configurations with a low energetic cost are realized with high probability.
Since $\bm{\sigma}$ is a random variable, to properly characterize its \emph{stable average configuration}, we will focus our study on its statistical mean. The mean $\bm{m^*}=\langle \bm{\sigma} \rangle$, as well as the covariance $\bm{\chi^*}=\{\langle \sigma_i\sigma_j \rangle\}_{i,j}$ and all subsequent moments of $\mu$ (here $\langle \cdot \rangle$ denotes an expectation taken over the distribution \eqref{eq:Boltzmann_weight}), can be obtained from the explicit expression of $Z$ which, seen as a function of the temperature, is a moment generating function. But an explicit expression of $Z$ cannot be obtained in general. A common approximation, adapted to sparse systems, is the \emph{Bethe} approximation \citep{bethe1935statistical} $Z_{\rm Bethe}(\bm{m},\bm{\chi})\approx Z$, which is asymptotically exact as $n\to\infty$ (see Appendix~\ref{app:mapping_to_Ising} for details).
The function $F_{\rm Bethe}(\bm{m},\bm{\chi}) = -\log Z_{\rm Bethe}(\bm{m},\bm{\chi})$ is called the \emph{Bethe free energy} and is a variational approximation of the actual free energy $F = -\log Z = \langle \mathcal{H}(\bm{\sigma})\rangle - S[\mu(\bm{\sigma})]$, where $S[\cdot]$ denotes the entropy of a distribution. One can show (see Appendix \ref{app:mapping_to_Ising}) that $F_{\rm Bethe}(\bm{m},\bm{\chi}) \geq F$, so the best estimate of the \emph{stable configuration} $(\bm{m}^*,\bm{\chi}^*)$ is
\begin{equation}
(\bm{m}^*, \bm{\chi}^*) \approx {\rm arg}~\underset{\bm{m},\bm{\chi}}{\rm min}\big( - {\rm log}~Z_{\rm Bethe}(\bm{m},\bm{\chi})\big) \equiv {\rm arg}~\underset{\bm{m},\bm{\chi}}{\rm min}~F_{\rm Bethe}(\bm{m},\bm{\chi}).
\label{eq:stable}
\end{equation}
For $r$ sufficiently large (that is, at high temperature) the free energy favors disordered configurations with high entropy. In this case, the function $F_{\rm Bethe}(\bm{m},\bm{\chi})$ has a unique minimum at $\bm{m}^* = 0$ and the system is said to be in the \emph{paramagnetic phase}. For smaller (more interesting) values of $r$, the free energy tends to favor configurations with a small energetic cost (minimizing Equation~\eqref{eq:Ising_G}). In this case the minima satisfy $\bm{m}^* \neq 0$ and $\bm{m} = 0$ becomes a saddle point \citep{leone2002ferromagnetic}: the stable average configuration of the spins has a non trivial ordering. In Appendix~\ref{app:mapping_to_Ising} we show that, on a graph $\mathcal{G}$ with $k$ communities, there exist exactly $k$ directions from the point $\bm{m} = 0$ along which the free energy \emph{can} exhibit a local minimum. The limitation to $k$ directions is a direct consequence of the fact that ${\rm rank}(\mathbb{E}[A]) = k$. These directions are those along which a non-trivial organization of the spins becomes stable at sufficiently low temperature: they are naturally correlated to the underlying community structure of $\mathcal{G}$.

\medskip

In order to formally carry out the stability analysis (i.e., to identify the above non-trivial directions of free energy descent), one needs to study the eigenvalues and eigenvectors of the Hessian matrix $\nabla^2_{\bm{m}} F_{\rm Bethe}$ of the function $F_{\rm Bethe}$ at the paramagnetic point $\bm{m} = 0$. In \citep{watanabe2009graph,saade2014spectral}, the authors show that this Hessian matrix at $\bm{m}=0$ is strictly proportional to the Bethe-Hessian, $H_r$. This induces a natural link to $H_r$, which in the previous section was defined from the non-backtracking matrix $B$ (arising from a linearization of the belief propagation equations).
The eigenvectors associated to the negative eigenvalues of the Hessian $H_r$ precisely correspond to the sought-for directions towards the local (or global) minima of the Bethe free energy. According to our earlier discussion, only $k$ such directions may exist, so that only up to $k$ eigenvalues (the $k$ smallest) of $H_r$ can be negative (i.e., in the limit of large $n$, $s_{k+1}^{\uparrow}(H_r) \geq 0, \forall~r > 1$), and their corresponding eigenvectors are \emph{all} correlated to the class labels. 

\medskip

We are thus left to showing why specifically the $p$-th smallest eigenvalue of $H_r$ for the particular choice of $r=\zeta_p$ is of utmost importance, and why it corresponds to the null isolated eigenvalue of $H_{\zeta_p}$.

\begin{figure}[t!]
	\centering
	\includegraphics[width = \columnwidth]{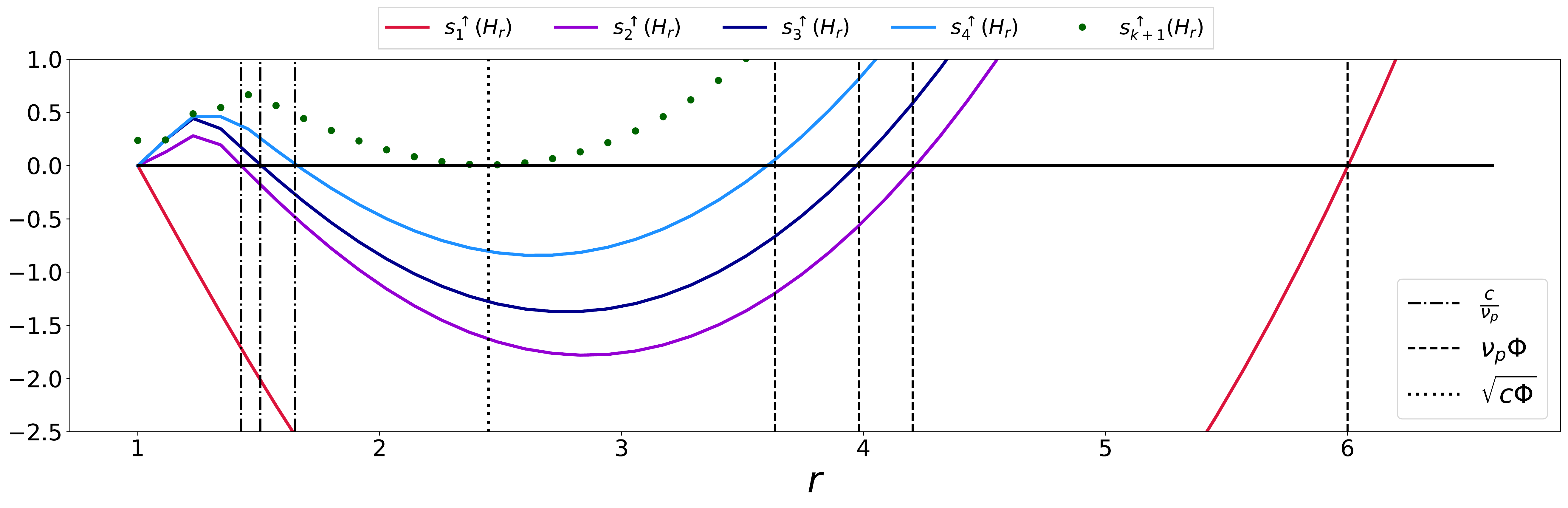}
	\caption{\vspace{-1cm}Behavior of $s_p^{\uparrow}(H_r)$ as a function of $r$ for $1 \leq p \leq k$ in solid lines, and for $p = k+1$ in dotted line. The vertical lines are the theoretical values of $c/\nu_p$(dashed dotted line), of $\sqrt{\rho(B)}$ (dotted line) and of $\nu_p\Phi$ (dashed lines). For this simulation, $n = 50.000$, $c = 6$, $c_{\rm out} = 2$, $k = 4$, $C_{p>q}\sim\mathcal{N}(c_{\rm out},2c_{\rm out}/k)$ and $\pi \propto \bm{1}_k$.\vspace{-0.5cm}}
	\label{fig:zeta}
\end{figure}

\subsubsubsection{~Zero is an Isolated Eigenvalue of $H_{\zeta_p}$}

This result follows from the observation, reported in Figure~\ref{fig:zeta}, that the $k$-th smallest eigenvalues $s_1^{\uparrow}(H_r),\ldots,s_k^\uparrow(H_r)$ of $H_r$ successively equal zero in this order as $r$ increases. Starting from $r = 1$, $H_r=H_1 = D-A$ for which we know that $s_j^{\uparrow}(H_1) = 0$ for $1 \leq j \leq n_{\rm CC}$ with $n_{\rm CC}$ the number of connected components and $s_i^{\uparrow}(H_1) \geq 0$, $i > n_{\rm CC}$. Increasing $r$ beyond $r=1$, the successive smallest eigenvalues of $H_r$ first all increase and remain equal to the left edge of the bulk before escaping, each in turn (and in the order $s_2^{\uparrow}(H_r),\ldots,s_k^\uparrow(H_r)$), the bulk of non-informative eigenvalues. At their point of escape, they successively shift until they cross zero: this is where the successive values $\zeta_2\leq ... \leq \zeta_k$ are defined. This in particular implies that the $p$-th smallest eigenvalue of $H_{\zeta_p}$ coincides with (i) the null eigenvalue of $H_{\zeta_p}$, as well as with (ii) its largest isolated (so the last informative) eigenvalue. This allows us to redefine $\zeta_p$ as in Equation~\eqref{eq:zeta} of Claim~\ref{claim:1}. 

Note in passing that, letting $r$ further increase beyond $\zeta_k$, the left edge of the non-informative bulk of $H_r$ progressively shifts back (from positive values) towards zero until it reaches asymptotically zero in the limit where $r=\sqrt{\rho(B)}$ \citep{saade2014spectral}, before increasing again away for $r>\sqrt{\rho(B)}$.
These last findings may be summarized as follows.
\begin{argument}
\label{el:2}
	Given a graph $\mathcal{G}(\mathcal{V},\mathcal{E})$ with $k$ classes, generated from the DC-SBM, the $p$ smallest eigenvalues of the matrix $H_r$ are isolated for $\zeta_p \leq r \leq \sqrt{c\Phi}$, for all $2 \leq p \leq k$. The entries of the $p$ smallest eigenvectors are correlated with the class labels. Besides, in the specific case where $r = \zeta_p$, the $p$-th smallest eigenvalue of $H_{\zeta_p}$ is equal to zero.
\end{argument}

Element~\ref{el:2} answers the problem of locating the informative eigenvalue-eigenvector pair $(0,\bm{x}_p)$ of $H_{\zeta_p}$ introduced in Element~\ref{el:1}, and of justifying why it is isolated. The last step, developed in the next section, consists in better understanding the structural content of the informative eigenvector $\bm{x}_p$.

\subsubsection{Parametrization to Provide Resilience to Degree Heterogeneity}
\label{subsec:resilience}

From a purely algebraic standpoint, the Bethe-Hessian matrix $H_r$ may be seen as a regularized combinatorial Laplacian. In \citep{dall2019revisiting} we studied the problem of a good parametrization $r$ of $H_r$ for which the informative eigenvectors are resilient to degree heterogeneity. We briefly report here the main conclusions, based on a local approximation of the neighborhood of each node. See \citep[Section~2]{dall2019revisiting} for further details.

The argument goes as follows: exploiting sparsity, the graph $\mathcal{G}$ can be shown to locally converge to a Galton-Watson tree in which the offsprings are statistically independent \citep{dembo2010gibbs,mossel2015reconstruction,salez2011some,mossel2014belief, gulikers2015impossibility}. 

Fixing $A$ (and thus the degrees $d_i$), we now perform a Bayesian analysis \emph{on a random allocation of the class labels}. Considering node $j$, labeled as $\ell_j$, as the \emph{root} of the tree, the probability for offspring $i$ to have label $\ell_i$ reads
\begin{align}
\mathbb{P}(\ell_i|\ell_j,A_{ij}=1) &= \frac{\mathbb{P}(\ell_i,\ell_j|A_{ij}=1)}{\mathbb{P}(\ell_j|A_{ij}=1)}
=\frac{\iint d\theta_i d\theta_j \mathbb{P}(\ell_i,\ell_j,\theta_i\theta_j|A_{ij}=1)}{\mathbb{P}(\ell_j)} \nonumber\\
&= \frac{\iint d\theta_id\theta_j 	\mathbb{P}(A_{ij}=1|\theta_i,\theta_j,\ell_i,\ell_j)\mathbb{P}(\ell_i)\mathbb{P}(\ell_j)\mathbb{P}(\theta_i)\mathbb{P}(\theta_j)}{Z \pi_{\ell_j}}\nonumber\\
&= \frac{\pi_{\ell_i}C_{\ell_i,\ell_j}}{c} =  \frac{(\Pi C)_{\ell_i,\ell_j}}{c} =  \frac{(C \Pi)_{\ell_j,\ell_i}}{c}.
\label{eq:GW}
\end{align}

Recalling the definition of the eigenvalue-eigenvector pairs $(\nu_p,\bm{v}_p)$ of $C\Pi$ (i.e., $C\Pi\bm{v}_p = \nu_p \bm{v}_p$), let us define $\bm{u}_p\in\mathbb{R}^n$ with $u_{p,i} = v_{p,\ell_i}$ the $n$-dimensional class-wise expansion of $\bm{v}_p$: this vector $\bm{u}_p$ is inherently random as the class allocations $\ell_1,\ldots,\ell_n$ are here considered random. Then, for $n$ large, under the limiting tree approximation with conditionally independent offsprings, we take an expectation over the random allocation of the labels different from $\ell_i$ $(\bm{\ell}_{\setminus i})$, with $A$ and $\ell_i$ known:
\begin{align}
\mathbb{E}_{\bm{\ell}_{\setminus i}}[(A\bm{u}_p)_i|A, \ell_i] &= \sum_{j \in \partial i} \mathbb{E}_{\bm{\ell}_{\setminus i}}[ u_{p,j}|A,\ell_i] = \sum_{j \in \partial i} \mathbb{E}_{\bm{\ell}_{\setminus i}}[ v_{p,\ell_j}|A,\ell_i]  
\approx \sum_{j \in \partial i} \sum_{\ell_j} \mathbb{P}(\ell_j|\ell_i,A_{ij}=1) v_{p,\ell_j}  \nonumber \\
&= \frac{1}{c} \sum_{j \in \partial i}\sum_{\ell_j}(C\Pi)_{\ell_i,\ell_j}v_{p,\ell_j} 
= \frac{1}{c}\sum_{j \in \partial i} (C\Pi v_p)_{\ell_i} = \frac{d_i}{c} \nu_{p} v_{p,\ell_i} = d_i \frac{\nu_{p}}{c} u_{p,i}
\label{eq:expectation}
\end{align}
where the approximation follows from the fact that conditional independence of the neighbors of a given node only holds asymptotically; in particular, the approximation becomes an equality on a tree rooted at node $i$.
As a consequence of Equation~\eqref{eq:expectation}, using $H_r=(r^2-1)+D-rA$, we find that
\begin{equation}
\mathbb{E}_{\bm{\ell}_{\setminus i}}[H_r\bm{u}_{p}|A,\ell_i] \approx \left[(r^2-1)I_n +D\left(1-r\frac{\nu_p}{c}\right)\right]\bm{u}_{p}.
\label{eq:exp_on_u}
\end{equation}
In order for this equation to be an approximate eigenvector equation for arbitrary degrees in $D$, the right hand-side term proportional to $D$ must vanish. That is, one must select
\begin{equation*}
r = \frac{c}{\nu_p} \approx \zeta_p
\end{equation*}
(this last approximation having been introduced and discussed in Section~\ref{subsec:connection.BP}).
This result implies that the eigenvectors $\bm{x}_p$ defined as $H_{\zeta_p}\bm{x}_p=0$, for $1\leq p\leq k$, correspond to a noisy version of $\bm{u}_p$ which is not affected \emph{on average over the class allocation} by the degree distribution. In \citep{dall2019revisiting}, we confirmed ---so long that the average node degree is not too small--- that the approximation holds, beyond the average, for every typical realization of the class allocation. In a nutshell, this behavior unfolds from the following remark: denoting $\bm{\xi}_p$ the ``noise'' vector satisfying $\bm{x}_p = \bm{u}_p + \bm{\xi}_p$, and thus $H_{\zeta_p}(\bm{u}_p + \bm{\xi}_p) = 0$,
\begin{align}
    {\rm Var}_{\bm{\ell}_{\setminus i}}[(H_{\zeta_p}\bm{u}_p)_i|A,\ell_i] &= {\rm Var}_{\bm{\ell}_{\setminus i}}[(H_{\zeta_p}\bm{\xi}_p)_i|A,\ell_i] \nonumber \\
    &= {\rm Var}_{\bm{\ell}_{\setminus i}}[ (\zeta_p^2-1 + d_i)\xi_{p,i}-\zeta_p (A\bm{\xi}_{p,i})|A,\ell_i] \nonumber \\
    &\approx O_{d_i}(d_i^2) {\rm Var}_{\bm{\ell}_{\setminus i}}[\xi_{p,i}|A,\ell_i] + O_{d_i}(d_i) {\rm Var}_{\bm{\ell}_{\setminus i}}[\xi_{p,i}|A,\ell_i] \nonumber\\
    &\approx O_{d_i}(d_i^2) {\rm Var}_{\bm{\ell}_{\setminus i}}[\xi_{p,i}|A,\ell_i]
    \label{eq:24}
\end{align}
where we recall that the underlying random variable is the random class allocation. There we used the fact that the random variables $\xi_{p,i}$ and $[A\bm{\xi}_p]_i$ are essentially independent and that the variance of the sum of the $d_i$ asymptotically independent variables $\xi_{p,1},\ldots,\xi_{p,n}$ grows linearly with $d_i$ (and thus becomes essentially negligible). Now, proceeding as in Equation~\eqref{eq:expectation}, we can compute ${\rm Var}_{\bm{\ell}_{\setminus i}}[(H_{\zeta_p}\bm{u}_p)_i|A,\ell_i]$ from a direct calculation of $\mathbb{E}_{\bm{\ell}_{\setminus i}}[(H_{\zeta_p}\bm{u}_p)_i^2|A,\ell_i]$ and $\mathbb{E}_{\bm{\ell}_{\setminus i}}^2[(H_{\zeta_p}\bm{u}_p)_i|A,\ell_i]$, and we obtain ${\rm Var}_{\bm{\ell}_{\setminus i}}[(H_{\zeta_p}\bm{u}_p)_i|A,\ell_i]=O_{d_i}(d_i)$.

Combining with Equation~\eqref{eq:24}, we get that, for $d_i$ sufficiently large, ${\rm Var}_{\bm{\ell}_{\setminus i}}[\xi_{p,i}|A,\ell_i] = O_{d_i}(d_i^{-1})$.
The vector $\bm{x}_p$ can therefore be written as the sum of the deterministic information and a noise with amplitude inversely proportional to the square root of the degree, consistently predicting that nodes with higher degrees are easier to classify.

\medskip 

These results are summarized under the form of our last argument.
\begin{argument}
\label{el:3}
	The eigenvector $\bm{x}_p$ ($1\leq p\leq k$), solution to $H_{\zeta_p}\bm{x}_p=0$, is a noisy version of the vector $\bm{u}_p$, defined as $u_{p,i} = v_{p,\ell_i}$, where $C\Pi\bm{v}_p = \nu_p\bm{v}_p$, where the noise for entry $i$ scales as $1/\sqrt{d_i}$ for $d_i$ sufficiently large and is zero on average. Consequently, the entries of $\bm{x}_p$ do not, to first order, depend on the degree distribution but only on the labels. 
\end{argument}

Element~\ref{el:3} combined with Elements~\ref{el:1},~\ref{el:2} completes our justification of Claim~\ref{claim:1}.

\subsection{Comments and Performance Comparison}

Let us now discuss how Claim~\ref{claim:1} can be exploited in practice to obtain an efficient spectral clustering algorithm for sparse graphs with a heterogeneous degree distribution.

From Claim~\ref{claim:1} and our detailed analysis in Sections~\ref{subsec:connection.BP}--\ref{subsec:resilience}, 
for all large $n$ with high probability, under Assumption~\ref{ass:cpi}, the $k-1$ Bethe-Hessian matrices $H_{\zeta_2},\ldots,H_{\zeta_k}$ have an eigenvalue equal to zero, which is isolated. The corresponding eigenvectors $\bm{x}_2,\ldots,\bm{x}_k$ (which are \emph{not} necessarily orthogonal as they correspond to distinct matrices) are all informative in the sense that they are noisy realizations of piece-wise constant versions of the eigenvectors $\bm{v}_2,\ldots,\bm{v}_k$ of the matrix $C\Pi$ (each ``piece'' identifying each class). Besides, $\bm{x}_2,\ldots,\bm{x}_k$ are, to first order, resilient to the degree distribution in the network.

\medskip

This so far assumes that the number of classes $k$ is known, and that all classes satisfy the separability condition of Assumption~\ref{ass:cpi}. Yet, in practice, $k$ is generally unknown. The following remark provides a consistent estimator for $k$, as proposed in \citep{krzakala2013spectral, saade2014spectral}.
\begin{rem}[Estimation of the number of classes]
	Under Assumption~\ref{ass:cpi}, for all large $n$ with high probability, $\hat{k}_{\rm B}= \hat{k}_{\rm H} = k$ where
	\begin{align*}
	\hat{k}_{\rm B} &= \left|\left\{i,~ \mathscr{R}[s_i(B)] > \sqrt{c\Phi}\right\}\right| \\
	\hat{k}_{\rm H}	&= \left|\left\{i,~ \left.s_i(H_r)\right|_{r = \sqrt{c\Phi}} < 0\right\}\right|,
	\end{align*}
	where $\mathscr{R}$ indicates the real part.
	\label{rem:k}
\end{rem}
The remark is justified by the fact that $r = \sqrt{c\Phi}$ is the (limiting) right edge of the bulk of the non-backtracking matrix $B$ and that the eigenvalues of $B$ whose real part exceeds $\sqrt{c\Phi}$ (in the limit $n\to \infty$) are the isolated largest real eigenvalues of $B$. These are mapped, according to Figure~\ref{fig:zeta} and the discussion of Section~\ref{subsec:connection.ising}, to the number of communities. Likewise, from the connection between $H_r$ and $B$, for $r = \sqrt{c\Phi}$, the negative eigenvalues of $H_r$ are one-to-one mapped to the largest real eigenvalues of $B$.

\begin{algorithm}[t!]
	\begin{algorithmic}[1]
		\State \textbf{Input} : adjacency matrix of undirected graph $\mathcal{G}$
		\State Estimate $c = \frac{1}{n}\sum_{i}d_i$ and $\Phi = \frac{1}{nc}\sum_{i}d_i^2$
		\State Detect number of classes  $\hat{k}$ according to Remark 1
		\For{$2 \leq p \leq \hat{k}$}
		\State  $\zeta_p \leftarrow r$ such that $s_p^{\uparrow}(H_{r}) = 0$ on $1 < r \leq \sqrt{c\Phi}$
		\State  $X_{\bullet,p} \leftarrow \bm{x}_p$ such that $H_{\zeta_p}\bm{x}_p = 0$
		\EndFor
		\State Estimate community labels $\hat{\bm{\ell}}$ from the node embedding $\bm{X} = [\bm{X}_{\bullet,1}, \ldots, \bm{X}_{\bullet,\hat{k}}]$.
		\State
		\Return Estimated number $\hat{k}$ of communities and label vector $\hat{\bm{\ell}}$.
		\caption{Community Detection with the Bethe-Hessian}
		\label{alg:1}
	\end{algorithmic}
\end{algorithm}

\medskip

The arguments above naturally lead to Algorithm~\ref{alg:1} for community detection using the Bethe-Hessian matrix. Algorithm~\ref{alg:1} is a meta-algorithm, written for the user readability. Section~\ref{sec:sim} will provide an efficient implementation (Algorithm~\ref{alg:2}) of Algorithm~\ref{alg:1} accounting for deeper algorithmic considerations.

\begin{figure}[!t]
	\centering
	\includegraphics[width = \columnwidth]{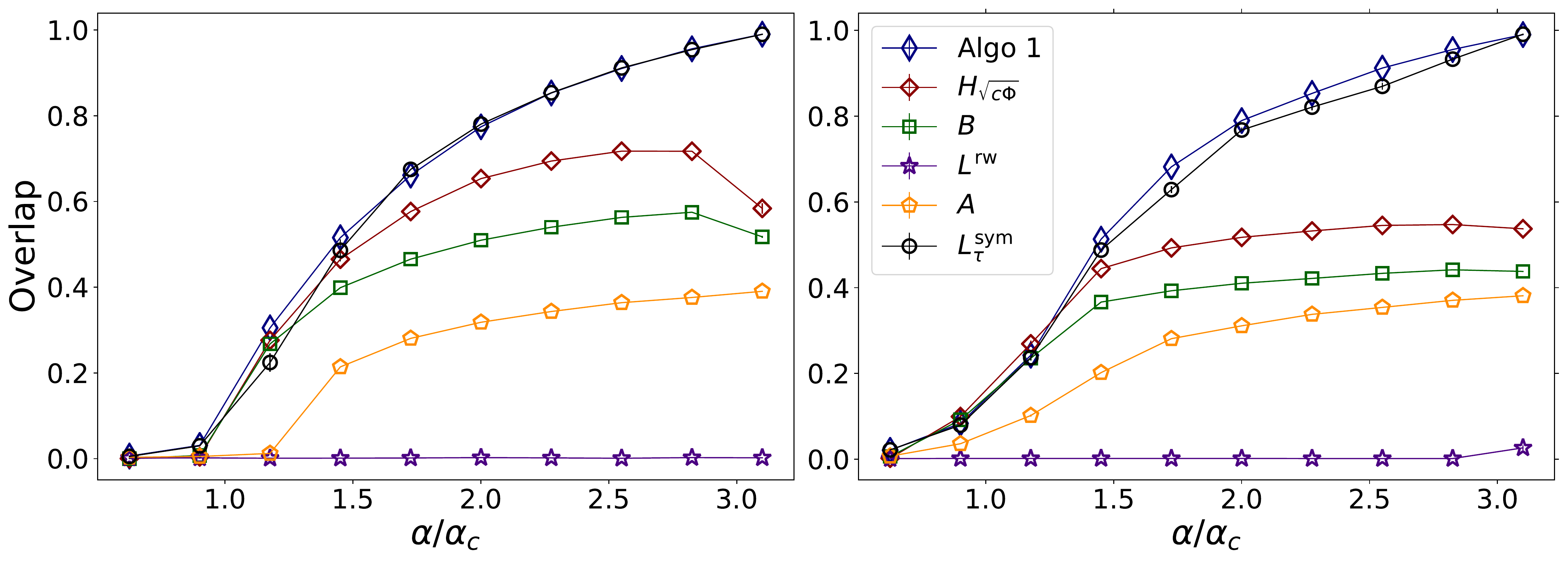}
	\caption{Overlap comparison as a function of the hardness $\alpha$ of the detection problem. Blue sharp diamonds are the result given by Algorithm \ref{alg:1}, red diamonds the algorithm using the Bethe-Hessian of \citep{saade2014spectral}, green squares are for the non-backtracking of \citep{krzakala2013spectral}, yellow pentagons the adjacency matrix, purple stars the random walk Laplacian, black dots are the algorithm of \citep{qin2013regularized}. For both graphs $n = 50.000$, $\theta_i \sim [\mathcal{U}(3,10)]^4$, $\bm{\pi} \propto \bm{1}_k$, $c = 5$. (Left): $k = 2$, $c_{\rm out} : 0.05 \to 4$. (Right): $k = 5$, $\pi_p \sim \mathcal{N}(k^{-1},(4k)^{-1})$, $C_{a>b} \sim \mathcal{N}(c_{\rm out}, c_{\rm out}k^{-1})$, $c_{\rm out} : 0.05 \to 4$. $\alpha := 2(c - c_{\rm out})/\sqrt{c}$ and $\alpha_c = 2/\sqrt{\Phi}$. Averages over $10$ samples.
		\vspace{-0.7cm}}
	\label{fig:overlap}
\end{figure}

Applying Algorithm~\ref{alg:1}, Figure~\ref{fig:overlap} compares the overlap performance (defined in Equation~\eqref{eq:overlap}), achieved by Algorithm~\ref{alg:1} versus competing methods for synthetic DC-SBM graphs as a function of the hardness of the problem $\alpha$.
\begin{equation}
{\rm Overlap} = \underset{\bm{\hat{\ell}} \in \mathcal{P}}{\rm max}~\frac{1}{1-\frac{1}{k}}\left(\sum_{i=1}^n \delta(\hat{\ell}_i,\ell_i)-\frac{1}{k}\right),
\label{eq:overlap}
\end{equation}

where $\hat{\bm{\ell}}$ is the estimate of the label eigenvector and $\mathcal{P}$ the set of the permutations of $\bm{\hat{\ell}}$.

\medskip

On the left hand-side of Figure~\ref{fig:overlap} is depicted a symmetric two-class scenario, for which $\alpha$ is defined as per Equation~\eqref{eq:detectability_SBM} and community reconstruction is asymptotically feasible if and only if $\alpha > \alpha_c$. On the right hand-side a more involved multiple class scenario is devised: the matrix $C$ is obtained here by drawing random Gaussian numbers with mean $c_{\rm out}$ and variance $f = c_{\rm out}/k$, before being symmetrized and tuning the diagonal elements to guarantee $C\Pi\bm{1}_k = c\bm{1}_k$. In this case, we still define $\alpha = 2(c-c_{\rm out})/\sqrt{c}$ but $\alpha_c$ now only represents the transition conjectured in \citep{decelle2011asymptotic} of the  planted partition model for which $f = 0$. In practice, in our simulation with $f\neq 0$, there are multiple transitions that are close (but in general different) to $\alpha_c$.

In both cases, Algorithm~\ref{alg:1} outperforms the competing algorithms. For $k = 2$, as $\alpha \to \alpha_c$, Algorithm~\ref{alg:1}, the algorithm of \citep{saade2014spectral} based on $H_{\sqrt{c\Phi}}$ and the one of \citep{krzakala2013spectral} based on $B$ give essentially the same result, which confirms that they are indeed equivalent at the phase transition. 
For easier detection problems though, except for the algorithm of \citep{qin2013regularized} which is only slightly less accurate, the performance of all methods is largely improved by Algorithm~\ref{alg:1}. Interestingly, it can be shown here that $|\sqrt{c\Phi} - \zeta_2| < |\nu_2\Phi - \zeta_2|$, giving an intuition on why the Bethe-Hessian method of \citep{saade2014spectral} performs better than the non-backtracking approach of \citep{krzakala2013spectral}. As for the standard spectral clustering algorithm which exploits the dominant eigenvectors of $A$, it can only perform non trivial community reconstruction for $\alpha$ far beyond the threshold $\alpha = \alpha_c$. The algorithm of \citep{shi2000normalized}, based on the popular random walk Laplacian $L^{\rm rw}$, is here incapable of making any non-trivial reconstruction for the considered set of parameters, suggesting that its dominant eigenvectors are not informative. 

This last point is quite interesting and raises the possibility that the informative eigenvectors of $L^{\rm rw}$ are possibly lost in the bulk. The next section will confirm this affirmation and make the statement more precise.

\section{Relation Between $H_r$ and the Regularized Random Walk Laplacian $L_{\tau}^{\rm rw}$}
\label{sec:reg_lap}

In this section we show that there is a strong connection between the Bethe-Hessian matrix and the regularized random walk Laplacian matrix $L_{\tau}^{\rm rw} = D_{\tau}^{-1}A$, where $D_{\tau} = D + \tau I_n$.
We will specifically prove that, when $\tau$ is carefully chosen, the matrix $L_{\tau}^{\rm rw}$ has isolated informative eigenvalues in its largest positions. 
Remarking that $L_{\tau}^{\rm rw}$ has the same eigenvalues as the matrix $L_{\tau}^{\rm sym} = D_{\tau}^{-1/2}AD_{\tau}^{-1/2}$, we also find a strong connection between $L_{\tau}^{\rm rw}$ and the community detection method of \citep{qin2013regularized}. 

The mapping between $H_r$ and $L_\tau^{\rm rw}$ unfolds from the following basic remark: for $\zeta_p > 1$ (as defined in Equation~\eqref{eq:zeta}),
\begin{align}
[(\zeta_p^2-1)I_n + D - \zeta_pA]\bm{x}_p &= 0 \label{eq:info_BH}\\
\Leftrightarrow [D + (\zeta_p^2-1)I_n]^{-1}A\bm{x}_p &= \frac{1}{\zeta_p}\bm{x}_p \label{eq:info_regL}
\end{align}
inducing a natural mapping between the Bethe-Hessian matrix at $r = \zeta_p$ to the regularized random walk Laplacian at $\tau=\zeta_p^2-1$. In particular, the vector $\bm{x}_p$ solution of Equation~\eqref{eq:info_BH} coincides with the eigenvector of $L_{\zeta_p^2-1}^{\rm rw}$ associated with the eigenvalue $1/\zeta_p$. 

In the following this relation is further analyzed and will be used to provide an alternative method to detect communities down to the detectability threshold.

\subsection{Main Result}
\label{sec:main}

The spectral methods of \citep{shi2000normalized,ng2002spectral} do not lead to good partitions in the sparse regime, as evidenced in Figure~\ref{fig:overlap}. The reason, which we briefly commented in Section~\ref{sec:intro}, is that the bulk of uninformative eigenvalues of both $L^{\rm rw}$ and $L^{\rm sym}$ undergo a certain spreading in the sparse regime (see Figure~\ref{fig:sparse_het}B) which ``swallows'' the informative eigenvalues within the bulk. This is not to say that informative eigenvectors vanish: rather, these eigenvectors are associated to \emph{non-isolated} eigenvalues; this was already observed in \citep{joseph2013impact} for the benchmark network \emph{Political blogs} \citep{adamic2005political}. 
This has two major negative consequences: (i) it may be practically infeasible (especially as the clustering task is more difficult) to identify the correct informative eigenvalue, and (ii) since the eigenvectors associated to close-by eigenvalues (in the bulk, the typical distance between consecutive eigenvalues is $O_n(1/n)$) tend to ``spread'' across the eigenvectors of these neighboring eigenvalues, not a single eigenvector but a collection of neighboring eigenvectors need to be considered. This is clearly deleterious to spectral clustering.

This problem is partially solved through regularization through a parameter $\tau$ (added to $D$ as $D_\tau=D+\tau I_n$): it was indeed shown in a series of related works that regularization helps clustering in sparse networks \citep{lei_consistency_2015,joseph2013impact, qin2013regularized}. In all these works, the results obtained (or inferred) are not straightforwardly applicable to the sparse regime. Furthermore, according to the analysis performed in these works (however only far from the transition point), large values of $\tau$ would seem preferable; however, in practice, it is rather observed that small values entail better performances.

\medskip

We answer here to the question of why small parametrizations $\tau$ should be preferred to large $\tau$, as well as which is the smallest value of $\tau$ for which we can guarantee the existence of isolated informative eigenvectors of $L_{\tau}^{\rm rw}$, down to the  detectability threshold. Besides, in accordance to the previous sections, we determine a value for $\tau$ which is optimal in its (i) ensuring the existence of these informative eigenvectors all the way to the detectability threshold, and most importantly (ii) which is resilient to degree heterogeneity in the graph. Formally, the result is formulated as follows.

\begin{claim}
	\label{prop:1}	
	Consider the graph $\mathcal{G}(\mathcal{V},\mathcal{E})$, built on a sparse DC-SBM as per Equation~\eqref{eq:DC-SBM} with $k$ communities. Let $\zeta_p$ for $2 \leq p \leq k$ be defined as per Equation~\eqref{eq:zeta} (imposing $\zeta_1 = 1$) and $\tau\in\mathbb{R}$ be such that $\zeta_p^2-1 \leq \tau \leq c\Phi - 1$. Then, under Assumption~\ref{ass:cpi}, for all large $n$ with high probability, the $p$ largest eigenvalues of the matrix $L_{\tau}^{\rm rw}$ are isolated. In particular, 
	\begin{align*}
	s_p^{\downarrow}(L^{\rm rw}_{\zeta_p^2-1}) = \frac{1}{\zeta_p}.
	\end{align*}
\end{claim}

Note that the proposition is not an obvious consequence of the equivalence between Equation~\eqref{eq:info_BH} and Equation~\eqref{eq:info_regL} as it is not clear that the eigenvalue $1/\zeta_p$ of $L^{\rm rw}_{\zeta_p^2-1}$ corresponds to the $p$-th largest and that it remains isolated (as is zero in the spectrum of $H_{\zeta_p}$).

\medskip

Since Claim~\ref{prop:1} asserts that the eigenvalue $1/\zeta_p$ of $L_{\zeta_p^2-1}^{\rm rw}$ is the $p$-th largest and is isolated, it can be clearly identified for all finite but large $n$: its corresponding informative eigenvector $\bm{x}_p$ (from Equation~\ref{eq:info_regL}) can thus not be confused with other eigenvectors. This eigenvector $\bm{x}_p$ is also the solution to $H_{\zeta_p}\bm{x}_p = 0$ (from Equation~\eqref{eq:info_BH}); its properties have been extensively discussed in Section~\ref{sec:connection}: it is in particular asymptotically insensitive to the degrees of the graph. Intuitively, by picking $\tau$ away from $\zeta_p^2-1$, the entries of the $p$-th eigenvector are likely to be more polluted by the degrees of the network. This suggests that large values of  $\tau$ \citep[as studied by the authors of][] {joseph2013impact} are likely to lead to sub-optimal partitions.

Claim~\ref{prop:1} further asserts that, for any $\tau$ in the interval $[\zeta_p^2-1,c\Phi-1]$, the $p$ dominant eigenvalues of $L_{\tau}^{\rm rw}$ are isolated. Since in addition $\zeta_p \leq \zeta_k \leq \sqrt{c\Phi}$, regardless of the hardness of the detection problem, the $k$ largest eigenvalues of $L_{c\Phi-1}^{\rm rw}$ (i.e., $L_{r^2-1}^{\rm rw}$ for $r=\sqrt{c\Phi}$) must be isolated. Note in passing that, in \citep{qin2013regularized}, the authors propose the regularization $\tau = \bar{d} = \frac{1}{n}\sum_i d_i$ from a heuristic intuition. While suboptimal according to the claim, this choice is somewhat meaningful as $\tau$ must indeed essentially grow with the average degree $\bar d\approx c$.

Besides, as a corollary, this provides a convenient alternative method to estimate the number of communities, based on the regularized Laplacian matrix.

\begin{rem}[Second estimation of the number of classes]
\label{rem:est_k}
	As a direct consequence of Claim~\ref{prop:1}, we have that, for all large $n$ with high probability $k=\hat{k}_{\rm B} = \hat{k}_{\rm H} = \hat{k}_{\rm L}$, where
	\begin{equation*}
	\hat{k}_{\rm L} = \left|\left\{i~: ~ s_i(L_{\tau}^{\rm rw})|_{\tau=c\Phi-1}\geq \frac{1}{\sqrt{c\Phi}}\right\}\right|.
	\end{equation*}
\end{rem}
Note that, since $S(L_{\tau}^{\rm rw}) = S(L_{\tau}^{\rm sym})$,  this calculation can be performed on a symmetric matrix, bringing gain in computational efficiency. 

\medskip

The technical supporting elements of Claim~\ref{prop:1} can be found in Appendix~\ref{app:proof}. In a nutshell, letting $\tau(r) = -s_p^{\uparrow}(D-rA)$, we show that there is a one-to-one map between the isolated eigenvalues of $L_{\tau(r)}^{\rm rw}$ and of $H_r$, only reversed in order (the smallest of $H_r$ are mapped to the largest of $L_{\tau(r)}^{\rm rw}$). This is particularly valid for $r=\zeta_p$ for which $-s_p^{\uparrow}(D-\zeta_pA) = \zeta_p^2-1$. We further show that the function $\tau(r)$ is bijective, thereby extending the results of Claim~\ref{claim:1} for $r \in [\zeta_p,\sqrt{c\Phi}]$ into $\tau \in [\zeta_p^2-1, c\Phi-1]$.

\subsection{Side Comments}

This section provides further interpretations and insights of the results introduced in the previous chapters. Firstly, we relate Algorithm~\ref{alg:1} to other commonly adopted methods for spectral clustering and secondly we discuss how our result can be extended in the presence of disassortative communities.

\subsubsection{Connection Among the Spectral Algorithms Based on $B, H_r, L_{\tau}^{\rm rw}, L^{\rm rw}, L$}

We showed in the previous sections the deep connection between the belief propagation equations and the regularized Laplacian matrix $L_{\tau}^{\rm rw}$, by successively passing through the non-backtracking $B$ and the Bethe-Hessian matrices $H_r$, so far treated in parallel (and with different tools) in the literature. From a practical perspective, we notably pointed that adding the regularization $\tau I_n$ to the degree matrix $D$ (i) favors efficient clustering in sparse networks but (ii) is optimally tuned for $\tau$ taken as a function of the hardness of the detection task (explicitly, for $k = 2$, a function of $\alpha=(c_{\rm in}-c_{\rm out})/\sqrt{c}$). In particular, for more challenging clustering tasks, larger values of $r$ in $H_r$ and $\tau$ in $L^{\rm rw}_\tau$ should be employed. On the opposite, in the limit of trivially simple clustering, $r\to 1$ (so that $H_r \to D-A$, the standard Laplacian) and $\tau \to 0$ (so $L_{\tau}^{\rm rw} \to L^{\rm rw}$, the standard random walk Laplacian). This suggests that the classical Laplacian matrices are still ``optimal'' in the sparse regime, yet only for easy clustering tasks, i.e., possibly far beyond the detectability threshold.

Most importantly, while the aforementioned ``difficulty'' of a clustering task is of course not accessible to the practitioner, we showed that the optimal tuning of the hyperparameters $r$ and $\tau$ can be practically obtained by retrieving isolated eigenvalues in the spectra of $H_r$, $B$, or $L^{\rm rw}_\tau$ (through the fundamental quantities $\zeta_2,\ldots,\zeta_k$). In passing, the relative distance of the $\zeta_p$'s to the bulk of non-informative eigenvalues is a further clue for the practitioner of the level of confidence of the ultimate clustering result.

\subsubsection{Extension to Disassortative Communities}
\label{subsec:extension}
Another remark concerns the possible existence of disassortative communities in the graph, i.e., groups of nodes which are identified as a class because they repel (rather than attract) each other. As a concrete example one may think of the vertices of a graph as the words contained in a text with edges if two words appear next to each other: on this graph, adjectives and nouns represent two disassortative communities \citep{newman2004finding}. Under a two-class DC-SBM model, a disassortative network can be easily generated by imposing $c_{\rm out} > c_{\rm in}$. In this case, the second largest eigenvalue of $C\Pi$ will therefore be negative and equal to $\nu_2 = (c_{\rm in} - c_{\rm out})/2 < 0$.

The problem of detection of two disassortative communities can be addressed both with the non-backtracking and the Bethe Hessian matrices \citep[as shown in][] {krzakala2013spectral,saade2014spectral}. In particular, in the presence of two disassortative communities, $s_2^{\downarrow_{|\cdot|}}(B) < 0$, where we recall $s_p^{\downarrow_{|\cdot|}}(M)$ indicates the $p$-th largest eigenvalues in modulus. Similarly, the matrix $H_r$ can detect two disassortative communities for $r < 0$, becoming a deformed version (equal in the limit of easy clustering) of the signless Laplacian $D+A$, frequently used to study bipartite graphs \citep{cvetkovic2007signless}.

\medskip

While defining communities when both assortative and disassortative elements are present is debatable, at least the result of Claim~\ref{prop:1} can be extended to the case where the matrix $C\Pi$ has possibly both positive and negative eigenvalues, as follows.

\begin{figure}[!t]
	\centering
	\includegraphics[width=\columnwidth]{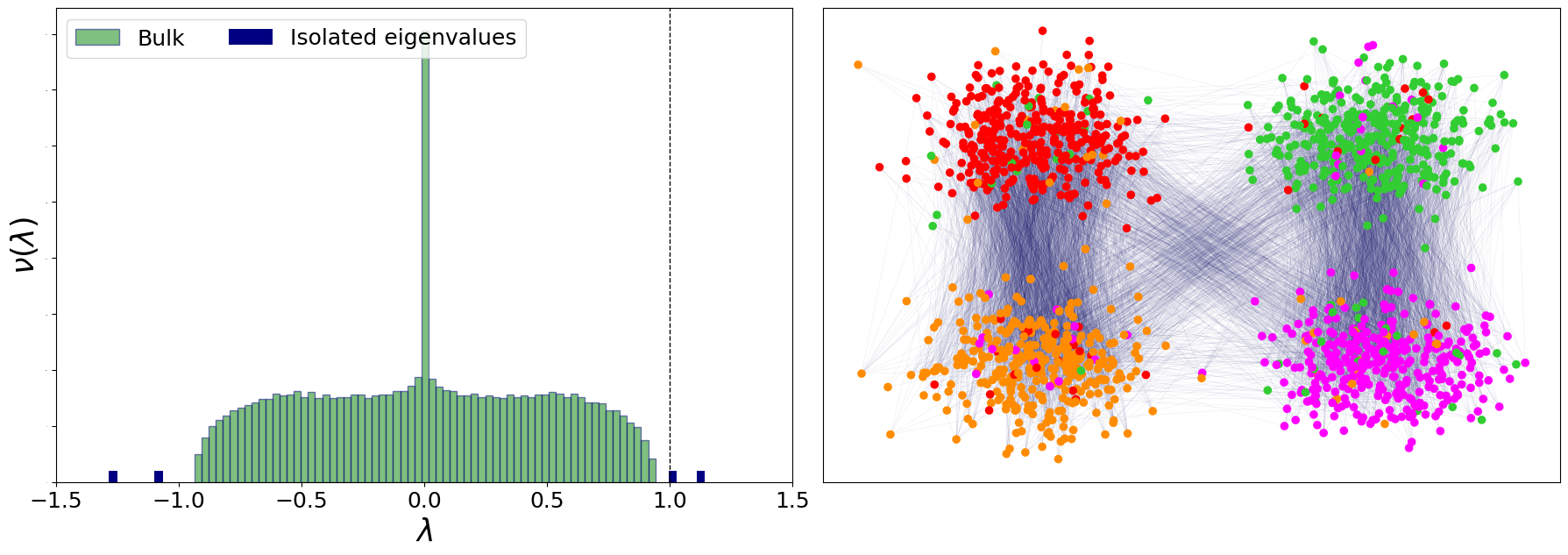}
	\caption{\vspace{-0.5cm}(Left) Spectrum of $\zeta_4L_{\zeta_4^2-1}^{\rm rw}$, $n = 5000$,  with $c_{\rm in} = 4$, $c_{\rm out} =0.5$, $\theta_i \sim [\mathcal{U}(3,5)]^4$ with $C$ generated according Equation \eqref{eq:C_hierarchy}. (Right) Realization of $\mathcal{G}$ from the SBM with $C$ defined as per Equation~\eqref{eq:C_hierarchy}, with $c_{\rm in} = 5$ and $c_{\rm out} = 1$, $n = 1500$. The position of the nodes is assigned according to the true labels as $\mathcal{N}(\mu_{\ell_i},I_2)$ and the colors are the output of Algorithm~\ref{alg:1} on $\mathcal{G}$.}
	\label{fig:dis}
\end{figure}

\begin{rem}[Eigenvalues of $C\Pi$ of arbitrary sign]
    Assume, unlike stipulated in Assumption~\ref{ass:cpi}, that the sign of the eigenvalues $\nu_p$ of $C\Pi$ is arbitrary (except for $\nu_1 = c$), but that $|\nu_p| > \sqrt{\frac{c}{\Phi}}$ in absolute value. We sort the eigenvalues $\nu_p$ according to their  modulus as $\nu_1 > |\nu_2| \geq \ldots \geq |\nu_k|$. In this setting, Claim~\ref{prop:1} generalizes by stating that the $p$ eigenvalues of $L_{\tau}^{\rm rw}$ \emph{with largest modulus} are isolated and that
	\begin{equation}
	s_{p}^{\downarrow_{|\cdot|}}(L_{\zeta_p^2-1}^{\rm rw}) = \frac{1}{\zeta_p} = \frac{\nu_p}{c} + o_n(1).
	\end{equation}
	\label{rem:2}
\end{rem}

As suggested above, while the mathematical result is well defined for $\nu_p < 0$, the interpretation of the negative eigenvalues of $C\Pi$ is not straightforward for $k>2$ and the very definition of a \emph{community} might not be evident. As an example, the right frame of Figure~\ref{fig:dis} displays a network designed from the SBM with $k = 4$, $\Pi \propto I_k$ and $C$ defined as:
\begin{align}
C = \begin{pmatrix}
c_{\rm in} & c_{\rm out} \\ c_{\rm out} &c_{\rm in} 
\end{pmatrix}\otimes \begin{pmatrix}
c_{\rm out} & c_{\rm in} \\ c_{\rm in} &c_{\rm out} 
\end{pmatrix}
\label{eq:C_hierarchy}
\end{align}
with `$\otimes$' the Kronecker product. This graph has a \emph{hierarchical} structure: it can be divided into two assortative communities, each of them composed of two disassortative communities. In Figure~\ref{fig:dis} (right) the position of each node is assigned according to its true label, while the color is assigned according to the output of $k$-means on the vectors $\bm{x}_p$ of $L_{\zeta_p^2-1}^{\rm rw}$ with largest modulus. We can observe that our algorithm performs well also in this particular case. 
For the sake of clarity, we close here the discussion on the generalization to cases where $\nu_p$ may be negative and, unless otherwise stated, we are thus once again under Assumption~\ref{ass:cpi}.

\section{Implementation of the Algorithm}
\label{sec:sim}

Based on the various discussions and results from the previous section, we now formally introduce our proposed algorithm along with pragmatic discussions on implementation cost, optimization and robustness to real-world network configurations.

\subsection{Applicability to Real Networks}

As opposed to the sparse stochastic block model, the DC-SBM model of Equation~\eqref{eq:DC-SBM} accounts both for the sparsity and heterogeneity of real networks. Yet, it does not capture other fundamental aspects of real graphs. For instance, the presence of many triangle in social networks \citep{holland1971transitivity} and more generally of short loops may invalidate the local tree-like approximation that underlies our theoretical findings. Nevertheless, its usability on real data sets suggest that some properties of $B$ are valid for various graph topologies. In particular, the following properties of $B$ seem to hold in general:
\begin{itemize}
  \setlength\itemsep{-0.5em}
	\item All complex eigenvalues come in pairs of complex conjugates and most of them are bounded by a circle on the complex plane of radius $r_{\rm radius} \approx \sqrt{\rho(B)}$. This statement originates both from empirical observations and from heuristic arguments on the asymptotic density of the eigenvalues of $B$ dicussed in \citep{krzakala2013spectral}.
	\item The number of real eigenvalues of $B$, different from $\{-1,0,1,\rho(B)\}$, is even. Half of the eigenvalues are larger in modulus than $r_{\rm radius}$, while the other half lies between $1$ and $r_{\rm radius}$. All of them are isolated.
\end{itemize}
It is easy to verify that the two earlier properties hold for \emph{any} graph topology if $D = cI_n$ and can be extended to the case in which $D \approx cI_n$ as suggested in \citep{coste2019eigenvalues}.
Given the connection between $B$ and the Bethe Hessian matrix $H_r$, from these two points, we can claim that the steps of Algorithm~\ref{alg:1} are all well defined on arbitrary graphs. 

\medskip

Some properties of $B$ are however likely model-dependent and thus not resilient to arbitrary graphs, such as
\begin{itemize}
  \setlength\itemsep{-0.5em}
	\item $\rho(B)$ may be far from $\frac{\sum_i d_i^2}{\sum_i d_i}$, which is a natural estimator for $c\Phi$ in the DC-SBM model.
	\item For an arbitrary network, $\zeta_p$ (as defined in Claim~\ref{claim:1}) may be far from $\frac{\rho(B)}{s_{R,p}^{\downarrow}(B)}$ in general.
\end{itemize} 
These observations impose that, when devising a spectral clustering algorithm adapted to real networks, these purely DC-SBM considerations should not be exploited.

\medskip

A further specificity of real graphs is their possibility to be inherently made of disjoint components. We have instead so far worked with the assumption that $\mathcal{G}$ has a giant component and that the communities are contained within the giant component. In practice, communities may live in disconnected subgraphs (in a two-class DC-SBM, this would correspond to setting $c_{\rm out} = 0$). In this case, one can perform a two-stage clustering. First the connected components are detected, looking into the eigenvectors with zero eigenvalue of $H_{\zeta_1} = D-A$. Afterwards, each connected component is treated independently. We will thus consider, without loss of generality, that the real graphs which we will consider are connected.

\subsection{Estimation of $k$}

The problem of estimating the number of communities in an unsupervised manner is in general non-trivial. Methods based on the non-backtracking rather than Bethe-Hessian matrix have been studied and exploited to efficiently recover communities regardless of the generative model \citep{le2015estimating}. In the course of our argumentation, we suggested different ways to estimate the number of communities, which are asymptotically equivalent (recall Remark~\ref{rem:k} and Remark~\ref{rem:est_k}):
\begin{enumerate}
	\item $\hat{k}_{\rm B} = |\{s_{R,i}(B) > \sqrt{\rho(B)}\}|$
	\item $\hat{k}_{\rm H} = |\{s_i\left(H_{\sqrt{\rho(B)}}\right) < 0\}|$
	\item $\hat{k}_{\rm L} = \left|\left\{s_i\left(L^{\rm sym}_{\rho(B)-1}\right) > \frac{1}{\sqrt{\rho(B)}}\right\}\right|$.
\end{enumerate}
From a computational standpoint, note that the eigenvalues of the matrix $B$ different from $\pm 1$ are the same as the eigenvalues of the matrix $B' \in \mathbb{R}^{2n\times 2n}$ \citep{krzakala2013spectral,coste2019eigenvalues},  defined as
\begin{align}
B' = \begin{pmatrix}
A&I_n-D\\I_n&0
\end{pmatrix}
\label{eq:Bprime}
\end{align}
so all computation involving the eigenvalues of $B$ can be performed in an efficient manner on the matrix $B'$. All three estimators above require the value of $\rho(B)$ that can be obtained efficiently by direct computation. But not all estimates perform the same in practice. Estimator~1 is in particular not efficient as not only real but also complex eigenvalues of $B$ need be evaluated, which sensibly slows down the algorithm. Estimators~2 and 3 do not suffer this limitation. Estimator~3 is nonetheless preferred as estimating eigenvalues with largest, rather than smallest, algebraic value can in general be performed more efficiently. Furthermore, when the matrix $C\Pi$ has at the same time both positive and negative eigenvalues, all communities can be detected using the same matrix $L_{\rho(B)-1}^{\rm sym}$, as opposed to the Bethe-Hessian for which one needs to consider both $H_{\sqrt{\rho(B)}}$ and $H_{-\sqrt{\rho(B)}}$. Consequently, the estimator $\hat{k}_{\rm L}$ is the one that will be adopted in our final Algorithm~\ref{alg:2}. The complexity of estimating $\hat{k}$ scales as $O(n\hat{k}^3)$, as computing the $p$  largest eigenvalues of a sparse matrix of size $n\times n$ (that is, containing $O(n)$ non-null elements) costs $O(np^2)$ with state-of-the-art methods such as restarted Arnoldi methods~\citep[see for instance][]{saad_numerical_2011}.

Subroutine \ref{sub:1} details how to estimate\footnote{Subroutine~\ref{sub:1} is actually a na\"ive implementation to estimate  the number of communities, as it requires to compute several times the same eigenvectors. In the \href{https://github.com/lorenzodallamico/CoDeBetHe.jl}{CoDeBetHe.jl} package a more efficient implementation is proposed, based random projections, that allows to efficiently estimate $\hat{k}$.} $\hat{k}$ and it will be referenced in Algorithm \ref{alg:2}. 

\begin{protocol}
\begin{algorithmic}[1]
	\State \textbf{Input} : adjacency matrix of a connected, undirected graph $\mathcal{G}$
	\State Compute $\rho(B) \leftarrow s_1^{\downarrow_{|\cdot|}}(B')$, with $B'$ defined in Equation~\eqref{eq:Bprime}
	\State $j \leftarrow 1$
	\While{$v > 0$}
	\State $v \leftarrow s_{j+1}^{\downarrow}(L_{\rho(B)-1}^{\rm sym}) - \frac{1}{\sqrt{\rho(B)}}$
	\State $j \leftarrow j+1$
	\EndWhile
	\State $\hat{k} \leftarrow j$\\
	\Return Estimated number $\hat{k}$ of communities.
	\caption{\texttt{estimate$\_$number$\_$of$\_$classes}}
	\label{sub:1}
\end{algorithmic}
\end{protocol}

Of course, as per Assumption~\ref{ass:cpi} in the DC-SBM setting, since for all $1\leq p\leq k$, $\nu_p > \sqrt{\frac{c}{\Phi}}$, all communities were claimed ``visible''. In practice though, this becomes a stringent condition. If, in particular, only $\kappa<k$ eigenvalues of $C\Pi$ exceed the threshold, fewer eigenvectors will be exploited and fewer classes will be looked for by the algorithm  \citep[as is also the case of the algorithms in][]{krzakala2013spectral,saade2014spectral}. In particular, $k-\kappa$ communities might be merged to close-by communities or spread across the $\kappa$ detected communities.

\subsection{Estimation of $\{\zeta_p\}$}

From Section~\ref{subsec:connection.BP}, the values of the $\zeta_p$ for $1 \leq p \leq k$ can be estimated as
\begin{align}
\zeta_p = \frac{c}{\nu_p} + o_n(1) = \frac{\rho(B)}{s_p^{\downarrow}(B)} + o_n(1).
\label{eq:lazy_zeta}
\end{align}
This estimation of $\zeta_p$ (via $B'$) is computationally efficient but may be quite inaccurate for graphs not generated from the DC-SBM model. 
Conversely, a naive line search for $\zeta_p \in (1,\sqrt{\rho(B)})$ satisfying $s^{\uparrow}_p(H_{\zeta_p}) = 0$ is computationally inefficient.

\medskip

We propose here a faster method, motivated by a Courant-Fischer theorem argument. The details and proof of convergence of the proposed method are given in Appendix~\ref{app:zeta}.

In a few words, starting from an initial guess $r_0$, we devise an iterative sequence $r_0, r_1, \dots$, such that $r_t \to \zeta_p$. In Appendix~\ref{app:zeta} we show that the convergence is guaranteed when setting $r_0 = \zeta_{p+1}$, under the convention $\zeta_{k+1} = \sqrt{\rho(B)}$. The values of $\zeta_p$ are then estimated from the largest, $\zeta_k$, to the smallest, $\zeta_2$. 

The algorithm builds on two parts: (i) the computation of a $p\times p$ matrix obtained from the eigenvectors of $H_{r_t}$ with computational cost of $O(np^2)$, (ii) a subsequent line-search using this matrix, with computational cost $O(p^3)$. The advantage of this method is that the line search (which requires many iterations) is computationally cheap, while the most expensive part of the algorithm needs to be performed much fewer times with respect to the greedy line-search to obtain the same accuracy. The total complexity of the algorithm needed to compute the vector $\bm{{\zeta}}=({\zeta}_1,\ldots,{\zeta}_k)^T$, scales as $O(nk^3)$.

\medskip

The proposed algorithm is described in the following subroutine. We indicate with $X_r \in \mathbb{R}^{n \times p}$ the matrix containing in its columns the $p$ smallest eigenvectors of $H_r$ and $S_r \in \mathbb{R}^{p \times p}$ the diagonal matrix with the corresponding eigenvalues. For further details, the reader is referred to Appendix~\ref{app:zeta}.

\begin{protocol}[!h]
\begin{algorithmic}[1]
	\State \textbf{Input} : adjacency matrix of a connected, undirected graph $\mathcal{G}$, number of classes $k$
	\State $r \leftarrow \sqrt{s_1^{\downarrow_{|\cdot|}}(B')}$, where $B'$ is defined in Equation~\eqref{eq:Bprime} 
    \State $p \leftarrow k$
	\While{$p > 1$}
	\Repeat
	\State   Compute $S_r = {\rm diag}\left(s_1^{\uparrow}(H_r),\dots, s_p^{\uparrow}(H_r)\right)$ and $X_r \in \mathbb{R}^{n\times p}$, with $H_r X_r = X_r S_r$
	\State	$\Lambda_r \leftarrow X_r^TDX_r \in \mathbb{R}^{p\times p}$ ($D \in \mathbb{N}^{n\times n}$, the diagonal degree matrix)
	\State $r^* \leftarrow r' : s_1^{\downarrow}(r'S_r + (r-r')\Lambda_r) = (r-r')(1+rr')$, with line-search on $r' \in (1,r)$
	\State $ r \leftarrow r^*$ 
	\Until{convergence}
	\State $\delta \leftarrow$ multiplicity of $s_p^{\uparrow}(H_r) = 0$ 
	\State Set $\zeta_{p}, \zeta_{p-1}, \cdots, \zeta_{p-\delta+1} {\rm ~to~} r$
	\State $p \leftarrow p - \delta$
	\EndWhile
	\Return Vector $\bm{{\zeta}}=({\zeta}_1,{\zeta}_2,\ldots,{\zeta}_k)^T$.
	\caption{\texttt{compute$\_\bm{\zeta}$}}
	\label{sub:2}
\end{algorithmic}
\end{protocol}

Note that, although the subroutine \texttt{compute}$\_{\bm{\zeta}}$ only outputs the vector $\bm{{\zeta}}$, it can also be used to directly compute the informative eigenvectors $\{\bm{x}_p\}$.

\begin{figure}[!t]
	\centering
	\includegraphics[width=\columnwidth]{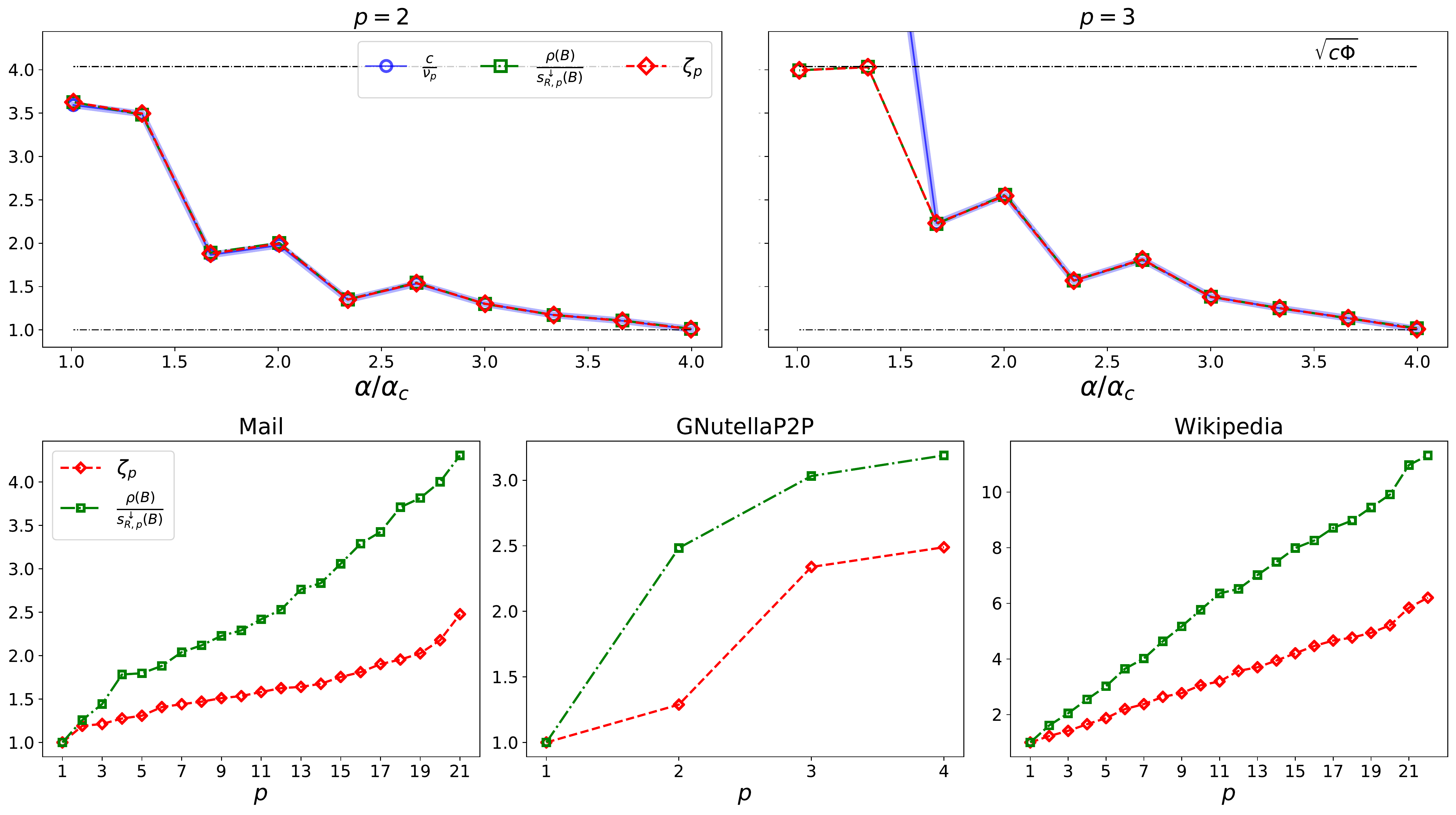}
	\caption{\vspace{-0.5cm}(Top: DC-SBM setting) Exact value of $c/\nu_p$ ($p = 2$ on the left and $p = 3$ on the right) in blue, compared to the estimate of $\zeta_p$ obtained using the dominant eigenvalues of $B$ (green) and its direct computation using Subroutine \ref{alg:2} (red), for problems of different hardness. For this simulation, $n = 50.000$, $k = 3$, $\theta_i \sim [\mathcal{U}(3,10)]^3$, $c = 10$. For each point $\pi_i \sim \mathcal{N}(1/k,1/2k)$ and the off-diagonal elements of $C$ are distributed as $\sim \mathcal{N}(c_{\rm out}, c_{\rm out}/k)$ and $c_{\rm out} : 0.1 \to 7.5$.
(Bottom: real networks) Estimate of the values of $\zeta_p$ as function of $p$, computed as $\rho(B)/s_p^{\downarrow}(B)$ (green squares) vs the direct computation of $\zeta_p$ as the output of Subroutine \ref{sub:2} (red diamonds)
 on the three real networks taken from \citep{snapnets}. The value of $\hat{k}$ is estimated according to Subroutine \ref{sub:1}.\vspace{-0.5cm}}
 \label{fig:zeta_estimation}
\end{figure}

Figure~\ref{fig:zeta_estimation} provides a typical output of the computation of $\bm{\zeta}$ and confirms the accuracy of the proposed algorithm. 
On the top line the algorithm is tested on a network created from the DC-SBM model, for which $\rho(B)/s_{R,p}^{\downarrow}(B)$ is a valid estimator for $\zeta_p$. The horizontal line indicates $\sqrt{\rho(B)} = \sqrt{c\Phi}$ which is the upper bound of $\zeta_p$. In our simulations, $C$ is generated randomly (see the caption of Figure~\ref{fig:zeta}), for the first two points we obtained $\nu_3 < \sqrt{c\Phi}$, invalidating Assumption \ref{ass:cpi}. In this case we see that $c/\nu_p > \sqrt{c\Phi}$ and the corresponding estimated value of $\zeta_p$ saturates at $\sqrt{c\Phi}$. On the contrary, whenever Assumption \ref{ass:cpi} is verified, the estimate of $\zeta_p$ is correct.

In the bottom line we compare the two methods to estimate $\zeta_p$ on the three real networks that clearly show that these two methods are different on graphs not generated from the DC-SBM. Given that Subroutine~\ref{sub:2} provides a direct computation of the eigenvalues of $B$, it should be preferred.

\medskip

For illustrative purposes, we tested the execution time of Subroutine~\ref{sub:2} for a SBM network with $k = 4$ classes of equal size, $n = 10^5$, $c_{\rm in} = 26.20$ on the diagonal elements of $C$ and $c_{\rm out} = 4.60$ for the off diagonal elements. The values of $\zeta_p$ were computed to machine precision in approximately $40$ seconds on a standard laptop using our Python implementation, while in approximately $5$ seconds with the \href{https://github.com/lorenzodallamico/CoDeBetHe.jl}{CoDeBetHe.jl} Julia implementation. The complexity of this subroutine scales linearly with $n$, and thus can be applied to large networks but cubically with respect to $k$ (and not quadratically as usual in the spectral clustering context) decreasing the computational efficiency when a rather large number of classes is present.

\subsection{Projection of the Embedded Points on a Hyper-Sphere}

The study performed so far identifies the presence of $k$ informative eigenvalues and describes the content of the associated $k$ eigenvectors $\bm{X}\equiv [\bm{x}_1,\ldots,\bm{x}_k]\in\mathbb{R}^{n\times k}$ with $\bm{x}_p$ the eigenvector associated to the $p$-th smallest eigenvalue of $H_{\zeta_p}$ (in particular, the vectors $\bm{x}_p$'s need not be orthogonal). 
The rows $\bm{X}_{1,\bullet},\ldots,\bm{X}_{n,\bullet} \in\mathbb{R}^k$ of the matrix $\bm{X}$ form a $k$-dimensional feature for every node, which are used in a last small-dimensional clustering step, usually employing the \emph{k-means} algorithm. The fact that \emph{k-means} is particularly efficient when the low-dimensional clusters are quite ``isotropic'' strongly motivates the need for the entries of the vectors $\bm{x}_p$ not to be affected by the node degrees (which would otherwise spread the clusters unevenly).

Yet, to further tackle residual degree dependence, a classical method, prior to \emph{k-means}, consists in normalizing all vectors $\bm{X}_{i,\bullet}$ to $\|\bm{X}_{i,\bullet}\|=1$ (this is Step~4 of Algorithm~\ref{alg:0}) . This method is motivated by the assumption that the degree dependence in each ${X}_{ij}$ is separable from the label dependence, a fact that is verified in sufficiently dense DC-SBM networks \citep{jin2015fast} and to some extent also in sparser graphs \citep{qin2013regularized}. Besides, under this normalization, the \emph{k-means} algorithm is restricted to the unitary hypersphere, improving its convergence to the genuine solution, especially when in presence of many communities.

\medskip

As such, while our proposed algorithm naturally discards degree dependence in the entries of $\bm{X}$ under the DC-SBM setting, the reality of practical networks may disrupt this expected behavior and the projection of the vectors $\bm{X}_{i,\bullet}$ on the unit hyper-sphere both alleviates this deleterious effect and further improves the convergence of \emph{k-means}. We thus adopt this normalization step in our final Algorithm~\ref{alg:2} and will confirm its practical gains when clustering real graphs.

The total theoretical complexity of Alg. 2 is dominated by subroutines 1 and 2, that both run in $O(n\hat{k}^3)$, as the \emph{k-means} step costs only $O(n\hat{k}^2)$. This is to compare with the usual complexity of spectral clustering algorithms on sparse graphs that are in $O(n\hat{k}	^2)$~\citep[see for instance][]{tremblay_approximating_2020}. This additional cost comes with a better classification performance in many real-world graphs, as discussed in the next section. In practice, to give an order of magnitude of computation times\footnote{The laptop’s RAM is 7.7Gb with Intel Core i7-6600U CPU @ 2.6GHz x 4}, running Algorithm 2 on a SBM\footnote{These times refer to our Python implementation, while the \href{https://github.com/lorenzodallamico/CoDeBetHe.jl}{CoDeBetHe.jl} implementation runs for $n = 10^5$ (resp., $10^6$) in approximately $15$ (resp., $450$) seconds if $k$ is not known and in approximately $6$ (resp., $100$) seconds if $k$ is known a priori.} with $k = 4$ classes of equal size, $n = 10^5$ (resp. $10^6$), $c_{\rm in} = 26.20$ on the diagonal elements of $C$ and $c_{\rm out} = 4.60$ for the off diagonal elements, takes approximately $85$ (resp., $2200$) seconds on a laptop. If $k$ is known in advance, this times drop to approximately $45$ (resp., $700$) seconds.

\begin{algorithm}[t!]
	\begin{algorithmic}[1]
		\State \textbf{Input} : adjacency matrix of a connected, undirected graph $\mathcal{G}$
		\State $\hat{k} \leftarrow \texttt{estimate\_number\_of\_classes}$ (Subroutine \ref{sub:1})
		\State $\bm{{\zeta}} \leftarrow$
		\texttt{compute$\_\bm{\zeta}$} (Subroutine \ref{sub:2})
		\State Initialize $\bm{X} \in \mathbb{R}^{n \times \hat{k}}$
		\For{$p =1:\hat{k}$}
		\State $\bm{X}_{\bullet,p} \leftarrow \bm{x}_p \in \mathbb{R}^n$, where $\bm{x}_p$ is the eigenvector with eigenvalue $s_p^{\uparrow}(H_{{\zeta}_p}) = 0$.
		\EndFor
		\State Normalize the rows of $\bm{X}_{i,\bullet} \leftarrow \bm{X}_{i,\bullet}/{\Vert \bm{X}_{i,\bullet} \Vert}$
		\State Estimate community labels $\hat{\bm{\ell}}$ as output of $\hat{k}$-class \emph{k-means} on the rows of $\bm{X}$.\\
		\Return Estimated number $\hat{k}$ of communities and label vector $\hat{\bm{\ell}}$.
		\caption{Community Detection in sparse and heterogeneous graphs}
		\label{alg:2}
	\end{algorithmic}
\end{algorithm}

\subsection{Algorithm and Performance on Real Networks}

In this section we compare the performance of Algorithm~\ref{alg:2} versus competing spectral methods on real-world networks that do not have ground-truth label assignment as well as LFR synthetic benchmark networks. 

Measuring the quality of an inferred partition $\bm{\hat{\ell}}$ is in general not straightforward since communities are not uniquely defined. Two different scores are then adopted adopted. One is the modularity\footnote{Note that the measure of the modularity is meaningful on \emph{assortative} or \emph{disassortative} networks but not on ``hybrid'' networks for which a more involved description would be needed.} \citep{newman2004finding,Newman-2006}, $\mathcal M$:
\begin{equation}
\mathcal{M} = \frac{1}{2|\mathcal{E}|}\sum_{i,j = 1}^n \left(A_{ij} - \frac{d_id_j}{2|\mathcal{E}|}\right)\delta(\hat{\ell}_i,\hat{\ell}_j).
\label{eq:modularity}
\end{equation}
High values of $\mathcal{M}$ correspond to good quality partitions. Alternatively, the partition quality is evaluated in terms of (normalized) posterior negative log-likelihood of the DC-SBM, $\mathcal{L}$:
\begin{align}
\mathcal{L} = - \frac{1}{2|\mathcal{E}|}\left[\sum_{(ij)\in\mathcal{E}} {\rm log}\left(\hat{\theta}_i\hat{\theta}_j\frac{\hat{C}_{\hat{\ell}_i,\hat{\ell}_j}}{n}\right) + \sum_{(ij)\notin\mathcal{E}} {\rm log}\left(1-\hat{\theta}_i\hat{\theta}_j\frac{\hat{C}_{\hat{\ell}_i,\hat{\ell}_j}}{n}\right)\right],
\label{eq:loglikelihood}
\end{align}
where $\hat{\theta}_i = d_i/\bar{d}$ and $\hat{C}_{ab} = \left( \sum_{i~:~\hat{\ell}_i = a} \sum_{j~:~\hat{\ell}_j = b} A_{ij}\right)/\left(\sum_{i~:~\hat{\ell}_i = a} \sum_{j~:~\hat{\ell}_j = b} \theta_i\theta_j\right)$. 
Good quality clustering correspond, in this case, to low values $\mathcal{L}$. 

\medskip

Table~\ref{tab:unlabeled} compares the results of different clustering algorithms\footnote{The choice of the spectral algorithm considered for comparison is based on two criterion: i) $H_{\sqrt{c\Phi}}$, $B$ and $L_{\tau}^{\rm sym}$ are the state-of-the-art spectral methods for community detection in \emph{sparse} graphs, while $A$ and $L^{\rm rw}$ are algorithms of great relevance in the literature; ii) these are the methods that are well explained by our unified framework.} of 15 real-world networks of increasing size. For all networks, the number of communities, when not available, is estimated through $\hat{k}_{\rm L}$ (see Subroutine \ref{sub:1}) and then the same value is used for all competing techniques (which in general do not provide their own dedicated estimator of $k$). The underlined numbers in the $k$ column indicate instead that $k$ is known. Furthermore, for all networks, community detection is performed only on the largest connected component of the graph and $n, c, \Phi, k$ refer to the characteristics of this dominant connected component.

			\begin{sidewaystable}
			\centering
			\footnotesize
			\hspace*{-.5cm}
			\begin{tabular}{|c||c|c|c|c||c|c|a|c|c|c|c|c|}
				\hline
				{Data set} & $n$ &$c$ &$\Phi$ & $k$ & Alg~\ref{alg:1}$_B$ & Alg~\ref{alg:2}$_{\rm wp}$ & Alg \ref{alg:2} 
				& $A$& $H_{\sqrt{c\Phi}}$ & $B$& $L^{\rm rw}$& $L_{\tau}^{\rm sym}$ \\
				\hline\hline
				
				
				Karate & 34 & 4.6 & 1.7 & \underline{2} & 
				
				\begin{tabular}{@{}l@{}}
					\textbf{0.86}\\
					\textbf{0.37}\\
				\end{tabular} & 
				
				\begin{tabular}{@{}l@{}}
					\textbf{0.86}\\
					\textbf{0.37}\\
				\end{tabular} &
				
				\begin{tabular}{@{}l@{}}
					\textbf{0.86}\\
					\textbf{0.37}\\
				\end{tabular} &
				
				\begin{tabular}{@{}l@{}}
					\textbf{0.86}\\
					\textbf{0.37}\\
				\end{tabular} &
				
				\begin{tabular}{@{}l@{}}
					\textbf{0.86}\\
					\textbf{0.37}\\
				\end{tabular} &
				
				\begin{tabular}{@{}l@{}}
					\textbf{0.86}\\
					\textbf{0.37}\\
				\end{tabular} &
				
				\begin{tabular}{@{}l@{}}
					0.98\\
					0.36\\
				\end{tabular} &
				
				\begin{tabular}{@{}l@{}}
					\textbf{0.86}\\
					\textbf{0.37}\\
				\end{tabular}\\ \hline

				
				{Dolphins} & 62 & 5 & 1.3 & \underline{2} &
				
				\begin{tabular}{@{}l@{}}
					\textbf{1.25}\\
					\textbf{0.38}\\
				\end{tabular} & 
				
				\begin{tabular}{@{}l@{}}
					\textbf{1.25}\\
					\textbf{0.38}\\
				\end{tabular} &
				
				\begin{tabular}{@{}l@{}}
					\textbf{1.25}\\
					\textbf{0.38}\\
				\end{tabular} &
				
				\begin{tabular}{@{}l@{}}
					{1.38 \ppm 0.01}\\
					0.22 \ppm 0.01\\
				\end{tabular} &
				
				\begin{tabular}{@{}l@{}}
					1.30 \ppm 0.03\\
					0.32 \ppm 0.03\\
				\end{tabular} &
				
				\begin{tabular}{@{}l@{}}
					1.38\\
					0.22\\
				\end{tabular} &
				
				\begin{tabular}{@{}l@{}}
					\textbf{1.25}\\
					\textbf{0.38}\\
				\end{tabular} &
				
				\begin{tabular}{@{}l@{}}
					\textbf{1.25}\\
					\textbf{0.38}\\
				\end{tabular}\\ \hline
				
				
				{Polbooks} & 105 & 8.4 &  1.4 & \underline{3} & 
				
				\begin{tabular}{@{}l@{}}
					\textbf{1.17}\\
					\textbf{0.51}\\
				\end{tabular} & 
				
				\begin{tabular}{@{}l@{}}
					1.16\\
					0.50\\
				\end{tabular} &
				
				\begin{tabular}{@{}l@{}}
					\textbf{1.17}\\
					\textbf{0.51}\\
				\end{tabular} &
				
				\begin{tabular}{@{}l@{}}
					1.26 \ppm 0.01\\
					~~~~0.46\\
				\end{tabular} &
				
				\begin{tabular}{@{}l@{}}
					1.21\\
					0.50\\
				\end{tabular} &
				
				\begin{tabular}{@{}l@{}}
					1.28 \ppm 0.01\\
					0.47 \ppm 0.01\\
				\end{tabular} &
				
				\begin{tabular}{@{}l@{}}
					\textbf{1.16}\\
					0.50\\
				\end{tabular} &
				
				\begin{tabular}{@{}l@{}}
					1.22\\
					\textbf{0.51}\\
				\end{tabular}\\ \hline
				
				
				{Football} & 115 & 10.7 & 1 & \underline{12}&
				
				\begin{tabular}{@{}l@{}}
					\textbf{0.83} \\
					\textbf{0.60}
				\end{tabular} & 
				
				\begin{tabular}{@{}l@{}}
					\textbf{0.83} \\
					\textbf{0.60}
				\end{tabular} &
				
				\begin{tabular}{@{}l@{}}
					\textbf{0.83} \\
					\textbf{0.60}
				\end{tabular} &
				
				\begin{tabular}{@{}l@{}}
					\textbf{0.83} \\
					\textbf{0.60}
				\end{tabular} &
				
				\begin{tabular}{@{}l@{}}
					\textbf{0.83} \\
					\textbf{0.60}
				\end{tabular} &
				
				\begin{tabular}{@{}l@{}}
					\textbf{0.83} \\
					\textbf{0.60}
				\end{tabular} &
				
				\begin{tabular}{@{}l@{}}
					\textbf{0.83} \\
					\textbf{0.60}
				\end{tabular} &
				
				\begin{tabular}{@{}l@{}}
					\textbf{0.83} \\
					\textbf{0.60}
				\end{tabular}\\ \hline
				
				
				{Mail} & 1133 & 9.6 & 1.9 & 21&
				
				\begin{tabular}{@{}l@{}}
					1.95 \ppm 0.01\\
					0.45 \ppm 0.01\\
				\end{tabular} & 
				
				\begin{tabular}{@{}l@{}}
					\textbf{1.89 \ppm 0.02}\\
					0.49 \ppm 0.01\\
				\end{tabular} &
				
				\begin{tabular}{@{}l@{}}
					\textbf{1.88 \ppm 0.01}\\
					\textbf{0.52 \ppm 0.01}\\
				\end{tabular} &
				
				\begin{tabular}{@{}l@{}}
					2.13 \ppm 0.01\\
					~~~~0.31\\
				\end{tabular} &
				
				\begin{tabular}{@{}l@{}}
					2.03 \ppm 0.01\\
					~~~~0.40\\
				\end{tabular} &
				
				\begin{tabular}{@{}l@{}}
					2.05 \ppm 0.01\\
					0.36 \ppm 0.01\\
				\end{tabular} &
				
				\begin{tabular}{@{}l@{}}
					1.99 \ppm 0.03\\
					0.50 \ppm 0.02\\
				\end{tabular} &
				
				\begin{tabular}{@{}l@{}}
					\textbf{1.87 \ppm 0.01}\\
					~~~~\textbf{0.51}\\
				\end{tabular}\\ \hline
				
				
				{Polblogs} & 1222 & 27,4 & 3 & \underline{2}&
				
				\begin{tabular}{@{}l@{}}
					\textbf{1.50}\\
					\textbf{0.43}\\
				\end{tabular} & 
				
				\begin{tabular}{@{}l@{}}
					\textbf{1.50}\\
					\textbf{0.43}\\
				\end{tabular} &
				
				\begin{tabular}{@{}l@{}}
					\textbf{1.50}\\
					\textbf{0.43}\\
				\end{tabular} &
				
				\begin{tabular}{@{}l@{}}
					1.62\\
					0.25\\
				\end{tabular} &
				
				\begin{tabular}{@{}l@{}}
					1.63\\
					0.27\\
				\end{tabular} &
				
				\begin{tabular}{@{}l@{}}
					1.65\\
					0.24\\
				\end{tabular} &
				
				\begin{tabular}{@{}l@{}}
					1.73\\
					0.00\\
				\end{tabular} &
				
				\begin{tabular}{@{}l@{}}
					\textbf{1.50}\\
					\textbf{0.43}\\
				\end{tabular}\\ \hline
				
				
				{Tv} & 3892 &8.9&3&41&
				
				\begin{tabular}{@{}l@{}}
					2.01 \ppm 0.03 \\
					0.55 \ppm 0.02\\
				\end{tabular} & 
				
				\begin{tabular}{@{}l@{}}
					1.89 \ppm 0.09 \\
					0.70 \ppm 0.06\\
				\end{tabular} &
				
				\begin{tabular}{@{}l@{}}
					\textbf{1.50 \ppm 0.01}\\
					\textbf{~~~~0.84}\\
				\end{tabular} &
				
				\begin{tabular}{@{}l@{}}
					1.99 \ppm 0.09\\
					0.53 \ppm 0.04\\
				\end{tabular} &
				
				\begin{tabular}{@{}l@{}}
					2.02 \ppm 0.05\\
					0.53 \ppm 0.02\\
				\end{tabular} &
				
				\begin{tabular}{@{}l@{}}
					2.05 \ppm 0.05\\
					0.53 \ppm 0.02\\
				\end{tabular} &
				
				\begin{tabular}{@{}l@{}}
					2.06 \ppm 0.15\\
					0.59 \ppm 0.12\\
				\end{tabular} &
				
				\begin{tabular}{@{}l@{}}
					\textbf{1.47 \ppm 0.01}\\
					0.79 \ppm 0.01\\
				\end{tabular}\\ \hline
				
				
				{Facebook} & 4039 &  43.7 &2.4& 55&
				
				\begin{tabular}{@{}l@{}}
					1.17 \ppm 0.04\\
					0.52 \ppm 0.01\\
				\end{tabular} & 
				
				\begin{tabular}{@{}l@{}}
					1.05 \ppm 0.06\\
					\textbf{0.76 \ppm 0.02}\\
				\end{tabular} &
				
				\begin{tabular}{@{}l@{}}
					\textbf{0.88 \ppm 0.01}\\
					\textbf{~~~~0.76}\\
				\end{tabular} &
				
				\begin{tabular}{@{}l@{}}
					1.13 \ppm 0.03\\
					0.44 \ppm 0.03\\
				\end{tabular} &
				
				\begin{tabular}{@{}l@{}}
					1.21 \ppm 0.03\\
					0.47 \ppm 0.01\\
				\end{tabular} &
				
				\begin{tabular}{@{}l@{}}
					1.21 \ppm 0.03\\
					0.47 \ppm 0.02\\
				\end{tabular} &
				
				\begin{tabular}{@{}l@{}}
					1.07 \ppm 0.14\\
					\textbf{0.75 \ppm 0.05}\\
				\end{tabular} &
				
				\begin{tabular}{@{}l@{}}
					\textbf{0.87 \ppm 0.01}\\
					0.59 \ppm 0.01\\
				\end{tabular} \\ \hline
				
				
				{GrQc} & 4158& 6.5&2.8& 29&
				
				\begin{tabular}{@{}l@{}}
					2.40 \ppm 0.04\\
					0.50 \ppm 0.02\\
				\end{tabular} & 
				
				\begin{tabular}{@{}l@{}}
					2.22 \ppm 0.11\\
					0.68 \ppm 0.09\\
				\end{tabular} &
				
				\begin{tabular}{@{}l@{}}
					\textbf{1.89 \ppm 0.01}\\
					\textbf{~~~~0.80}\\
				\end{tabular} &
				
				\begin{tabular}{@{}l@{}}
					2.39 \ppm 0.06\\
					0.49 \ppm 0.04\\
				\end{tabular} &
				
				\begin{tabular}{@{}l@{}}
					2.40 \ppm 0.04\\
					0.50 \ppm 0.02\\
				\end{tabular} &
				
				\begin{tabular}{@{}l@{}}
					2.39 \ppm 0.01\\
					0.50 \ppm 0.01\\
				\end{tabular} &
				
				\begin{tabular}{@{}l@{}}
					2.44 \ppm 0.09\\
					0.58 \ppm 0.10\\
				\end{tabular} &
				
				\begin{tabular}{@{}l@{}}
					\textbf{1.90 \ppm 0.02}\\
					~~~~0.78\\
				\end{tabular}\\ \hline
				
				
				{Power grid} & 4941 & 2.7&1.5&25&
				
				\begin{tabular}{@{}l@{}}
					3.63 \ppm 0.02\\
					0.32 \ppm 0.01\\
				\end{tabular} & 
				
				\begin{tabular}{@{}l@{}}
					2.61 \ppm 0.01\\
					~~~~0.92\\
				\end{tabular} &
				
				\begin{tabular}{@{}l@{}}
					\textbf{2.57 \ppm 0.01}\\
					\textbf{~~~~0.93}\\
				\end{tabular} &
				
				\begin{tabular}{@{}l@{}}
					3.84 \ppm 0.01\\
					0.17 \ppm 0.01\\
				\end{tabular} &
				
				\begin{tabular}{@{}l@{}}
					3.63 \ppm 0.03\\
					0.35 \ppm 0.03\\
				\end{tabular} &
				
				\begin{tabular}{@{}l@{}}
					3.71 \ppm 0.01\\
					0.28 \ppm 0.01\\
				\end{tabular} &
				
				\begin{tabular}{@{}l@{}}
					2.61 \ppm 0.01\\
					\textbf{~~~~0.92}\\
				\end{tabular} &
				
				\begin{tabular}{@{}l@{}}
					2.80 \ppm 0.02\\
					0.86 \ppm 0.01\\
				\end{tabular}\\ \hline
				
				
				{Politicians} & 5908 & 14.1 & 3 & 62 &
				
				\begin{tabular}{@{}l@{}}
					1.78 \ppm 0.04\\
					0.52 \ppm 0.02\\
				\end{tabular} & 
				
				\begin{tabular}{@{}l@{}}
					1.60 \ppm 0.05\\
					\textbf{0.83 \ppm 0.01}\\
				\end{tabular} &
				
				\begin{tabular}{@{}l@{}}
					1.43 \ppm 0.01\\
					\textbf{0.84 \ppm 0.01}\\
				\end{tabular} &
				
				\begin{tabular}{@{}l@{}}
					1.94 \ppm 0.06\\
					0.54 \ppm 0.02\\
				\end{tabular} &
				
				\begin{tabular}{@{}l@{}}
					1.90 \ppm 0.02\\
					0.50 \ppm 0.02\\
				\end{tabular} &
				
				\begin{tabular}{@{}l@{}}
					1.92 \ppm 0.02\\
					0.47 \ppm 0.01\\
				\end{tabular} &
				
				\begin{tabular}{@{}l@{}}
					1.77 \ppm 0.11\\
					0.78 \ppm 0.07\\
				\end{tabular} &
				
				\begin{tabular}{@{}l@{}}
					\textbf{1.40 \ppm 0.01}\\
					0.73 \ppm 0.01\\
				\end{tabular}\\ \hline
				
				
				{GNutella P2P} & 6299 & 6.6 & 2.7& 4 &
				
				\begin{tabular}{@{}l@{}}
					3.32 \ppm 0.01\\
					0.20 \ppm 0.01\\
				\end{tabular} & 
				
				\begin{tabular}{@{}l@{}}
					3.28\\
					0.26\\
				\end{tabular} &
				
				\begin{tabular}{@{}l@{}}
					\textbf{3.21}\\
					\textbf{0.40}\\
				\end{tabular} &
				
				\begin{tabular}{@{}l@{}}
					~~~~3.33\\
					0.19\ppm 0.01\\
				\end{tabular} &
				
				\begin{tabular}{@{}l@{}}
					3.37\\
					0.14\\
				\end{tabular}&
				
				\begin{tabular}{@{}l@{}}
					3.38\ppm 0.01\\
					0.13\ppm 0.01\\
				\end{tabular} &
				
				\begin{tabular}{@{}l@{}}
					3.41\\
					0.00\\
				\end{tabular} &
				
				\begin{tabular}{@{}l@{}}
					3.27\\
					0.35\\
				\end{tabular}\\ \hline
				
				
				{Wikipedia} & 7066 & 28.3&5.1&22&
				
				\begin{tabular}{@{}l@{}}
					1.93\\
					0.20\\
				\end{tabular} & 
				
				\begin{tabular}{@{}l@{}}
					\textbf{1.88 \ppm 0.01}\\
					0.23 \ppm 0.01\\
				\end{tabular} &
				
				\begin{tabular}{@{}l@{}}
					\textbf{1.88}\\
					0.26\\
				\end{tabular} &
				
				\begin{tabular}{@{}l@{}}
					1.99\ppm 0.01\\
					~~~~0.15\\
				\end{tabular} &
				
				\begin{tabular}{@{}l@{}}
					1.97\\
					0.17\\
				\end{tabular} &
				
				\begin{tabular}{@{}l@{}}
					~~~~1.96\\
					0.14\ppm 0.01\\
				\end{tabular} &
				
				\begin{tabular}{@{}l@{}}
					1.94 \ppm 0.03\\
					\textbf{0.38 \ppm 0.02}\\
				\end{tabular} &
				
				\begin{tabular}{@{}l@{}}
					\textbf{1.88}\\
					0.27\\
				\end{tabular}\\ \hline
				
				
				{HepPh} &11204     & 21.0 & 6.2& 60 &
				
				\begin{tabular}{@{}l@{}}
					1.75 \ppm 0.06\\
					0.39 \ppm 0.02\\
				\end{tabular} & 
				
				\begin{tabular}{@{}l@{}}
					1.98 \ppm 0.05\\
					0.37 \ppm 0.06\\
				\end{tabular} &
				
				\begin{tabular}{@{}l@{}}
					1.52 \ppm 0.01\\
					\textbf{~~~~0.57}\\
				\end{tabular} &
				
				\begin{tabular}{@{}l@{}}
					1.74 \ppm 0.08\\
					0.42 \ppm 0.04\\
				\end{tabular} &
				
				\begin{tabular}{@{}l@{}}
					1.79 \ppm 0.04\\
					0.39 \ppm 0.02\\
				\end{tabular}&
				
				\begin{tabular}{@{}l@{}}
					1.68 \ppm 0.05\\
					0.44 \ppm 0.03\\
				\end{tabular} &
				
				\begin{tabular}{@{}l@{}}
					2.12 \ppm 0.07\\
					0.25 \ppm 0.09\\
				\end{tabular} &
				
				\begin{tabular}{@{}l@{}}
					\textbf{1.44 \ppm 0.01}\\
					0.51 \ppm 0.01\\
				\end{tabular}\\ \hline
				
				
				{Vip} & 11565 & 11.6 & 4.4 & 53 &
				
				\begin{tabular}{@{}l@{}}
					2.60 \ppm 0.03\\
					0.34 \ppm 0.01\\
				\end{tabular} & 
				
				\begin{tabular}{@{}l@{}}
					2.48 \ppm 0.13\\
					\textbf{0.58 \ppm 0.15}\\
				\end{tabular} &
				
				\begin{tabular}{@{}l@{}}
					2.22 \ppm 0.01\\
					\textbf{0.60 \ppm 0.01}\\
				\end{tabular} &
				
				\begin{tabular}{@{}l@{}}
					2.72 \ppm 0.02\\
					0.27 \ppm 0.01\\
				\end{tabular} &
				
				\begin{tabular}{@{}l@{}}
					2.64 \ppm 0.03\\
					0.32 \ppm 0.01\\
				\end{tabular}&
				
				\begin{tabular}{@{}l@{}}
					2.67 \ppm 0.03\\
					0.30 \ppm 0.01\\
				\end{tabular} &
				
				\begin{tabular}{@{}l@{}}
					2.65 \ppm 0.01\\
					0.50 \ppm 0.12 \\
				\end{tabular} &
				
				\begin{tabular}{@{}l@{}}
					\textbf{2.18 \ppm 0.01}\\
					0.54 \ppm 0.01\\
				\end{tabular}\\ \hline
			\end{tabular}
		\vspace{0.15cm}
		\caption{Modularity and log-likelihood comparison on real networks \citep{snapnets,zachary1977information,lusseau2003bottlenose,girvan2002community,adamic2005political}. The value of $k$ is estimated according to Algorithm \ref{alg:2} only if it is not known.  When $k$ is known in advance, it appears in the table as an underlined number. All algorithms are performed on the largest connected component of the network. For each data set, the first row in the score $\mathcal{L}$ Equation~\eqref{eq:loglikelihood}, while the second is $\mathcal{M}$, Equation~\eqref{eq:modularity}. When the standard deviation is not indicated it means that it is below $0.01$.}
		\label{tab:unlabeled}
	\end{sidewaystable}

Given the embedding $\bm{X}$, $5$ iterations of \emph{k-means} are run and the best partition $\bm{\hat{\ell}}$ (in terms of \emph{k-means} is kept) and the scores $\mathcal{M}, \mathcal{L}$ are computed. To keep stochastic of \emph{k-means} into account, this step is repeated for $8$ times and Tables~\ref{tab:unlabeled} reports the results of $\mathcal{M, L}$ in terms of $\texttt{mean} \pm \texttt{standard~deviation}$.

\medskip

\begin{figure}
	\centering
	\includegraphics[width=0.6\columnwidth]{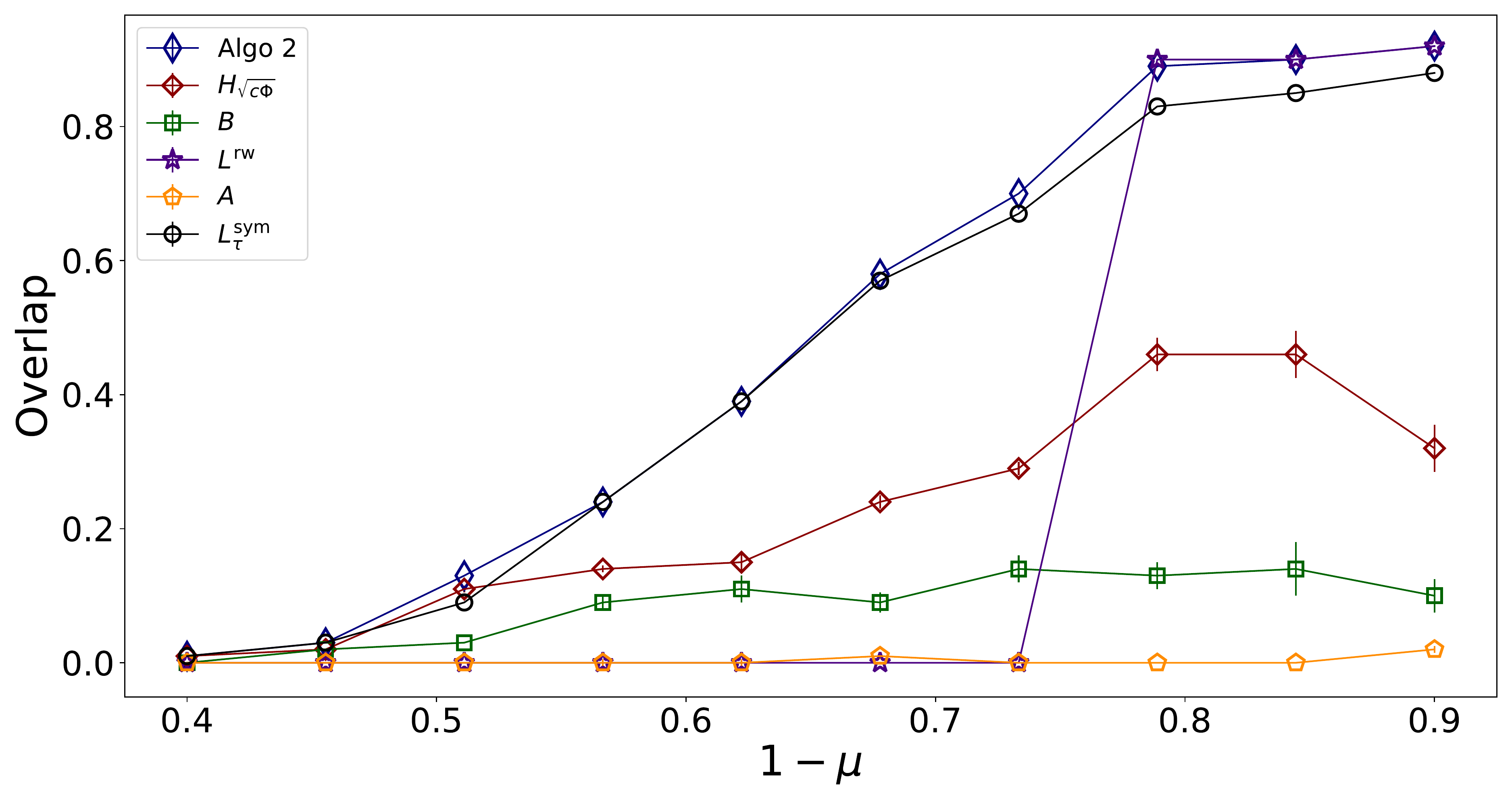}
	\caption{Overlap comparison on different LFR \citep{lancichinetti2008benchmark} benchmark graphs as a function of the mixing parameter $\mu$. Blue sharp diamonds are the result given by Algorithm \ref{alg:2}, red diamonds the algorithm using the Bethe-Hessian of \citep{saade2014spectral}, green squares are for the non-backtracking of \citep{krzakala2013spectral}, yellow pentagons the adjacency matrix, purple stars the random walk Laplacian, black dots are the algorithm of \citep{qin2013regularized}. The parameters of the LFR are $n = 50.000$, $d_{\rm max} = 15~ {\rm log}(n)$, $\hat{d} = 5$, $\tau_1 = -3$, $\tau_2 = 1$, \texttt{min\_class\_size} = $n/8$, \texttt{max\_class\_size} = $n/8$. Variances are taken over $8$ runs of \emph{k-means}. \vspace{-0.7cm}}
	\label{fig:LFR}
\end{figure}

The first score column of Table~\ref{tab:unlabeled} indicates the output of the meta-algorithm Algorithm~\ref{alg:1} for which $\zeta_p$ is estimated from $\rho(B)/s_{R,p}^{\downarrow}(B)$. The second score column provides the output of Algorithm~\ref{alg:2} in which the last step of projection on the hypersphere (Step~4 of Algorithm~\ref{alg:0}) is not performed. Comparing these two columns, it is clear that Algorithm~\ref{alg:2} generally provides much better partitions both in terms of $\mathcal{M}$ and $\mathcal{L}$, as a consequence of $\rho(B)/s_{R,p}^{\downarrow}(B)$ being an inappropriate estimator for $\zeta_p$ in general. The following columns display the results scores obtained by Algorithm~\ref{alg:2} with the projection step (highlighted in cyan), by clustering based on the leading eigenvectors of $A$, by the Bethe-Hessian as per \citep{saade2014spectral}, by the non-backtracking matrix as per \citep{krzakala2013spectral}, by the random walk Laplacian as per \citep{shi2000normalized} and by the symmetric normalized Laplacian of \citep{qin2013regularized}.

The algorithm of \citep{shi2000normalized} based on $L^{\rm rw}$ provides in certain cases very competitive partitions (\emph{e.g.}, in the Wikipedia data set) but is quite unreliable as it may dramatically fail in others (GNutella P2P and Polblogs). Algorithm~\ref{alg:2} without the normalization step provides systematically good partitions, all comparable to those of $L_\tau^{\rm sym}$. This is an evidence that Algorithm~\ref{alg:2} effectively produces a node embedding which is significantly resilient to degree heterogeneity. Finally, the improved version of Algorithm~\ref{alg:2} including the projection step further improves the quality of the partition on most data sets, providing on all data sets but Wikipedia the highest reported modularity and lowest, or second lowest measured value of $\mathcal{L}$.

To test Algorithm~\ref{alg:2} we computed the vector $\bm{\zeta}$ up to machine error precision, that is, the convergence stopping criterion in Subroutine \ref{sub:2} is met when $r$'s update is below machine precision. Notably, due to the non-linearity of the \emph{k-means} step, larger errors in the estimate of the values of $\bm{\zeta}$ lead, within a certain range, to the same partition. The same classification precision can therefore be reached in fewer iterations, thus faster, if needed.

\medskip

\begin{figure}[t!]
	\centering
	\includegraphics[width = \columnwidth]{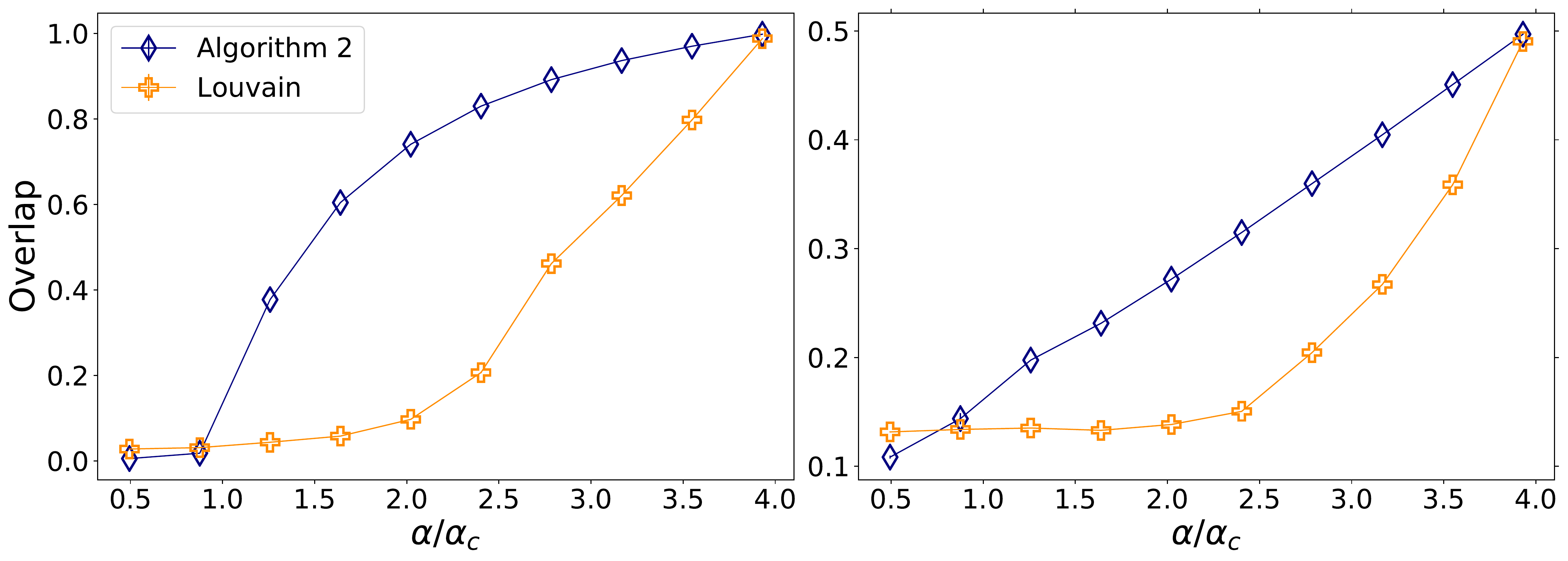}
	\caption{\vspace{-0.5cm}
		Comparison of the overlap (left) and modularity (right) resulting from the label estimation of Algorithm~\ref{alg:2} (blue diamonds) and Louvain algorithm (orange crosses) applied on the giant component of $\mathcal{G}$, as function of the detection hardness. For this simulation, $n = 50.000$, $k = 2$, $\pi \propto \bm{1}_2$, $c = 8$, $c_{\rm out} = 0.05 \to 7$. Averages are taken over 10 samples.\vspace{-0.5cm}}
	\label{fig:vs_Louvain}
\end{figure}

To further test the efficiency of Algorithm~\ref{alg:2}, we compared the clustering performances on some large size Lancichinetti-Fortunato-Radicchi networks \citep{lancichinetti2008benchmark} that notoriously constitute a benchmark to evaluate clustering performance on synthetic graphs with ground truth labels that resemble real-world networks. In this case we have a clear definition of communities (provided by the generative model), hence the performance is evaluated in terms of overlap. Figure~\ref{fig:LFR} shows the performance of the different spectral algorithms considered earlier as a function of the mixing parameters $1-\mu$, the fraction of links connecting nodes in the same community. The result is in complete agreement with the observations in Table~\ref{tab:unlabeled} and Figure~\ref{fig:overlap} which all evidence: i) the importance of choosing a proper parametrization for $H_r$; ii) that Algorithm~\ref{alg:2} is robust on real-world networks.

Let us mention that, comparing the columns ``Alg~\ref{alg:2}$_{\rm wp}$" and ``$H_{\sqrt{c\Phi}}$", it appears evident that Algorithm~\ref{alg:2} indeed achieves good quality embeddings on real-world graphs. In addition, it appeared from our simulations that the projection step on the unitary hypersphere of the rows of $\bm{X}$ significantly helps (both in terms of quality of the partitioning and of the variance of the score) the convergence of \emph{k-means} at a low computational cost.

\medskip

As a side comment, although formally not strictly comparable on even grounds, we evaluated the performances reached by Algorithm~\ref{alg:2} against the popular greedy Louvain method \citep{blondel2008fast} using its \emph{scikit-network} implementation \citep{scikit-learn}. The Louvain method comes with an estimate of the number of communities and relies on a different notion of communities than the one we used. Specifically, it is a hierarchical algorithm which looks for a partition maximizing the modularity, while Algorithm \ref{alg:2} relies on the DC-SBM assumption. Figure~\ref{fig:vs_Louvain} shows that Algorithm~\ref{alg:2} largely outperforms the Louvain method on a DC-SBM graphs, both in terms of overlap (left) and modularity (right), 
except below $\alpha_c$. There, the Louvain method reaches higher modularity, but this is likely an incidental artifact of the modularity optimization constraint \citep{guimera2004modularity}.

The situation of real networks is less straightforward. We indeed observed that when the number of communities estimated by both algorithms is similar, they both produce node partitions with a similar modularity. This is however no longer the case when the estimated $k$ are more distinct (\emph{e.g.}, on GNutella P2P). While we recommend 
our estimation method of $k$ due to its interpretability (that we discussed all along the article), we acknowledge that the Louvain algorithm often provides very competitive outcomes, at a smaller computational cost but at the expense of any theoretical guarantee. Similar considerations can be made in relation to the more recently proposed Leiden method \citep{traag2019louvain}.

\section{Comments and Conclusions}
\label{sec:conclusions}

This article provides a critical analysis of the current state-of-the-art spectral algorithms for community detection in sparse graphs. While these methods were mostly developed in parallel and from differing lines of arguments, we showed that they all relate and can be jointly optimized to tackle community detection in heterogeneous and sparse graphs. This in passing bridges ideas from statistics and physics. Notably, we discovered that the hyperparameters involved in these algorithms, so far chosen statistically, must instead be smartly selected, this selection being a function of the difficulty of the clustering problem.
Besides this fundamental observation, we also showed that there is an efficient (with computational complexity linear in $n$) routine capable of estimating these optimal parameters, which comes from a deep relation between the non-backtracking matrix and the Bayes optimal solution.

\medskip

The necessity in all these methods to adapt the choice of the graph representation matrix to the hardness of the problem at hand is a profound observation, likely not restricted to community detection on sparse graphs. Indeed, the classical spectral algorithms based on Laplacian matrices may all be interpreted as a continuous relaxation of an NP-hard discrete optimization problem, providing good (and sometimes provably optimal) solutions on dense but generally not on sparse networks or data. 
In light of this observation, it may appear that some common optimization problems, beyond clustering, have similarly been devised in the past to perform well in easy scenarios, but would require a task-hardness related update to perform better in harder settings. 

\medskip

Practically, examples of how the results of this article can be generalized to more involved scenarios include the dynamical community detection studied in \citep{dall2020community} on time-evolving graphs and the i.i.d.\@ sparsification (improving cost efficiency) of kernel matrices for spectral clustering \citep{dall2021nishimori}. The latter work shows that the common kernel spectral clustering algorithms can be interpreted as a dense limit of a weighted Bethe-Hessian, that can be conveniently adopted in both the sparse and dense regimes.
Beyond spectral clustering, we envision that a carefully parametrized Bethe-Hessian matrix can be exploited as a powerful alternative to the combinatorial graph Laplacian.

\acks{Couillet's  work  is  supported  by  the  IDEX  GSTATS  DataScienceChair  and  the  MIAI  LargeDATA  Chair  at  University  GrenobleAlpes.  Tremblay's work is partly supported by the CNRS PEPS I3A (Project RW4SPEC) as well as the French National Research Agency in the frameworks of the ANR GraVa (ANR-18-CE40-0005), the "Investissements d’avenir” program (ANR-15-IDEX-02) and the LabEx PERSYVAL (ANR-11-LABX-0025-01). The authors thank Lennart Gulikers and Simon Barthelm\'e for fruitful discussions and the anonymous reviewers for the constructive comments that helped to improve the article.}

\appendix

\section{Existence of a Giant Component in the Sparse DC-SBM}
\label{app:giant}

In this appendix we state and prove Theorem~\ref{th:giant} which provides a necessary and sufficient condition for the existence of a giant component in a multi-class (with unequal sizes) sparse DC-SBM. This theorem is an application of \cite[Theorem~3.1]{bollobas2007phase} to the present DC-SBM setting. Note in particular that the number of nodes in each class $p$ is deterministic and equal to $n\pi_p$. In Bollobas' setting, however, the class of each node $i$ is drawn independently and is equal to $p$ with probability $\pi_p$: $n\pi_p$ is thus only the expected number of nodes in class $p$. Asymptotically, we will see that this of course does not affect the result. To be precise, the generative procedure under analysis in the theorem is the following:
\begin{itemize}
    \item letting $k$ be the number of classes, and $n$ the number of nodes,  we consider a symmetric matrix $C\in\mathbb{R}^{k\times k}$ where $C_{\ell\ell'}>0$ encodes the affinity between classes $\ell$ and $\ell'$; a diagonal matrix $\Pi=\text{diag}(\pi_1,\ldots,\pi_k)$, where $0<\pi_{\ell}<1$ is the expected proportion of nodes in class $\ell$ and $\text{Tr}(\Pi)=1$.
    \item we generate $n$ nodes and, for each node, draw independently the class $p$ to which it belongs with probability $\pi_p$. Also, for each node $i$, we draw independently its intrinsic connectivity variable $\theta_i\in[\theta_{\rm min}, \theta_{\rm max}]$, according to an underlying distribution $\nu(\theta)$ verifying $\mathbb{E}(\theta)=1$ and $\mathbb{E}(\theta^2)=\Phi$. 
    \item For each pair of nodes $(i,j)$, we generate an edge with probability ${\rm min}\left(\theta_i\theta_j\frac{ C_{\ell_i, \ell_j}}{n},1\right)$ where $\ell_i$ is the class of node $i$.
\end{itemize}

\setcounter{theorem}{0}

\begin{theorem}[Percolation threshold in the DC-SBM]
	Consider a graph $\mathcal{G}$ generated according to the above DC-SBM procedure. Suppose that the constant vector is the Perron-Frobenius eigenvector of $C\Pi$ associated to eigenvalue $c>0$. Then, for all large $n$ with high probability, the graph $\mathcal{G}$ has a giant component if and only if $c\Phi > 1$.
	\label{th:giant}
\end{theorem}

\begin{proof} 
In our proof we will use \citep[Theorem 3.1]{bollobas2007phase} which requires to write the graph $\mathcal{G}$ as a graphon. Let us define the ``node'' variable $\bm{x} = \theta\bm{v} \in \mathcal{S}$, with probability distribution $\mu(\bm{x})$, where
\begin{itemize}
	\item $\theta \in \Theta = [\theta_{\rm min}, \theta_{\rm max}]$ is a scalar that encodes the degree heterogeneity, distributed according to $\nu(\cdot)$ and satisfying $\int_{\Theta} d\nu(\theta) = 1$ (normalization), $\mathbb{E}[\theta] = \int_{\Theta} \theta~d\nu(\theta) = 1$ and $\mathbb{E}[\theta^2] = \int_{\Theta} \theta^2~d\nu(\theta) =  \Phi$.
	\item $\bm{v} \in \{ \bm{e}_1,\ldots,\bm{e}_k\}$ with $\bm{e}_p\in\mathbb{R}^{k}$ satisfying ${e}_{p,\ell}=\delta_{\ell p}$; $\bm{v}$ is a $k$-dimensional random vector with law $\mathbb{P}(\bm{v} = \bm{e}_p) = \pi_p$; that is, $\bm{v}$ encodes the class to which node $\bm{x}$ belongs.
	\item The probability density of $\bm{x}$ is denoted by $\mu(\bm{x})$ and is defined on $\mathcal{S}=\cup_{\ell=1}^k \mathcal{S}_{\ell}$ where $\mathcal{S}_{\ell}=[\theta_{\rm min}\bm{e}_{\ell}, \theta_{\rm max}\bm{e}_{\ell}]$. The measure $\mu(\bm{x})$ is equal to $\pi_{\ell}\nu(\theta)$ if $\bm{x}\in\mathcal{S}_{\ell}$ and it is indeed normalized:
	\begin{align*}
	    \int_{\mathcal{S}} d\mu(\bm{x}) = \sum_{\ell = 1}^k \pi_{\ell} \int_{\Theta} d\nu(\theta) = \sum_{\ell = 1}^k \pi_{\ell} = 1.
	\end{align*}
\end{itemize}

Given two such nodes, we next define the kernel $\kappa(\bm{x},\bm{y}) = \bm{x}^TC\bm{y}$.
The kernel $\mathcal{\kappa}$ satisfies the three conditions of \citep[Definition 2.7]{bollobas2007phase} and is a so-called graphical kernel.

With these notations at hand, the generative procedure of the DC-SBM under consideration is equivalent to: drawing $n$ independent values $\{\bm{x}_i\}_{i = 1,\dots, n}$ from $\mu(\cdot)$ and then generating the edges independently according to $p_{ij} = {\rm min}\left(\frac{\kappa(\bm{x}_i,\bm{x}_j)}{n},1\right)$; we thus fall precisely under Bollobas' framework.

\medskip

 In order to use the core argument of \citep[Theorem 3.1]{bollobas2007phase}, we still need to define the linear operator $T_{\kappa}$, an operator on $f:\mathbb{R}^k\to \mathbb{R}$, endowed with the norm $\|f\|^2_2=\int_{\mathcal{S}} f^2(\bm{x})~d\mu(\bm{x})$, as:
\begin{align*}
\forall ~\bm{x} \in \mathcal{S}, \quad (T_kf)(\bm{x}) &= \int_\mathcal{S} \kappa(\bm{x},\bm{y})f(\bm{y})~d\mu(\bm{y})
\end{align*}
The square $2$-norm of this operator, $\Vert T_{\kappa} \Vert_2$, reads
\begin{align}
\Vert T_{\kappa}\Vert_2^2 &= \underrel{\|f\|_2\leq 1}{\rm sup}\int_{\mathcal{S}} (T_{\kappa}f)^2(\bm{x})~d\mu(\bm{x}).
\label{eq:norm_bollo}
\end{align}

According to \citep[Theorem 3.1]{bollobas2007phase}, with high probability, a giant component appears in $\mathcal{G}$ if and only if $\Vert T_{\kappa} \Vert > 1$. We thus are left to evaluating $\Vert T_{\kappa} \Vert$ for the kernel $\bm{x}^TC\bm{y}$: we will show that it equals $c\Phi$, first by showing that $\Vert T_{\kappa} \Vert \leq c\Phi$ and then finding a particular function $f$ for which the bound is attained.

Let $f$ be a function from $\mathbb{R}^k$ to $\mathbb{R}$. We introduce its associated vector $\bm{\omega}_f\in\mathbb{R}^k$ as follows.

\begin{align}
(T_{\kappa}f)(\bm{x}) &= \int_\mathcal{S} \kappa(\bm{x},\bm{y})f(\bm{y})~d\mu(\bm{y}) = \int_\mathcal{S} \bm{x}^TC\bm{y}f(\bm{y})~d\mu(\bm{y}) \nonumber \\
&= \bm{x}^TC\sum_{\ell = 1}^k \pi_{\ell}\bm{e}_{\ell}\int_{{\Theta}}\theta~f(\theta\bm{e}_{\ell})~d\nu(\theta)
\equiv \bm{x}^TC\Pi\bm{\omega}_f,
\label{eq:omega_def}
\end{align}
where $\forall \ell,\;\omega_{f,\ell} = \int_{{\Theta}}\theta~f(\theta\bm{e}_{\ell})~d\nu(\theta)$.  From Equation~\eqref{eq:omega_def}, one thus has:
\begin{align}
\int_\mathcal{S} \left(T_\kappa f\right)^2(\bm{x})d\mu(\bm{x})
&=\int_\mathcal{S} \bm{\omega}^T_f \Pi C \bm{xx}^T C\Pi\bm{\omega}_f ~d\mu(\bm{x})\nonumber\\
&=\Phi \;\bm{\omega}^T_f \Pi C \left(\sum_{\ell=1}^k \pi_{\ell} \bm{e}_{\ell}\bm{e}_{\ell}^T\right) C\Pi\bm{\omega}_f\nonumber\\
&=\Phi \;\bm{\omega}^T_f \Pi C \Pi C\Pi\bm{\omega}_f.
\label{eq:useful}
\end{align}
Injecting the result of Equation~\eqref{eq:useful} into Equation~\eqref{eq:norm_bollo} we obtain
\begin{align}
\Vert T_{\kappa} \Vert_2^2 &= \Phi \underrel{\Vert f \Vert_2\leq 1}{\rm sup}\bm{\omega}^T_f\Pi C \Pi C\Pi \bm{\omega}_f   = \Phi \underrel{\Vert f \Vert_2\leq 1}{\rm sup}\frac{\bm{\omega}^T_f\Pi C \Pi C\Pi \bm{\omega}_f}{\bm{\omega}_f^T\Pi\bm{\omega}_f}\cdot \bm{\omega}_f^T\Pi\bm{\omega}_f \nonumber \\
&\leq \Phi \underrel{\Vert f \Vert_2\leq 1}{\rm sup}\frac{\bm{\omega}^T_f\Pi C \Pi C\Pi \bm{\omega}_f}{\bm{\omega}_f^T\Pi\bm{\omega}_f}~\cdot \underrel{\Vert f \Vert_2\leq 1}{\rm sup} \bm{\omega}_f^T\Pi\bm{\omega}_f
\label{eq:almost_eig}
\end{align}
Analyzing the first element, we can write for $\bm{v}_f = {\Pi^{1/2}}\bm{\omega}_f$:
\begin{align*}
    &\underrel{\Vert f \Vert_2\leq 1}{\rm sup}\frac{\bm{\omega}^T_f\Pi C \Pi C\Pi \bm{\omega}_f}{\bm{\omega}_f^T\Pi\bm{\omega}_f} = \underrel{\Vert f \Vert_2 \leq 1}{\rm sup}\frac{\bm{v}_f^T\Pi^{1/2} C \Pi C\Pi^{1/2} \bm{v}_f}{\bm{v}_f^T\bm{v}_f} \leq \underrel{\bm{v} \in \mathbb{R}^k}{\rm sup}\frac{\bm{v}^T\Pi^{1/2} C \Pi C\Pi^{1/2} \bm{v}}{\bm{v}^T\bm{v}} 
\end{align*}
i.e.:
\begin{align*}
    \underrel{\Vert f \Vert_2\leq 1}{\rm sup}\frac{\bm{\omega}^T_f\Pi C \Pi C\Pi \bm{\omega}_f}{\bm{\omega}_f^T\Pi\bm{\omega}_f}\leq s_1^{\downarrow}(\Pi^{1/2}C\Pi C\Pi^{1/2}) = s_1^{\downarrow}\big((C\Pi)^2\big) = c^2,
\end{align*}
as, by hypothesis, $c$ is the Perron eigenvalue of $C\Pi$, and as such is larger than the modulus of any other eigenvalue: $s_1^{\downarrow}\big((C\Pi)^2\big)$ is indeed $c^2$. 
We can therefore rewrite Equation~\eqref{eq:almost_eig} as
\begin{align*}
\Vert T_{\kappa} \Vert_2^2 \leq c^2\Phi  \underrel{\Vert f \Vert_2\leq 1}{\rm sup} \bm{\omega}_f^T\Pi\bm{\omega}_f.
\end{align*}
Analyzing the right-hand side term, we have
 \begin{align*}
 \bm{\omega}_f^T\Pi\bm{\omega}_f=\sum_{\ell=1}^k \pi_{\ell} {\omega}_{f,\ell}^2=\sum_{\ell=1}^k \pi_{\ell} \left(\int_\Theta \theta f(\theta\bm{e}_{\ell})~d\nu(\theta)\right)^2
 \end{align*}
 which, by Cauchy-Schwartz's inequality is bounded as
 \begin{align*}
 \left(\int_\Theta \theta f(\theta\bm{e}_{\ell})~d\nu(\theta)\right)^2&\leq \left(\int_\Theta \theta^2 d\nu(\theta)\right)
 \left(\int_\Theta  f^2(\theta\bm{e}_{\ell}) ~d\nu(\theta)\right) \leq \Phi,
 \end{align*}
 where in the last step we used the fact that the norm of $f$ is less than or equal to one. We thus obtain that $\underrel{\Vert f \Vert_2\leq 1}{\rm sup}\bm{\omega}_f^T \Pi \bm{\omega}_f \leq \Phi$ and conclude that
 \begin{align}
     \Vert T_{\kappa}\Vert_2^2 \leq (c\Phi)^2.
     \label{eq:lower}
 \end{align}

We are thus left to showing that there exists a function $f$ for which the bound is reached. Let us consider $\bar{f}(\bm{x}) = \Vert\bm{x}\Vert/\sqrt{\Phi}$. It is easy to check that this function has unit norm
\begin{align}
    \Vert \bar{f} \Vert_2^2 = \int_{\mathcal{S}} \left(\frac{\Vert \bm{x} \Vert}{\sqrt{\Phi}}\right)^2~d\mu(\bm{x}) = \frac{1}{\Phi} \int_{\mathcal{S}} \theta^2 \Vert \bm{e}_{\ell} \Vert^2 ~d\mu(\bm{x}) = \frac{1}{\Phi} \int_{\mathcal{S}} \theta^2  ~d\mu(\bm{x}) = 1.
\end{align}
Furthermore, observe that $\bm{\omega}_{\bar{f}} = \sqrt{\Phi} \bm{1}_k$. Then, we have
\begin{align}
   \Vert T_{\kappa}\Vert_2^2 = \Phi \underrel{\Vert f \Vert_2\leq 1}{\rm sup} \bm{\omega}_f^T \Pi C \Pi C \Pi \bm{\omega}_{f} \geq \Phi \bm{\omega}_{\bar{f}}^T \Pi C \Pi C \Pi \bm{\omega}_{\bar{f}} = \Phi^2 \bm{1}_k^T \Pi (C\Pi)^2 \bm{1}_k = (c\Phi)^2
   \label{eq:upper}
    \end{align}
as, by hypothesis, we suppose that $C\Pi \bm{1}_k=c \bm{1}_k$.
    
Combining Equations~\eqref{eq:lower} and \eqref{eq:upper}, we find that $\Vert T_{\kappa} \Vert = c\Phi$ so that, from \citep[Theorem 3.1]{bollobas2007phase}, we conclude that the percolation transition is at $c\Phi = 1$ and a giant component exists if and only if $c\Phi > 1$.

\end{proof}

\section{Definition of $\zeta$ on the Sparse DC-SBM}
\label{app:zeta_p}

In this Appendix we show that for a graph generated from the sparse DC-SBM model, defined in Equation~\eqref{eq:DC-SBM}, the two following definitions of $\zeta_p^{(j)}$ (where  $j$ is the giant component) are indeed equivalent with high probability for $p \geq 2$ if $\zeta_p^{(j)}$ exists:

\begin{align}
\label{eq:zeta_equiv:1}
\zeta_p^{(j)} &= \underrel{r\geq1}{\rm min}\{r : s_p^{\uparrow}(H^{(j)}_r) = 0\}, \\
\label{eq:zeta_equiv:2}
\zeta_p^{(j)} &= \underrel{r>1}{\rm min}\{r : s_p^{\uparrow}(H_r) = 0\}.
\end{align}
We will further discuss that, with high probability, $\zeta_p^{(j)}>1$ exists only for the giant component.

\medskip

This appendix is structured as follows. First we enunciate two lemmas and one corollary to state that when an arbitrary connected graph has at most one loop, the spectrum of its associated non-backtracking matrix $B$ does not have real eigenvalues that are larger than one in modulus.

We then proceed by arguing that with high probability, for large DC-SBM graphs, the small connected components do not have more than one loop, implying that all the real eigenvalues of $B$ that are larger than one in modulus come from the giant component. Finally we show that this last statement, implies that the two definitions of $\zeta_p^{(j)}$ given in Equations (\ref{eq:zeta_equiv:1},\ref{eq:zeta_equiv:2}) are indeed equivalent with high probability.

\begin{lem}[Spectrum of $B$ on a tree]
\label{th:1}
Let $\mathcal{T}(\mathcal{V},\mathcal{E})$ be a tree. Then, all the eigenvalues of its associated non-backtracking matrix $B(\mathcal{T})$, as defined in Equation~\eqref{eq:B}, are equal to zero.
\end{lem}

\begin{lem}[Spectrum of $\mathcal{G}$ plus a node]
\label{th:2}
Let $\mathcal{G}'$ be a graph obtained by adding one node and one edge to the graph $\mathcal{G}$, i.e. for $i \notin \mathcal{V}$ and an arbitrary $j \in \mathcal{V}$, $\mathcal{V}' = \mathcal{V} \cup \{i\}$ and $\mathcal{E}' = \mathcal{E} \cup (ij)$. Then, all the non-zero eigenvalues of $B(\mathcal{G}')$ are equal to the non-zero eigenvalues of $B(\mathcal{G})$, where $B(\mathcal{G})$ is the non-backtracking matrix defined in Equation~\eqref{eq:B} of the graph $\mathcal{G}$.
\end{lem}

\begin{cor}[Spectral radius of a connected graph with $|\mathcal{E}| = |\mathcal{V}|$]
	\label{cor:1}
Let $\mathcal{G}(\mathcal{V},\mathcal{E})$ be a connected graph with $|\mathcal{E}| = |\mathcal{V}|$. Then $\rho(B) = 1$, where $B$ is defined in Equation~\eqref{eq:B} and $\rho(\cdot)$ is the spectral radius.
\end{cor}

\begin{proof}[Lemma \ref{th:1}]
Consider an arbitrary vector $\bm{g}\in\mathbb{R}^{2|\mathcal{E}|}$ and define  $\bm{g}^{(m)}{\in\mathbb{R}^{2|\mathcal{E}|}}$ as 
\begin{align}
{g}_{ij}^{(m)} = \sum_{(k\ell)~:~ d(ij, k\ell) = m} g_{k\ell}.
\end{align}
The notation $d(ij,k\ell) = m$ indicates that there exists a non-backtracking path from the edge $(ij)$ to the edge $(k\ell)$ of length $m$. From a straightforward calculation one obtains
\begin{align}
(B\bm{g}^{(m)})_{ij} =  \sum_{\ell \in \partial j\setminus i}~\sum_{(kq):d(j\ell,kq) = m} g_{kq} = g_{ij}^{(m+1)}.
\end{align} 
For all trees, for any two edges we have that $d(ij,k\ell) \leq n-1$, so there is a value  of $m_{\rm  c}\leq n-1$, which represents the maximal distance between any two directed edges, such that, for all vectors $\bm{g}$,
\begin{align}
B\bm{g}^{(m_{\rm c})} = 0.
\end{align}
This relation comes from the fact that, by  definition of $m_{\rm c}$, no two edges are at a distance equal to $m_{\rm c}+1$, so $\{(k\ell: d(ij,k\ell) = m_{\rm c} + 1\} = \emptyset$ for any edge $(ij)$.
Now, let us consider $\bm{g}$ to be an eigenvector of $B$, such that $B\bm{g} = \gamma \bm{g}$. Then we can write
\begin{align}
0 = B\bm{g}^{(m_{\rm c})} = B^2\bm{g}^{(m_{\rm c}-1)} = \dots = B^{m_{\rm c}-1}\bm{g} = \gamma^{m_{\rm c}-1}\bm{g}.
\end{align}
Thus concluding that any eigenvector $\bm{g}$ of $B$ is associated to eigenvalue zero. Note that in other words, this means that $B$ is nilpotent.
\end{proof}

\begin{proof}[Lemma \ref{th:2}]
Let $i$ be the newly added node and $j$ the node in $\mathcal{V}$ to which $i$ is attached. The matrix $B(\mathcal{G}')$ can be written by adding to the matrix $B(\mathcal{G})$ two rows and two columns corresponding to the directed edges $(ij)$ and $(ji)$. We introduce the notation $\mathds{1}^{(\bullet j)}\in \mathbb{R}^{2|\mathcal{E}|}$. Its element-wise definition reads for all $(yx) \in \mathcal{E}_d$ (the set of directed edges of $\mathcal{G}$) as $\mathds{1}_{yx}^{(\bullet j)} = \delta_{xj}$. Similarly, we define  $\mathds{1}^{(j\bullet)}\in \mathbb{R}^{2|\mathcal{E}|}$ with $\mathds{1}^{(j\bullet)}_{yx} = \delta_{yj}$. Denote with $M = (\bm{\mathds{1}}^{(j\bullet)}, \bm{0}_{2|\mathcal{E}|})\in \{0,1\}^{2|\mathcal{E}|\times 2}$ and $M' = (\bm{0}_{2|\mathcal{E}|},\bm{\mathds{1}}^{(\bullet j)})\in \{0,1\}^{2|\mathcal{E}|\times 2}$, where $\bm{0}_{2|\mathcal{E}|}$ is the null vector of size $2|\mathcal{E}|$.  The matrix $B(\mathcal{G}')$ can be written as:
\begin{align*}
    B(\mathcal{G}')= 
\begin{pmatrix}
B(\mathcal{G})& M' \\
M^T & 0_{2\times 2}
\end{pmatrix}
\end{align*}
We now look for the non-zero eigenvalues of $\gamma$ of $B(\mathcal{G}')$. Recall that $I_n$ is the $n\times n$ identity matrix. Consider $\gamma\in\mathbb{C}^*$. Using a block matrix determinantal formula one has:
\begin{align*}
    {\rm det}\left(B(\mathcal{G}')-\gamma I_{2|\mathcal{E}'|}\right) = {\rm det}(-\gamma I_2){\rm det}\left(B(\mathcal{G}) - \gamma I_{2|\mathcal{E}|} {+} \frac{1}{\gamma}M'M^T\right).
\end{align*}
It is straightforward to check that $M'M^T = 0_{2|\mathcal{E}|\times 2|\mathcal{E}|}$, thus we have ${\rm det}\left(B(\mathcal{G}')-\gamma I_{2|\mathcal{E}'|}\right) = {\gamma^2}{\rm det}\left(B(\mathcal{G})-\gamma I_{2|\mathcal{E}|}\right)$. {A non-zero eigenvalue of $B(\mathcal{G}')$ (cancelling the determinant) is thus necessarily also an eigenvalue of $B(\mathcal{G})$, ending the proof.}
\end{proof}

\begin{proof}[Corollary \ref{cor:1}]
A connected graph with $|\mathcal{E}| = |\mathcal{V}|$ can be obtained by adding an edge between any two nodes of a particular tree defined on $\mathcal{\mathcal{V}}$. The graph $\mathcal{G}$ thus contains only one loop. We now apply Lemma \ref{th:2} "backwards", i.e., removing leaves from $\mathcal{G}$ without affecting the non-zero eigenvalues of $B$. 

By iteratively removing all leaves, the graph $\mathcal{G}$ reduces to a loop. We are thus left to prove that $\rho(B) = 1$ on a loop. It is straightforward to check that on any $d$-regular graph (all nodes having $d$ neighbours), the vector $\bm{1}_{2|\mathcal{E}|}$ is the Perron-Frobenius  eigenvector of $B$ with eigenvalue equal to $d-1$. A loop is a $d$-regular graph with $d = 2$, hence the result. 
\end{proof}

With the results of the former two Lemmas, we proceed to argue that all the eigenvalues of $B$ that are larger than one in modulus come from the giant component. Since the graph is disconnected, $S(B) = \bigcup_{j=1}^{n_{\rm CC}} {S}\left(B^{(j)}\right)$, i.e., each connected component contributes independently to the eigenvalues of $B$. The expected size of the small connected components grows as\footnote{{Indeed, the kernel $\kappa$ defined in the proof of Theorem~\ref{th:giant} is irreducible in the sense of \citep[Def.~2.10 ]{bollobas2007phase}, as the nodes cannot be split into two parts that have no chance of being connected (all the entries of $C$ and all $\pi_i$ are supposed strictly positive). One can thus apply \citep[Thm.~3.12]{bollobas2007phase} stating that the small components are with high probability of order $\log{n}$.}} ${\rm log}(n)$. {We claim that this implies that} the probability that a small connected component contains two or more loops is $o_n(1)$, thus tending to zero as $n$ grows to infinity. Applying Lemmas \ref{th:1} (for zero loops) and Corollary \ref{cor:1} (for one loop), we conclude that, with high probability all the real eigenvalues of $B$ larger than one in modulus will come from the giant component. From the Ihara-Bass formula \citep{terras2010zeta}, for all the eigenvalues $\gamma \neq \pm 1$, $\gamma \in S(B)\iff {\rm det}[H_{\gamma}] = 0$. From the definition of $\zeta_{p\geq 2}^{(j)}$ we gave in Equation~\eqref{eq:zeta_j}, we have $\rm{det}[H_{\zeta_{p\geq 2}^{(j)}}] = 0$, thus all the $\zeta_p^{(j)} > 1$ are in the spectrum of $B$. We may conclude that, if $\exists j$ such that $\zeta_{p\geq 2}^{(j)}>1$, then with high probability $j$ is the giant component. 
We now conclude showing that this statement implies the equivalence between Equations~\eqref{eq:zeta_equiv:1} and \eqref{eq:zeta_equiv:2}.

\medskip

Thanks to the properties we discussed in Section \ref{sec:connection}, for $p>1$, then certainly $\zeta^{(j)}_p>1$, if it exists. We denote with $j$ the giant component and we want to prove that for $r,p >1$
\begin{align}
s_p^{\uparrow}(H_r^{(j)}) = s_p^{\uparrow}(H_r)
\end{align}
Consider $\zeta_p^{(j)} > 1$ such that $s_p^{\uparrow}(H^{(j)}_{\zeta_p^{(j)}}) = 0$, then certainly there exists\footnote{{As $S(H_r)=\cup_{j=1}^{n_{\rm CC}} S\left(H_r^{(j)}\right)$.}} $q\geq p$ such that $s_q^{\uparrow}(H_{\zeta_p^{(j)}}) = 0$ and so it is enough to show that $p=q$. We proceed with a proof by contradiction. Suppose that $p \neq q$, then there exists $j'\neq j$ such that $s_1^{\uparrow}\left(H^{(j')}_{\zeta_p^{(j)}}\right) \leq 0$. Applying Gershgorin circle theorem, it is easy to show that for $r > d_{\rm max}-1$, the matrix $H_r$ is positive definite. Consequently, there exists $r\geq \zeta_p^{(j)}>1$ such that $s_1^{\uparrow}(H_r^{(j')}) = 0$. From the Ihara-Bass formula, $r$ is thus  in $S(B)$, so there is a real eigenvalue of $B$ larger than one not coming from the giant component. This is in contradiction with what stated above, thus concluding our argument.

\section{The Eigenvalues of $T$}
\label{app:T}

We here want to show that $s_p^{\downarrow}(T) = \frac{\nu_{p+1}}{c}$ for $1 \leq p < k$. Consider the  following equivalent identities for the matrix $T = \frac{\Pi C}{c} - \Pi \bm{1}_k\bm{1}_k^T$:

\begin{align*}
    \left(\frac{\Pi C}{c} - \Pi \bm{1}_k\bm{1}_k^T\right)\bm{a}_p &= s_p^{\downarrow}(T)\bm{a}_p \\
    \left(\frac{C\Pi}{c} -  \bm{1}_k\bm{1}_k^T\Pi \right) \bm{b}_p &= s_p^{\downarrow}(T) \bm{b}_p; \quad\quad \bm{b}_p = \Pi^{-1}\bm{a}_p \\
\end{align*}
Since $C\Pi\bm{1} = c\bm{1}_k$ and $\bm{1}_k^T\Pi\bm{1}_k = 1$, then $\bm{b}_k = \bm{1}_k$. By introducing $\bm{d}_p = \Pi^{1/2}\bm{b}_p$, we can write:
\begin{align*}
    \left(\frac{\Pi^{1/2}C\Pi^{1/2}}{c} - \Pi^{1/2}\bm{b}_k\bm{b}_k^T\Pi^{1/2}\right)\bm{d}_p &= s_p^{\downarrow}(T)\bm{d}_p \\
    \left(\frac{\Pi^{1/2}C\Pi^{1/2}}{c} - \bm{d}_k\bm{d}_k^T\right)\bm{d}_p &= s_p^{\downarrow}(T)\bm{d}_p 
\end{align*}
Given that the $\bm{d}_k$ are eigenvectors of a symmetric matrix, then $\bm{d}_p^T\bm{d}_q = \delta_{pq}$, so, for $p < k$
\begin{align*}
    \frac{\Pi^{1/2}C\Pi^{1/2}}{c} \bm{d}_p &= s_p^{\downarrow}(T)\bm{d}_p \\
    \frac{C\Pi}{c}\bm{b}_p &= s_p^{\downarrow}(T)\bm{b}_p
\end{align*}
From this last equation the result follows directly.

\section{The Excited States of the Ising Hamiltonian on $\mathcal{G}$}
\label{app:mapping_to_Ising}

In this section we give an explicit connection between the problem described in Section~\ref{subsec:connection.ising} and its statistical physics analogue. Consider the graph $\mathcal{G}$ obtained from the $k$-class DC-SBM model as per Equation \eqref{eq:DC-SBM}. To study the stability of the configurations of $\langle\bm{\sigma}\rangle$ as a function of $r$, one needs to compute the free energy $F$, defined in Equation~\eqref{eq:free} (which we recall is a function of the moments of the Boltzmann distribution $\mu(\bm{\sigma})$):
\begin{align}
\mu(\bm{\sigma}) &= \frac{1}{Z}e^{-\mathcal{H}(\bm{\sigma})} \label{eq:Boltzmann} \\
F &= -{\rm log}~Z = \sum_{\bm{\sigma}} \mu(\bm{\sigma})[\mathcal{H}(\bm{\sigma})  + {\rm log}~\mu(\bm{\sigma})]. \label{eq:free}
\end{align}
To evaluate $F$, a common way to proceed is to exploit the \emph{Bethe} approximation that is exact on a tree and is thus relevant on sparse graphs. By denoting with $\mu_{ij}(\cdot)$ and $\mu_i(\cdot)$ the edge and the node marginals of $\mu(\cdot)$, the Bethe free energy is defined as:
\begin{align}
\mu^{\rm B}(\bm{\sigma};\bm{m},\bm{\chi}) &= \frac{\prod_{(ij)\in\mathcal{E}} \mu_{ij}^{\rm B}(\sigma_i,\sigma_j;\bm{m},\bm{\chi})}{\prod_{i \in \mathcal{V}} \left[\mu_i^{\rm B}(\sigma_i;\bm{m},\bm{\chi})\right]^{d_i-1}} \label{eq:factor}\\
F_{\rm Bethe}(\bm{m},\bm{\chi}) &= \sum_{\bm{\sigma}} \mu^{\rm B}(\bm{\sigma};\bm{m},\bm{\chi})\big[\mathcal{H}(\bm{\sigma}) + {\rm log}~\mu^{\rm B}(\bm{\sigma};\bm{m},\bm{\chi})\big].
\end{align}
By a direct computation, one can verify that $F_{\rm Bethe}(\bm{m},\bm{\chi}) = F + {\rm D_{KL}}\big(\mu^{\rm B}(\bm{\sigma};\bm{m},\bm{\chi})
\Vert \mu(\bm{\sigma})\big)$, where $D_{\rm KL}(\cdot \Vert \cdot)$ is the Kullback-Leibler divergence which is always greater or equal to zero. In order to find the best estimate of the exact free energy, one needs to find the minimum of $F_{\rm Bethe}$ with respect to $\bm{m}$, $\bm{\chi}$. At large $r$ one sees that $F_{\rm Bethe}$ has a unique minimum in $\bm{m} = 0$, $\bm{\chi} = \frac{1}{r}\bm{1}_{2|\mathcal{E}|}$, whereas for $r$ small enough, there appear multiple minima and $\bm{m} = 0$ is a saddle point. To quantitatively study this behavior, one needs to consider the Hessian matrix of the Bethe free energy at the  paramagnetic point, i.e.:
\begin{equation}
\left.  \frac{\partial^2F_{\rm Bethe}(\bm{m},\bm{\chi})}{\partial m_i \partial m_j}\right|_{\bm{m} = \bm{0};\bm{\chi} = \frac{1}{r}\bm{1}_{2|\mathcal{E}|}} = \frac{(H_r)_{ij}}{r^2-1}.
\end{equation}
The eigenvectors corresponding to the negative eigenvalues of the Bethe-Hessian matrix $H_r$ represent the directions along which the paramagnetic phase becomes unstable and they will have a non-trivial structure.

We now detail an approach inspired from \citep{suchecki2006ising} but adapted to the Bethe approximation to study the stability of the paramagnetic phase. The following equations hold:
\begin{align}
\mu_i^{\rm B}(\sigma_i) &= \frac{1}{z_i}\prod_{k \in \partial i} \xi_{ki}(\sigma_i)\label{eq:bp1}\\
\xi_{ji}(\sigma_i) &\propto \sum_{\bm{\sigma}_j}e^{{\rm ath}\left(\frac{1}{r}\right)\sigma_i\sigma_j}\prod_{k \in \partial j \setminus i}\xi_{kj}(\sigma_j)\label{eq:bp2}.
\end{align}
It is possible to verify that by plugging the ansatz of Equations (\ref{eq:bp1},\ref{eq:bp2}), one recovers the factorization of Equation \eqref{eq:factor}.
We define the following change of variable $\xi_{ji}(\sigma_i) = e^{h_{ji}\sigma_i}$, resulting in $h_{ji} = \frac{1}{2}{\rm log}\frac{\xi_{ji}(+1)}{\xi_{ji}(-1)}$. This leads to the following expressions:
\begin{align}
h_{ji} &= {\rm ath}\left[\frac{1}{r}{\rm th}\left(B^T\bm{h}\right)_{ji}\right] \label{eq:BP_ising_f} \\
m_i &= {\rm th}\left[\sum_{j \in \partial i}h_{ji}\right]. \label{eq:BP_ising_m}
\end{align}
We next linearize Equations~\eqref{eq:BP_ising_f} and \eqref{eq:BP_ising_m} at the paramagnetic point, i.e., for $\bm{h} \to 0$ in the high temperature regime (large $r$). For mathematical convenience we here assume the regime where $c \gg 1$ and $r$ of the order of $c$ or larger. We will comment in the end on how the results we obtain still hold in the sparse regime in which $c = O_n(1)$.
\begin{align}
m_i &= \sum_{j \in \partial i} h_{ji} = \frac{1}{r}\sum_{j \in \mathcal{V}} \left(B^T\bm{h}\right)_{ji}  = \frac{1}{r}\sum_{j,k,m \in \mathcal{V}} A_{ij}A_{km}\delta_{jm}(1-\delta_{ik})h_{km} \nonumber \\
&= \frac{1}{r}\left[\sum_{j,k \in \mathcal{V}} A_{ij}A_{jk}h_{kj} - \sum_{j \in\mathcal{V}} A_{ij}h_{ij}\right] = \frac{1}{r}\sum_{j \in\mathcal{V}}A_{ij}(m_j-h_{ij}).
\end{align}
Further developing the second term of the summation
\begin{align}
\sum_{j \in \mathcal{V}} A_{ij}h_{ij}=\sum_{j \in\partial i} h_{ij} = \frac{1}{r}\sum_{j \in \partial i}\left(B^T\bm{h}\right)_{ij} = \frac{(d_i-1)m_i}{r},
\end{align}
we obtain
\begin{align}
m_i = \frac{1}{r}\left[\sum_{j \in\mathcal{V}} A_{ij}m_j - \frac{(d_i-1)m_i}{r}\right] = \frac{1}{r}\sum_{j \in \mathcal{V}}A_{ij}m_j + O_n\left(\frac{1}{r^2}\right).  \label{eq:self}
\end{align}
Equation~\eqref{eq:self} gives a self consistent relation between the magnetizations of the stable (or metastable) configurations. We study the magnetizations taking an average over the realizations of $A$ as proposed in \citep{suchecki2006ising}
\begin{equation}
\mathbb{E}[m_i] \approx \frac{1}{r}\sum_{j \in \mathcal{V}} \theta_i\theta_j\frac{C_{\ell_i,\ell_j}}{n}\mathbb{E}[m_j].
\end{equation}
Defining $s_a = \frac{1}{n\pi_a}\sum_{i \in \mathcal{V}_a} \theta_i\mathbb{E}[m_i]$, where $\mathcal{V}_a = \{i \in \mathcal{V} : \ell_i = a \}$, gives
\begin{align}
s_a &= \frac{1}{n\pi_a}\sum_{i \in \mathcal{V}_a} \theta_i\mathbb{E}[m_i] \approx \frac{1}{rn^2\pi_a}\sum_{b = 1}^k \sum_{i \in \mathcal{V}_a}\sum_{j \in \mathcal{V}_b}\theta_i^2\theta_jC_{ab}\mathbb{E}[m_j] \nonumber \\
&= \frac{1}{rn\pi_a}\sum_{b = 1}^k\sum_{i \in \mathcal{V}_a} \theta_i^2 C_{ab}s_b \pi_b = \frac{\Phi}{r}\sum_{b = 1}^k (C\Pi)_{ab}s_b.
\end{align}
We thus obtain the relation controlling the stability of the solutions:
\begin{equation}
C\Pi\bm{s} = \frac{r}{\Phi}\bm{s} = \nu_p\bm{s}, \label{eq:transitions}
\end{equation}
The transition temperatures are hence given by $r = \nu_p\Phi$, as expected. For $r > c\Phi$ Equation~\eqref{eq:transitions} only admits the trivial solution $\bm{s} = 0$ which represents the paramagnetic configuration. At $r = c\Phi$ there is a paramagnetic-ferromagnetic transition, consistent with the result of \citep{leone2002ferromagnetic}. For smaller values of $r$ there appear other directions (aligned to the eigenvectors of $C\Pi$) along which the Bethe free energy has local minima.

The main result of this appendix is thus to argue that (i) there are only $k$ directions along which the paramagnetic phase can get  unstable, hence $H_r$ can have at most $k$ negative eigenvalues, so that $s_{k+1}^{\uparrow}(H_r) \geq 0$ for $r > 1$; (ii) the directions of instability precisely correspond to the eigenvectors of the matrix $C\Pi$.

\medskip

To conclude, we comment the case $c = O_n(1)$. 
We just detailed an approach to understand and interpret the problem at hand with tools taken from statistical physics. For mathematical convenience we had to assume the average degree $c$ to be rather large and we were able to predict the exact positions of the transition temperatures, which coincide with the largest eigenvalues of the non-backtracking matrix. Our approach in determining the positions of these eigenvalues is non-rigorous and does not intend to substitute the existing proofs, but rather to give a physical intuition of the problem at hand. In particular, the spectrum of $B$ has been rigorously studied in   \citep{gulikers2017spectral,bordenave2015non} in the regime $c = O_n(1)$. The conclusion above is not altered: the matrix $B$ has $k$ isolated eigenvalues, so only the $k$ smallest eigenvalues of the Bethe-Hessian matrix can become negative, in correspondence to  the emergence of new local minima in the Bethe free energy profile. According to our earlier argument, we expect these minima (represented by the eigenvectors of $H_r$) to be correlated with the class structure of $A$.

\section{Supporting Arguments for Claim \ref{prop:1}}
\label{app:proof}

To support Claim~\ref{prop:1}, we will rely on the following three intermediary results that will be proved subsequently.

\begin{lem}
\label{lem:1}
Let $D$ and $A$ be the degree and adjacency matrices of any graph of size $n$. Let $r>1$ and $p$ an integer between $2$ and $n$. If $s_p^{\uparrow}(D-rA) < 0$ then
\begin{equation*}
s_p^{\downarrow}\left(L_{-s_p^{\uparrow}(D-rA)}^{\rm rw}\right) = \frac{1}{r}
\end{equation*}
and both eigenvalues $s_p^{\uparrow}(D-rA)$ and $1/r$ share the same eigenvector.
\end{lem}
The fact that the matrix $L_{-s_p^{\uparrow}(D-rA)}^{\rm rw}$ has an eigenvalue equal to $1/r$ comes easily by construction of $L_{-s_p^{\uparrow}(D-rA)}^{\rm rw}$. The main result of Lemma~\ref{lem:1} is to state that this eigenvalue is the $p$-th largest. Note that the condition $s_p^{\uparrow}(D-rA) < 0$ could be loosened, but we don't need a stronger result in the following.

\begin{lem}
	\label{lem:3}
	Let $D$ and $A$ be the degree and adjacency matrices of any arbitrary graph. Let $p$ be an integer $\geq2$.
	Suppose that there exists $r_p>1$ such that i/~$\forall ~r\geq r_p$ the eigenvalue $s_p^{\uparrow}(D-r A)$ is simple, and ii/~$s_p^{\uparrow}(D-r_pA)<0$. Then:
	\begin{align*}
	\forall~r\geq r_p, \quad \partial_r s_p^{\uparrow}(D-rA) < 0.
	\end{align*}
	In words, once the function $s_p^{\uparrow}(D-rA)$ becomes negative, it is strictly decreasing.
\end{lem}

The eigenvalue simplicity assumption of this lemma is technical and enables us to properly define and manipulate the derivative of the $p$-th smallest eigenvalue. In case of multiplicity, the tools involved are more complicated~\citep[see for instance][]{greenbaum2019first} and not included here. Note that according to Claim \ref{claim:1} this assumption is verified for a network generated from the DC-SBM because on a large random graph with independent entries all eigenvalues are simple with probability one.
\begin{lem}
		\label{lem:2}
Let $\mathcal{G}(\mathcal{V},\mathcal{E})$ be a DC-SBM graph generated according to Equation~\eqref{eq:DC-SBM}, and $D$ and $A$ its degree and adjacency matrices respectively.
Consider $2 \leq p \leq k$ and let $\zeta_p$ be defined as per Equation~\eqref{eq:zeta}. Let $r \geq \zeta_p$. If $s_p^{\uparrow}(D-rA)<0$ is isolated with associated eigenvector $\bm{x}_p$, 
then $\bm{x}_p$ is an eigenvector of $L^{\rm rw}_{-s_p^{\uparrow}(D-rA)}$, whose corresponding eigenvalue is also isolated.
Otherwise, for $1 \leq p \leq n$ and $r>1$, if $s_p^{\uparrow}(D - rA)$ belongs to the bulk of non informative eigenvalues of $D-rA$, then the corresponding eigenvalue of $L^{\rm rw}_{-s_p^{\uparrow}(D-rA)}$ also belongs to the bulk of $L^{\rm rw}_{-s_p^{\uparrow}(D-rA)}$.
\end{lem}

Based on Lemmas \ref{lem:1}-\ref{lem:2} and Claim \ref{claim:1} we now proceed to the justification of Claim \ref{prop:1}, that we recall here for convenience.
\setcounter{claim}{1}
\begin{claim}
	Consider the graph $\mathcal{G}(\mathcal{V},\mathcal{E})$, built on a sparse DC-SBM as per Equation~\eqref{eq:DC-SBM} with $k$ communities. Let $\zeta_p$ for $2 \leq p \leq k$ be defined as per Equation~\eqref{eq:zeta} (imposing $\zeta_1 = 1$) and $\tau\in\mathbb{R}$ be such that $\zeta_p^2-1 \leq \tau \leq c\Phi - 1$. Then, under Assumption~\ref{ass:cpi}, for all large $n$ with high probability, the $p$ largest eigenvalues of the matrix $L_{\tau}^{\rm rw}$ are isolated. In particular, 
	\begin{align*}
	s_p^{\downarrow}(L^{\rm rw}_{\zeta_p^2-1}) = \frac{1}{\zeta_p}.
	\end{align*}
\end{claim}
For simplicity, will use the notation $s_p^{\uparrow}(D-rA) = s_p(r)$.
The first part of Claim \ref{prop:1} asserts that for $\zeta_p^2-1 \leq \tau \leq c\Phi-1$ the $p$ largest eigenvalues of the matrix $L_{\tau}^{\rm rw}$ are isolated. According to Claim~\ref{claim:1}, for $\zeta_p \leq r \leq \sqrt{c\Phi}$ the eigenvalue $s_p(r)$ is isolated. 
Thanks to Lemma~\ref{lem:3}, since $s_p(\zeta_p) < 0$ we have: $\partial_r s_p(r) < 0$ for all $r$ in the interval $\zeta_p \leq r \leq \sqrt{c\Phi}$, implying $s_p^{\uparrow}(D-rA) < 0$ for all $r$ in the interval $\zeta_p \leq r \leq \sqrt{c\Phi}$. The hypotheses of Lemma~\ref{lem:1} and Lemma~\ref{lem:2} are therefore verified.
We can then assert that the eigenvalue $1/r$ is the $p$-th largest of the matrix $L_{-s_p^{\uparrow}(D-rA)}^{\rm rw}$ and it is isolated. 

Further exploiting Lemma~\ref{lem:3}, letting $\tau(r) = -s_p^{\uparrow}(D -rA)$, we have $\partial_r \tau(r) > 0$ for all $r$ in the interval $\zeta_p \leq r \leq \sqrt{c\Phi}$. The function $\tau(r)$ is thus bijective and increasing on this interval. Since the eigenvalue $s_p^{\uparrow}(D-rA)$ is isolated for all $\zeta_p \leq r \leq \sqrt{c\Phi}$, the corresponding eigenvalue of $L_{\tau(r)}^{\rm rw}$ equal to $1/r$ is isolated for 
\begin{align*}
\zeta_p^2 - 1 = -s_p(\zeta_p)\leq \tau \leq -s_{p}(\sqrt{c\Phi}).
\end{align*}
Note now that, for $1 \leq p \leq k$, $s_p^{\uparrow}(D-\sqrt{c\Phi}A) \leq s_{k+1}^{\uparrow}(D - \sqrt{c\Phi}A) = c\Phi -1$. This  implies that for $\zeta_p^2-1\leq \tau \leq c\Phi - 1$,  the top $p$ eigenvalues of $L_{\tau}^{\rm rw}$ are certainly isolated.

For the particular case where $r = \zeta_p$, by definition of $\zeta_p$, $-s_p^{\uparrow}(D-\zeta_pA) = \zeta_p^2-1 > 0$ and the  result is straightforwardly obtained by applying Lemma \ref{lem:1}, concluding our argument.

\medskip

We now proceed to the proof of the three lemmas. For the sake of clarity we here enunciate the Courant-Fischer theorem that will be of fundamental use in the sequel.

\begin{theorem}\emph{\citep[Courant-Fischer, see for instance][]{bhatia2013matrix}}
	Let $M\in \mathbb{C}^{n\times n}$ be a Hermitian matrix and $U$ a vector subspace of $\mathbb{C}^n$. Then,
	\begin{align*}
	s_p^{\uparrow}(M) &= \underrel{U : {\rm dim}(U) = p}{\rm min} ~\underrel{\bm{z} \in U, \bm{z}\neq 0}{\rm max} \frac{\bm{z}^TM\bm{z}}{\bm{z}^T\bm{z}} \\
	s_p^{\uparrow}(M) &= \underrel{U : {\rm dim}(U) = n-p+1}{\rm max} ~\underrel{\bm{z} \in U, \bm{z}\neq 0}{\rm min} \frac{\bm{z}^TM\bm{z}}{\bm{z}^T\bm{z}}.
	\end{align*}
\end{theorem}

\medskip

\begin{proof}[Lemma \ref{lem:1}]
Let $r>1$. For simplicity of notation, we write $s_p(r)\equiv s_p^{\uparrow}(D-rA)$. Define the matrix $M_r$ as
	\begin{equation*}
	M_{r}=-s_p(r)I_n+D-rA = D_{-s_p(r)} - rA
	\end{equation*}
	where we recall the notation $D_\tau=D+\tau I_n$.
	Note that one has $s_p^{\uparrow}(M_r) = 0$.
	Letting $\tilde{M}_{r} \equiv D_{-s_p(r)}^{-1/2}M_{r}D_{-s_p(r)}^{-1/2} $, we then have
	\begin{equation*}
	L^{\rm sym}_{-s_p(r)} = \frac{1}{r}\left[I_n - \tilde{M}_{r}\right].
	\end{equation*}
	Recalling that $L_{\tau}^{\rm rw}$ and $L^{\rm sym}_{\tau}$ share the same spectrum, one has in particular:
	\begin{equation*}
	s_p^{\downarrow}(L_{-s_p(r)}^{\rm rw}) = s_p^{\downarrow}(L^{\rm sym}_{-s_p(r)}) = \frac{1}{r}s_p^{\downarrow}\left(I_n - \tilde{M}_{r}\right) = \frac{1}{r}\left(1-s_p^{\uparrow}(\tilde{M}_{r})\right).
	\end{equation*}
	Thus, proving the lemma amounts to proving that $s_p^{\uparrow}(\tilde{M}_{r})=0$, which we do now by first proving that $s_p^{\uparrow}(\tilde{M}_{r})\leq0$ and then that $s_p^{\uparrow}(\tilde{M}_{r})\geq0$.
	
	Denote by $X=(\bm{x}_1|\ldots|\bm{x}_p)\in\mathbb{R}^{n\times p}$ the matrix concatenating the eigenvectors associated to the $p$ smallest eigenvalues of $M_{r}$. One has:
	\begin{align*}
 \forall ~1 \leq q \leq p, \qquad	M_{r} \bm{x}_q = s_q^{\uparrow}(M_{r}) \bm{x}_q \quad\text{  and   }\quad 
s_q^{\uparrow}(M_{r}) \leq s_p^{\uparrow}(M_{r}) = 0.
\end{align*}
	
	Now define the vector space $V$ as  $V=\text{span}\left(D_{-s_p(r)}^{1/2}X\right)$. Since $s_p(r)$ is strictly negative by hypothesis, $D_{-s_p(r)} \succ 0$ (is positive definite), which in turn implies that  ${\rm dim}(V) = p$. By definition of $V$, $\forall~\bm{z}\in V,~\exists~ \bm{u}\in \mathbb{R}^p : \bm{z} = D_{-s_p(r)}^{1/2}X\bm{u}$.
	From the Courant-Fischer theorem, we can write:
	\begin{align*}
	s_p^{\uparrow}(\tilde{M}_{r}) &= \underrel{U : {\rm dim}(U) = p}{\rm min}~\underrel{\bm{z}\in U,\bm{z}\neq 0}{\rm max}\frac{\bm{z}^T\tilde{M}_{r}\bm{z}}{\bm{z}^T\bm{z}}
	\leq \max_{\bm{z}\in V, \bm{z}\neq 0} \frac{\bm{z}^T\tilde{M}_{r}\bm{z}}{\bm{z}^T\bm{z}} = \max_{\bm{u}\in \mathbb{R}^p, \bm{u}\neq 0} \frac{\bm{u}^TX^TD_{-s_p(r)}^{1/2}\tilde{M}_{r}D_{-s_p(r)}^{1/2}X\bm{u}}{\bm{u}^TX^TD_{-s_p(r)}X\bm{u}}
	\end{align*}
	\textit{i.e.}:
	\begin{align*}
	  s_p^{\uparrow}(\tilde{M}_{r}) \leq  \max_{\bm{u}\in \mathbb{R}^p, \bm{u}\neq 0}\frac{\bm{u}^TX^TM_{r}X\bm{u}}{\bm{u}^TX^TD_{-s_p(r)}X\bm{u}} = \max_{\bm{u}\in \mathbb{R}^p, \bm{u}\neq 0}\frac{\sum_{q= 1}^p (\bm{x}_q^T \bm{u})^2 s_q^{\uparrow}(M_r)}{\bm{u}^TX^TD_{-s_p(r)}X\bm{u}}\leq 0
	\end{align*}
	where for the last step we exploited $s_q^{\uparrow}(M_r) \leq 0$ and $D_{-s_p(r)} \succ 0$. We thus conclude that $s_p^{\uparrow}(\tilde{M}_{r}) \leq 0$. 
	
	To prove that the equality holds, we exploit the second relation of the  Courant-Fischer theorem. We define $\bar{X}  = (\bm{x}_p|\ldots|\bm{x}_n) \in \mathbb{R}^{n\times(n-p+1)}$ the matrix concatenating the  eigenvectors associated to the $n-p+1$ largest eigenvalues of $M_{r}$. For $q \geq p$, $s_q^{\uparrow}(M_r) \geq 0$. We further define $W=\text{span}\left(D_{-s_p(r)}^{1/2}\bar{X}\right)$, satisfying ${\rm dim}(W) = n-p+1$. We can write:
	\begin{align*}
	s_p^{\uparrow}(\tilde{M}_{r}) &= \underrel{U :~ {\rm dim}(U) = n-p+1}{\rm max}~\underrel{\bm{z}\in U, \bm{z}\neq 0}{\rm min} \frac{\bm{z}^T \tilde{M}_{r}\bm{z}}{\bm{z}^T\bm{z}}
	\geq \min_{\bm{z}\in W, \bm{z}\neq 0}\frac{\bm{z}^T \tilde{M}_{r}\bm{z}}{\bm{z}^T\bm{z}}
	\end{align*}
	\textit{i.e.}:
	\begin{align*}
	s_p^{\uparrow}(\tilde{M}_{r}) \geq \min_{\bm{u}\in \mathbb{R}^{n-p+1}, \bm{u}\neq 0} \frac{\bm{u}^T \bar{X}^T M_r \bar{X}\bm{u}}{\bm{u}^T \bar{X}^T D_{-s_p(r)} \bar{X}\bm{u}} = \min_{\bm{u}\in \mathbb{R}^p, \bm{u}\neq 0}\frac{\sum_{q= p}^n (\bm{x}_q^T\bm{u})^2 s_q^{\uparrow}(M_r)}{\bm{u}^T\bar{X}^TD_{-s_p(r)}\bar{X}\bm{u}}\geq 0.
	\end{align*}
	As a consequence, $s_p^{\uparrow}(\tilde{M}_{r}) \geq 0$. 
	Combining both inequalities, we obtain that $s_p^{\uparrow}(\tilde{M}_{r}) = 0$.
	
The fact that the eigenvectors are shared comes from the following. Let $\bm{x}_p(r)$ be the eigenvector of $D-rA$ associated to $s_p(r)$, \textit{i.e.},  $[D - rA]\bm{x}_p(r) = s_p(r)\bm{x}_p(r)$. This can be re-written as $L_{-s_p(r)}^{\rm rw}\bm{x}_p(r) = \frac{1}{r}\bm{x}_p(r)$.
	
\end{proof}

\begin{proof}[Lemma \ref{lem:3}]
    Once again, let us write $s_p(r) = s_p^{\uparrow}(D-rA)$ to lighten notations. Let $p, r_p$ be an integer $\geq2$ and a scalar $> 1$ (if they exist) such that $s_p(r_p) < 0$. Let $r\geq r_p$.
    As $s_p(r)$ is an eigenvalue, $\exists\;\bm{x}_p(r)$ with $||\bm{x}_p(r)||^2=1$ such that:
	\begin{equation*}
	[D - rA]\bm{x}_p(r) = s_p(r)\bm{x}_p(r),
	\end{equation*}
	which implies in particular that $\bm{x}_p(r)^T A \bm{x}_p(r)=\frac{1}{r}\bm{x}_p(r)^TD_{-s_p(r)}\bm{x}_p(r)$. 
	As we suppose $s_p(r)$ to be simple, we can apply the eigenvalue perturbation theorem~\citep[see, for instance][]{greenbaum2019first}:
	\begin{equation}
	\partial_r s_p(r) = -\bm{x}_p^T(r)A\bm{x}_p(r) = -\frac{1}{r}\bm{x}_p(r)^TD_{-s_p(r)}\bm{x}_p(r).
	\label{eq:derivative_sp}
	\end{equation}
	In Equation~\eqref{eq:derivative_sp}, $D_{-s_p(r_p)} \succ 0$ (as $s_p(r_p)<0$ by hypothesis), consequently $\left.\partial_r s_p(r)\right|_{r=r_p}<0$.
	We now want to show that for all $r > r_p$, $\partial_r s_p(r) < 0$. We proceed with a proof by contradiction. 
	
	\indent Suppose that there exists a value $r' > r_p$ such that $\left. \partial_r s_p(r)\right|_{r=r'} \geq 0$. From Equation~\eqref{eq:derivative_sp} it follows that a necessary (but not sufficient) condition to be verified is that $s_p(r') {\geq} 0$. From a continuity argument on the function $s_p(r)$, and as $s_p(r_p) < 0$ and $s_p(r') \geq 0$ with $r'>r_p$, there exists $r'' \in (r_p,r')$ such that $\left.\partial_r s_p(r)\right|_{r=r''} > 0$ and $s_p(r'') < 0$. Invoking once again Equation \eqref{eq:derivative_sp},  no such $r''$ can exist, invalidating the hypothesis we made by absurd. We thus conclude that
	\begin{align}
	    \forall~r\geq r_p, \quad \partial_r s_p(r) < 0,
	\end{align}
	finishing the proof.
	
\end{proof}

\begin{proof}[Lemma \ref{lem:2}]
Consider the values $\{\zeta_p\}_{2\leq p \leq k}$ as defined in Eq.\eqref{eq:zeta}. As $\mathcal{G}$ is generated from a DC-SBM we know from Claim~\ref{claim:1} that these $k-1$ values of $\zeta_p$ exist with high probability. Define $\zeta_{k+1} = \sqrt{c\Phi}$ and fix $p$ an integer such that $2\leq p  \leq k$. Let us write once more $s_p(r) = s_p^{\uparrow}(D-rA)$ to lighten notations. Note that $s_p^{\uparrow}(H_{\zeta_p}) = 0$, $\forall~2\leq p\leq k$ and $s_{k+1}^{\uparrow}(H_{\zeta_{k+1}}) = o_n(1)$, as shown in \citep{saade2014spectral}. We thus have $s_p(\zeta_p) < 0$, $\forall~2\leq p \leq k+1$ and it is isolated by hypothesis. We can apply Lemma~\ref{lem:3}, for $r_p = \zeta_p$: for all $r \geq \zeta_p$, $\partial_r s_p(r) < 0$ and $s_p(r) < 0$. Consequently, there exists a unique value of $r'\geq r$, satisfying $s_{p+1}(r') = s_p(r)$.
From lemma~\ref{lem:1}, we know that:
\begin{itemize}
\item as $s_p(r)<0$: $\bm{x}_p(r)$, the eigenvector of $D-rA$ associated to $s_p(r)$, is also the eigenvector of $L_{-s_p(r)}^{\rm rw}$ associated to its $p$-th largest eigenvalue, $1/r$.
\item as $s_{p+1}(r')<0$: $\bm{x}_{p+1}(r')$, the eigenvector of $D-r'A$ associated to $s_{p+1}(r')$, is also the eigenvector of $L_{-s_{p+1}(r')}^{\rm rw}=L_{-s_p(r)}^{\rm rw}$ associated to its $(p+1)$-th largest eigenvalue, $1/r'$.
\end{itemize}
Thus, the $p$-th (resp. $(p+1)$-th) largest eigenvalue of $L_{-s_p(r)}^{\rm rw}$ is $1/r$ (resp. $1/r'$). 
Consider $r \geq \zeta_p$. By hypothesis, $s_p(r)$  is an isolated eigenvalue, that is, we can write
\begin{align*}
O_n(1) = s_{p+1}(r) - s_{p}(r) = s_{p+1}(r) - s_{p+1}(r') = \int_{r'}^r dx~ \partial_x s_p(x) = \kappa(r - r')
\end{align*} 
for some constant $\kappa = O_n(1)$ independent of $n$, representing the average value of $\partial_x s_p(x)$ on the integration interval.
The constant $\kappa = O_n(1)$ because for any $r,r' = O_n(1)$, we have that $s_p(r), s_p(r') = O_n(1)$. 
The eigengap for $L_{-s_p(r)}^{\rm rw}$ is $1/r-1/r' = (r'-r)/(rr')$, thus
\begin{align*}
O_n(1) = O_n\big(s_{p+1}(r)-s_p(r)\big) = O_n(r-r') = O_n\left(\frac{1}{r} - \frac{1}{r'}\right) = O_n(1).
\end{align*}
So, if $s_p(r)$ is isolated, the eigenvalue $1/r$ of $L_{\tau}^{\rm rw}$ is isolated as well.

On the other hand, if $s_p(r)$ is in the bulk, then $\exists~q~:|s_p(r)-s_q(r)| = o_n(1)$. By an argument of continuity on $s_p(r)$, one can analogously define an $r'$ satisfying $s_p(r) = s_q(r)$, concluding that in this case $|r-r'| = o_n(1)$ and so eigenvalues in the bulk are mapped into eigenvalues in the bulk.

\end{proof}

\section{Fast Estimate of the ${\zeta_p}$}
\label{app:zeta}

We here describe a fast algorithm to estimate the values of the $\zeta_p$ using the Courant-Fisher theorem.

Let us denote by $X^T_{r_t} = (\bm{x}^T_1({r_t}),\ldots, \bm{x}^T_p({r_t}))^T\in \mathbb{R}^{n\times p}$ where $\bm{x}_p({r_t})$ is the eigenvector of $H_{r_t}$ corresponding to the $p$-th smallest eigenvalue, $s_p^{\uparrow}(H_{r_t})$, while $S_{r_t} = {\rm diag}\left(s_1^{\uparrow}(H_{r_t}),\ldots,s_p^{\uparrow}(H_{r_t})\right)$. For another value $r'\neq r_t$, and applying the Courant-Fischer theorem, we can write
\begin{align*}
s_p^{\uparrow}(H_{r'}) &= \underrel{U : {\rm dim}(U) = p}{\rm min} ~\underrel{\bm{z} \in U, \bm{z}\neq 0}{\rm max}\frac{\bm{z}^TH_{r'}\bm{z}}{\bm{z}^T\bm{z}} \leq \underrel{\bm{u} \in \mathbb{R}^p}{\rm max}\frac{\bm{u}^TX_{r_t}^TH_{r'}X_{r_t}\bm{u}}{\bm{u}^T\bm{u}} = s_1^{\downarrow}(X_{r_t}^T H_{r'}X_{r_t}) 
\end{align*}
i.e.:
\begin{align*}
s_p^{\uparrow}(H_{r'}) \leq s_1^{\downarrow}(X_{r_t}^T[(r'^2-{r_t}^2)I_n - (r'-{r_t})A + H_{r_t}]X_{r_t}) = (r'^2 - {r_t}^2) + s_1^{\downarrow}(S_{r_t} - (r'-{r_t})X_{r_t}^TAX_{r_t}).
\end{align*}
We can further simplify the earlier expression by exploiting the identity:
\begin{align*}
S_{r_t} = X_{r_t}^TH_{r_t}X_{r_t}  = ({r_t}^2-1)I_p + X_{r_t}^TDX_{r_t} - {r_t}X_{r_t}^TAX_{r_t}.
\end{align*}
We thus obtain
\begin{align}
s_p^{\uparrow}(H_{r'}) \leq \frac{1}{r_t}\left[(r'-r_t)(1+r'r_t)+ s_1^{\downarrow}\big(({r_t}-r')X_{r_t}^TDX_{r_t} + r'S_{r_t}\big)\right] \equiv \frac{f_{r_t}(r')}{r_t}.
\label{eq:ineq}
\end{align}
We now study the function $f_{r_t}(r')$ for $r_t \in (\zeta_p, \sqrt{\rho(B)})$ and define $r_{t+1} \in (\zeta_p, r_t)$ as the solution (if it exists) to
\begin{align}
f_{r_t}(r_{t+1}) = 0.
\label{eq:fixed_fast}
\end{align} 
The idea is to iteratively approach $\zeta_p$ 
from the right and substitute $r_t \leftarrow r_{t+1}$. If $r_{t+1} \in (\zeta_p,r_t)$ as defined above exists, then $\{r_t\}_{t \geq 0}$ is a lower-bounded decreasing sequence: it thus converges to a limit $r_{\infty}$ (potentially different from $\zeta_p$).
Exploiting \citep[Theorem in I.4]{rellich1969perturbation}, denoting with $A,B$ two Hermitian matrices and $\lambda$ a scalar, $s_1^{\downarrow}(A+\lambda B)$ is a convex function of $\lambda$. As a consequence, the function $f_{r_t}(r')$ is convex and has either no root or two roots. Since $\lim_{|r'|\to \infty} f_{r_t}(r') = +\infty$, it is enough to find a value of $r'$ for which $f_{r_t}(r') < 0$ to prove that this function has two roots. With a straightforward computation, one can verify that $f_{r_t}(r_t) < 0 $, so $f_{r_t}(r')$ has two roots, satisfying $r_{t+1} < r_t$ and $r^+ > r_t$.
By construction, since $s_p^{\uparrow}(H_{r'})$ is negative in the considered interval and $s_p^{\uparrow}(H_{r'}) \leq f_{r_t}(r')/r_t$, if $f_{r_t}(r_{t+1}) = 0$, then $\zeta_p \leq r_{t+1}$. Consequently, $f_{r_t}(r_{t+1}) = 0$ has a unique solution satisfying
\begin{align*}
\zeta_p \leq r_{t+1} < r_t
\end{align*}
and the algorithm converges to $r_{\infty} =\displaystyle \lim_{t \to \infty} r_t \geq \zeta_p$. We are left to prove that $r_{\infty} = \zeta_p$. By convergence of $r_t$, we have $r_{t+1} - r_t =  o_{t}(1)$. Plugging this relation solution into Equations~(\ref{eq:ineq},\ref{eq:fixed_fast}), we obtain
\begin{align*}
    s_1^{\downarrow}(r_{t+1}S_{r_{t+1}}) = r_{t+1}s_p^{\uparrow}(H_{r_{t+1}}) = o_t(1).
\end{align*}
Since $s_p^{\uparrow}(H_r) = 0$ has a unique solution ($r = \zeta_p$) in the interval $r \in (1,\sqrt{\rho(B)})$, we obtain $r_t = \zeta_p + o_t(1)$ and so
\begin{align*}
   r_{\infty} = \zeta_p.
\end{align*}

The initial value of $r$ can be chosen as $r_{0} = \zeta_{p+1}$ (setting $\zeta_{k+1} = \sqrt{\rho(B)}$ that certainly falls in the right interval for all the $\zeta_p$.

\begin{figure}
	\centering
	\includegraphics[width=0.5\columnwidth]{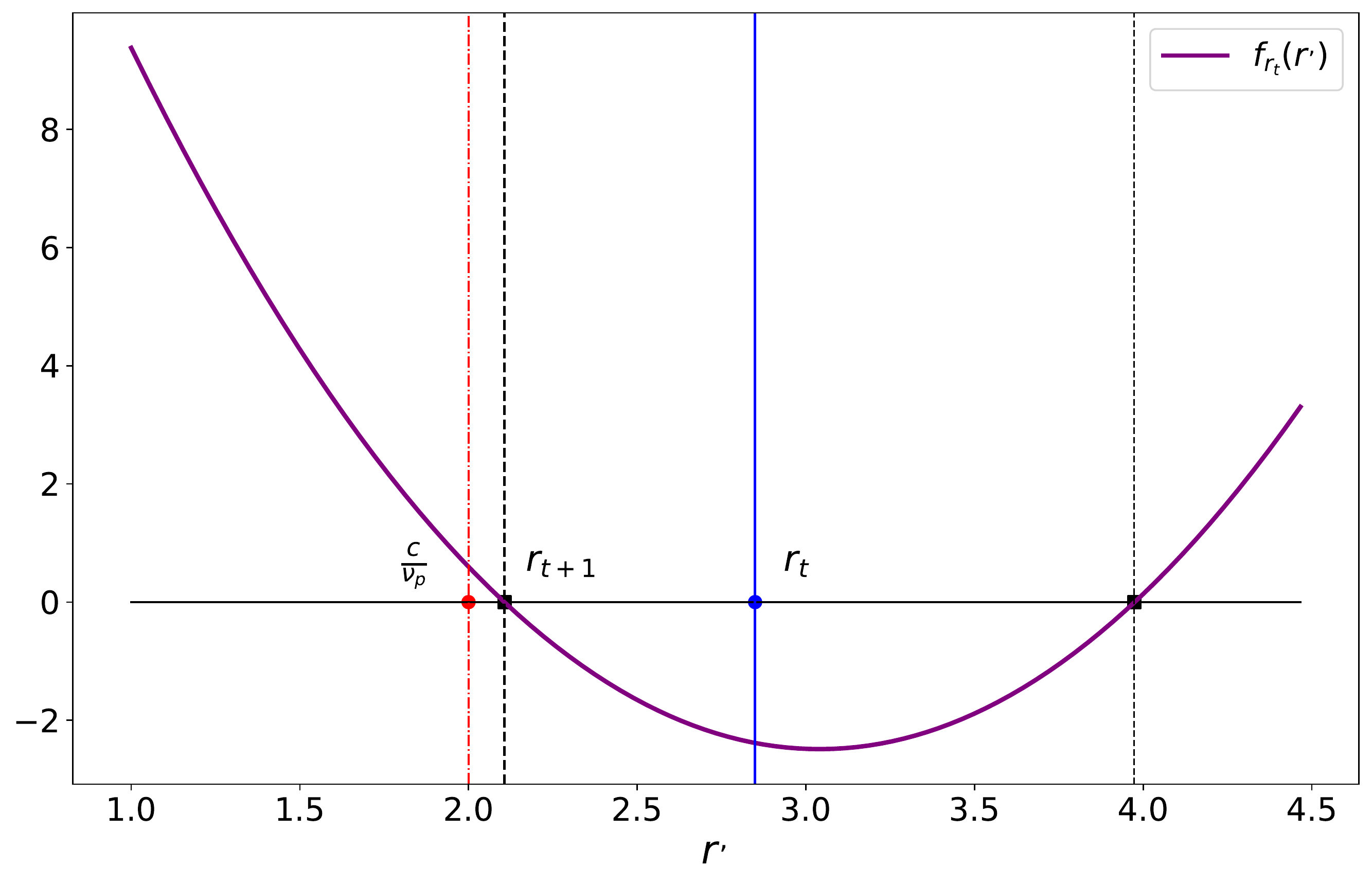}
	\caption{Plot of the $f_{r_t}(r')$ (in purple). The blue continuous line indicates the value of $r_t$, the red dashed dotted line the theoretical value of $\zeta_p = c/\nu_p$, the two dashed black lines are the two roots of $f_{r_t}(r')$, the smaller of which is $r_{t+1}$. For this simulation $n = 50\,000$, $k = 2$, $c = 5$, $c_{\rm out} = 2.5$, $r = \sqrt{c\Phi}$, $\theta_i \sim [\mathcal{U}(3,10)]^3$, $\bm{\pi} \propto I_k$.}
	\label{fig:numerical_method}
\end{figure}


\end{document}